\documentclass[10pt,twocolumn,letterpaper]{article}

\usepackage{float}
\usepackage[pagenumbers]{cvpr} 

\usepackage{ragged2e}
\definecolor{cvprblue}{rgb}{0.21,0.49,0.74}
\usepackage[pagebackref,breaklinks,colorlinks,allcolors=cvprblue]{hyperref}
\usepackage{amsmath}
\usepackage{amsthm}
\usepackage{amssymb}

\usepackage{algorithm}
\usepackage{algpseudocode}

\newtheorem{theorem}{Theorem}

\newtheorem{lemma}[theorem]{Lemma}

\theoremstyle{definition}
\newtheorem{definition}[theorem]{Definition}

\def\paperID{13666} 
\def\confName{CVPR}
\def\confYear{2025}

\title{Noise-Tolerant Coreset-Based Class Incremental Continual Learning}

\author{Edison Mucllari\\
University of Kentucky\\
Lexington, KY, USA\\
{\tt\small edison.mucllari@uky.edu}
\and
Aswin Raghavan\\
SRI International\\
Princeton, NJ, USA\\
{\tt\small aswin.raghavan@sri.com}
\and
Zachary Daniels\\
SRI International\\
Princeton, NJ, USA\\
{\tt\small zachary.daniels@sri.com}
}

\begin{document}
\maketitle
\begin{abstract}

Many applications of computer vision require the ability to adapt to novel data distributions after deployment. Adaptation requires algorithms capable of continual learning (CL).  
Continual learners must be plastic to adapt to novel tasks while minimizing forgetting of previous tasks. 
However, CL opens up avenues for noise to enter the training pipeline and disrupt the CL. 
This work focuses on label noise and instance noise in the context of class-incremental 
learning (CIL), where new classes are added to a classifier over time, and there is no access to external data from past classes.
We aim to understand the sensitivity of CL methods that 
work by replaying items from a memory constructed using the idea of Coresets. 
We derive a new bound for the robustness of such a method to uncorrelated instance noise under a general additive noise threat model, revealing several insights. Putting the theory into practice, we create two continual learning algorithms to construct noise-tolerant replay buffers. We empirically compare the effectiveness of prior memory-based continual learners and the proposed algorithms under label and uncorrelated instance noise on five diverse datasets. 
We show that existing memory-based CL are not robust whereas the proposed methods exhibit significant improvements in maximizing classification accuracy and minimizing forgetting in the noisy CIL setting.
\end{abstract}
\section{Introduction}
\label{sec:intro}

Many real-world applications of computer vision (CV) require the ability to adapt to novel data distributions after deployment due to changes in environmental factors, sensor characteristics, and task-of-interest: autonomous vehicles need to adapt to novel weather patterns; biomedical imaging systems need to adapt to novel diseases; remote sensing systems need to adapt to improvements in sensor technologies; and foundation models need to adapt to novel data sources. Most machine learning (ML) systems assume stationary and matching distributions between the data seen during training and deployment. The aforementioned applications require algorithms capable of performing continual (or lifelong) learning (CL) \cite{thrun1998lifelong}, where models must be plastic enough to adapt to new tasks while being stable enough to minimize catastrophic forgetting of previous tasks.

CL is a topic of great interest to the AI/ML community \cite{chen2022lifelong,khodaee2024knowledge}. Understanding the robustness of CL strategies to noise encountered during the data collection process is a significant underexplored problem since CL exposes a unique attack surface to the training pipeline. We identify three types of noise that may appear during CL training:
\begin{itemize}
\item \textit{Label Noise: } Data annotation is expensive, especially in a CL system, which requires continuous data collection and labeling. To reduce human labeling effort, it is useful to collect a batch of data, that is \emph{mostly} from a single class, and assign a single (imperfect) label to the entire batch. Label noise also occurs in crowd-sourced labels from non-subject matter experts. It would be advantageous if a CL algorithm can learn around large amounts of mislabeled samples, without assuming that the mislabeled samples belong to a known (seen) class.
\item \textit{Uncorrelated Instance Noise: } Sensors can degrade and fail over time. Further, the degradation can vary with environmental conditions, leading to a mix of clean and noisy instances. Novel instance noise can occur throughout the lifetime of a fielded sensor or system. 
\item \textit{Adversarial Poisoning: } CL opens an avenue for poisoning attacks, e.g., backdoor trigger attacks \cite{saha2020hidden} and forgetting attacks (aka brainwashing) \cite{umer2020targeted,li2023pacol,abbasi2024brainwash}. These attacks turn the advantages of CL into critical system flaws. 
\end{itemize}
This work focuses on CL under the first two scenarios: label noise and instance noise in the context of class-incremental learning (CIL) \cite{van2019three}. In CIL, new classes are added to a classifier over time, and there is no access to external data from past classes. 
Specifically, we aim to understand, \emph{theoretically and empirically}, the sensitivity of CL methods that work by replaying items from a memory constructed using the idea of \emph{Coresets}. Specifically, we consider the CRUST algorithm \cite{mirzasoleiman2020coresets} for finding Coresets. CRUST has been empirically and theoretically shown to have high label noise-tolerance in the static ML setting.


The major contributions of this work are as follows:
\begin{itemize}
\item We derive a new bound for the robustness of CRUST to uncorrelated instance noise under a general additive noise threat model. The theoretic analysis as a function of the proportion of corrupted data reveals the relative hardness of instance noise and label noise, provides a theoretical guarantee on the required number of training iterations and learning rate, and advances the applicability of CRUST in general.
\item Putting the theory into practice, we create two continual learning algorithms that extend CRUST to the CL setting (``Continual CRUST'' and ``Continual Cosine-CRUST'') to construct noise-tolerant replay buffers. The methods incrementally train both the classifier and the Coresets in a theoretically-sound manner. 
\item We empirically compare the effectiveness of prior memory-based continual learners and the proposed algorithms under label and uncorrelated instance noise on five diverse datasets including image classification benchmarks, Synthetic Aperture Radar (SAR) satellite imagery, and histology images in the medical domain. 
CRUST-based methods are evaluated for the first time in a Continual Learning CIL setting in the presence of both label noise and uncorrelated instance noise. Continual (Cosine)CRUST exhibits significant improvements in maximizing classification accuracy and minimizing forgetting in the noisy CIL setting.
\end{itemize}
\section{Preliminaries}
\label{sec:problem}

\subsection{Class Incremental Learning}
\label{sec:cil}
Class Incremental Learning (CIL) addresses the challenge of learning a classifier from an infinite stream of data with dynamic shifts in instance and label distributions. We formalize the CIL problem as follows. Let $\{(x_t, y_t)\}_{t=1}^\infty$ be a sequence of input-output pairs, viz.\! instance $x_t \in \Delta(\mathcal{X})$ (e.g., distribution over fixed size images) and labels $y_t \in \mathbb{N}$. Our labels are integers to allow for a increasing number of classes. 
The goal is to learn a classifier that is a parametrized function $f_\theta: \Delta(\mathcal{X}) \rightarrow \mathbb{N}$ that minimizes:
\begin{equation}
    \inf_\theta \mathbb{E}_{(x,y) \sim p_t(x,y)} [\mathcal{L}(f_\theta(x), y)]
\end{equation}
where $p_t(x,y)$ is the joint distribution at time $t$, and $\mathcal{L}$ is a training loss function. 
For the purpose of the theoretical analysis, we will use Mean Squared Error (MSE) as the loss. Following prior work \cite{li2020gradient}, the theory will treat the labels $y$ as ordinal values, whereas the experiments use categorical labels as in image classification benchmarks.

In practice, the CIL process is divided into experiences, each consisting of $K$ \emph{new} classes (added to the label space) with associated probabilities $p_K(x,y)$. CIL's objective is:
\begin{equation}
    \inf_\theta \sum_{k=1}^K \mathbb{E}_{(x,y) \sim p_k(x,y)} [\mathcal{L}(f_\theta(x), y)]
\end{equation}
This objective is often subject to a constraint that accuracy on previous tasks does not significantly degrade:
\begin{equation}
    s.t.\!\quad \mathbb{E}_{(x,y) \sim p_j(x,y)} [\mathcal{L}(f_\theta(x), y)] \leq \upsilon_j, \quad \forall j < k
\end{equation}
where $\upsilon_j$ are accuracy tolerances on previous experiences.

We address two natural CIL scenarios where the data for the new experience is noisy: (1) Label noise where the ground truth label $y_t$ is corrupted to a random label. We use the definition of a Corrupted Dataset introduced in \cite{li2020gradient}. (2) Additive instance noise in $x_t$ (aka pixel noise). In both cases, the bounds are a function of the fraction of noisy samples. In real-world applications, this fraction can potentially be estimated from the data. In our experiments, we vary the fraction and correlate the bounds to the impact on accuracy. 

\subsection{Coresets for Robust Training (CRUST)}
\label{sec:crust}
A Coreset $S$ for a stationary dataset $X$ with respect to a loss function $L$ is a weighted subset of $X$ that approximates the loss of the full dataset for any set of parameters $\theta$:
\begin{equation}
    |\mathcal{L}(X, \theta) - \mathcal{L}(S, \theta)| \leq \epsilon |\mathcal{L}(X, \theta)| \quad \forall \theta
\end{equation}
where $\epsilon$ is a small error parameter \cite{munteanu2018coresets}.

CRUST \cite{mirzasoleiman2020coresets} is a method for training of neural network (NN) classifiers robust to noisy labels.
CRUST uses the observation that samples with clean labels tend to have similar gradients w.r.t. the NN parameters. Let $V$ denote the complete dataset. CRUST formulates the selection of a Coreset $S$ of size $k$ as a $k$-medoids problem:
\begin{equation}
S^* = \arg \min_{S \subset V} \sum_{i \in V} \min_{j \in S} d_{ij}(W) \quad \text{s.t.} \quad |S| \leq k
\label{eq:optimization}
\end{equation}
where $d_{ij}(W) = \|\nabla \mathcal{L} (W, x_i) - \nabla \mathcal{L} (W, x_j)\|_2$ is the pairwise difference between gradients of sample $i$ from $V$ and $j$ from $S$. While the optimization does not yield the optimal rank-$k$ approximation of the Jacobian, it yields a close approximation of the Jacobian \emph{for} samples with clean labels. 
In order to improve efficiency, CRUST uses a greedy procedure using the submodular structure of the problem.
\begin{equation}
    S_t = S_{t-1} \cup \{\sup_{e \in V} \hat{F}(e | S_{t-1})\}
\end{equation}
where $F(S,W) = \sum_{i \in V} \max_{j \in S} d_0 - d'_{ij}(W)$ with $d_0$ to be a constant satisfying $d_0 \geq d_{ij}(W)$, for all $i, j \in V$, and $d'$ is the gradient of the loss wrt the input to the last NN layer. At each iteration $t$, an element $e \in V$ that maximizes the marginal utility $\hat{F}(e|S_t) = F(S_t \cup \{e\}) - F(S_t)$ is added to the Coreset $S_t$, 
with $S_0 = \varnothing$, the empty set. 
Theoretically, Coresets found by CRUST are ``mostly clean'' due to the underlying properties of the Jacobian spectrum. The classification error for training with the Coreset is bounded.

\begin{theorem}\label{thm:crust_convergence} (\textbf{Robustness to Label Flipping}) \cite{mirzasoleiman2020coresets}
Assume that gradient descent is applied to train a neural network with mean-squared error (MSE) loss on a dataset with noisy labels. 
Suppose that the Jacobian of the network is $L$-smooth. 
Assume that the dataset has label margin of $\delta$, and Coresets found by CRUST contain a fraction of $\rho < \frac{\delta}{8}$ noisy labels.
If the Coreset approximates the Jacobian by an error of at most $\epsilon \leq \mathcal{O} (\frac{\delta \alpha^2}{k \beta \log (\sqrt{k} / \rho)})$, where $\alpha = \sqrt{r_{min}} \sigma_{min} (\mathcal{J}(W, X_S)))$, $\beta = \|\mathcal{J} (W, X)\| + \epsilon$, then for $L \leq \frac{\alpha \beta}{L \sqrt{2 k}}$ and learning rate $\eta = \frac{1}{2 \beta^2}$, after $\tau \geq \mathcal{O} (\frac{1}{\eta \alpha^2} \log (\frac{\sqrt{n}}{\rho}))$ iterations the NN classifies all samples in the Coreset correctly.
\end{theorem}




where $r_{min}$ is the size of the smallest cluster over the $k$-medoids, $\sigma_{min}$ is the smallest singluar value, and the label margin $\delta$ is a dataset-specific constant \cite{li2020gradient}. Refer to the supplemental material for details. 
Theorem \ref{thm:crust_convergence} says that the number of iterations  $\tau$ to fit the Coreset is proportional to $\log \left( \frac{\sqrt{n}}{\rho} \right)$. As the noise fraction $\rho$ increases, 
$\tau$ decreases for the same size of Coreset, which may initially seem counterintuitive. 
However, with more noise, the Coreset becomes less representative of the true data distribution, making it easier to fit (see robustness due to early stopping in \cite{li2020gradient}) but potentially less useful for generalization.
Additionally, as $\rho$ increases, the approximation error bound $\epsilon$ for the Jacobian matrix grows inversely in the order $\log \left( \frac{\sqrt{k}}{\rho} \right)$.
\section{New Bound for Instance Noise}
Practical CL systems do not only face noise in the label space, but may also encounter perturbed data. Sensors may fail over time causing issues such as random noisy pixel variations to appear with some degree of regularity. Data quality is effected by environmental factors e.g.,\! shot noise in electro-optic and speckle in SAR imagery. 


Our \textbf{first contribution} is a new bound for CRUST under general additive noise in the instances, i.e.\!, $\tilde{X} = X + \text{E}_X$. The following terms $E_{J_1}$ and $E_{J_2}$ appear in the bound. They are associated with the Jacobian of the NN loss evaluated at the perturbation. For the Jacobian, we have 
\begin{align}
    \mathcal{J}(W, X + E_X)^T &= \mathcal{J}(W, X)^T + \mathcal{J}(W, E_X)^T \\
    &= \mathcal{J}(W, X)^T + E_{J_1}
\end{align}
\begin{definition} (Average Jacobian) \cite{li2020gradient}
    We define the Average Jacobian along the path connecting two points $u, v \in \mathbb{R}^p$ as 
    $$\bar{\mathcal{J}}(v,u) = \int_0^1 \mathcal{J} (u+\alpha (v-u))d \alpha$$
\end{definition}
Let $\bar{\mathcal{J}}(v,u,x)=\bar{\mathcal{J}}(v,u)\bigg|_x$ denote the average jacobian evaluated at $x$. Let $\hat{W}=W-\eta\nabla\mathcal{L}(W)$ denote a one-step gradient update to the NN weights. Then, 
\begin{align}
    \bar{\mathcal{J}}(\hat{W}, W, X + E_X) &= \bar{\mathcal{J}}(\hat{W}, W, X) + \bar{\mathcal{J}}(\hat{W}, W, E_X) \\
    &= \bar{\mathcal{J}}(\hat{W}, W, X) + E_{J_2}
\end{align}
where $E_{J_2} = \mathcal{J}(\hat{W}, W, E_X)$. Additionally, we define $E_{min}$ and $E_{max}$ as the numerical errors corresponding to the smallest and largest singular values, respectively.
In contrast to Theorem \ref{thm:crust_convergence}, our bound depends on the residual error after $T$ iterations of NN training on the Coreset.
\begin{definition} (Residual) $r_T = f(X,W_T)-Y$.
\end{definition}
\begin{theorem} (\textbf{Robustness to Data Perturbations})\label{thm:random_noise}\\
Assume that gradient descent is applied to train an NN with mean-squared error (MSE) loss on a dataset with $\delta$ fraction of perturbed data samples of the form  \text{$\tilde{X}=X+E_X$}. Suppose the Jacobian is $L$-smooth. If the Coreset 
approximates the Jacobian with an error of at most $\epsilon \leq \mathcal{O}\left(\frac{\delta \alpha^2}{k \beta \log (\delta)}\right)$, where $\alpha = \sqrt{r_{min}} \sigma_{min}(\mathcal{J}(W, X_S)) - E_{min}$ and $\beta = \|\mathcal{J}(W, X)\|_2 + \epsilon + E_{max}$,
then for $L \leq \frac{\alpha \beta}{L \sqrt{2k}}$ and a learning rate $\eta = \frac{1}{2 \beta^2}$, after {\scriptsize $$T \geq \mathcal{O}\left(\frac{1}{\eta}\left(\frac{\alpha}{2} + E_{J_2} \alpha + E_{J_1} \alpha - \eta E_{J_2} \frac{\beta^3}{2} - \eta E_{J_1} \frac{\beta}{3}\right)^{-1} \log \frac{\|r_0\|_2}{\nu}\right)$$ } iterations, 
$\nu > 0$ is a constant that ensures that $\|r_T\|_2 \leq \nu$, the NN classifies samples in the Coreset correctly.
\end{theorem}


The full proof of the theorem is included in the supplemental material. The theorem illustrates an important relationship between the fraction of perturbed data, $\delta$, and the error in approximating the Jacobian, $\epsilon$. As $\delta$ increases (more noisy data), $\epsilon$ also rises, but at a slower rate compared to the case of label noise. While noise does impact classification accuracy, its effect is more gradual.
Similarly, as $\delta$ increases, the number of iterations required to fit the Coreset decreases due to fewer clean examples available for learning. However, this decrease in iterations is not as rapid as seen with label noise. 
The provable guarantee extends the practical application of CRUST, providing confidence in the performance across a wide range of real-world data conditions. Armed with the theory, next we operationalize CRUST to create a novel method for CIL with noisy data. 

\section{Continual CRUST for Class-Incremental Continual Learning}
CRUST is designed for stationary ML (time invariant). Our \textbf{second contribution} is Continual CRUST, a method for Class-Incremental Continual Learning. We leverage CRUST's ability to select compact Coresets that contain essential information about a given dataset. Compact Coresets can be aggregated and refined over the lifetime of a learner allowing for efficient knowledge preservation and transfer across different learning experiences. Continual CRUST can handle both label noise and instance noise due to the theoretical guarantees in the previous sections.

Following previous work on the challenging one-class incremental learning setting \cite{asadi2022tackling},
we assume that each experience contains one new class. 
Our method trivially generalizes to settings where multiple classes may be added per experience. The overall flow of our method is shown in Algorithm~\ref{alg:continualCRUST}. The idea is to maintain and update Coresets denoted \emph{S[i]} for each class $i$ throughout CIL. 

The classifier is always trained on the combined
Coresets from all previous experiences. The Coreset for a new class is initialized to the entire data of the new experience. First, the classifier is trained for $m$ iterations in order to first distill the new class into the classifier. Then, the Coreset for the new class is refined over $n$ iterations by repeatedly applying CRUST on the gradients of the classifier. As in \cite{katharopoulos2018not}, for efficiency we only use the gradients of the final NN layer. 

Our \textbf{third contribution} is an improvement to Continual CRUST. As studied in \cite{mirzasoleiman2020coresets}, clean samples are well separated from noisy samples in the Jacobian spectrum in the case of label noise. We hypothesize that noisy examples will form small clusters of outliers in the spectrum. Further, for high dimensional data $X$, we hypothesize that noisy examples will be neither too similar nor too dissimilar to clean samples (in the Jacobian spectrum). Furthermore, in the CIL setting, we expect the new class data will exhibit gradients with different properties than those of previously seen classes. Samples from old classes should be well-understood by the network whereas training dynamics associated with new class may not yet have converged. Similarly, we expect out-of-distribution samples associated with labels from future classes to exhibit less stable gradients which may make them easier to filter out. 


We might be able to detect all of the above as small clusters by performing spectral clustering on the gradients using cosine distance as the metric. Cosine distance is a well-established metric that quantifies directional similarity between vectors, regardless of their magnitude. 
\begin{equation}
\text{cosine\_distance}(a, b) = 1 - \frac{a \cdot b}{\|a\|_2 \|b\|_2}
\label{eq:cos_dist}
\end{equation}
In contrast to CRUST that solves a single submodular optimization problem, we cluster the samples based on 
cosine similarity (using spectral clustering), filter out small 
clusters, which we hypothesize are noisy samples of new classes, and solve the submodular optimization within each cluster to form our final Coreset.
Samples for the Coreset are drawn proportional to the cluster size. When the gradients are well-behaved, we hypothesize that the filtering of likely noisy samples and optimization based on diverse clusters (instead of global optimization) may lead to improved performance on the downstream tasks.


\algdef{SE}[REPEATN]{REPEATN}{END}[1]{\algorithmicrepeat\ #1 \textbf{times}}{\algorithmicend}

\begin{algorithm}
\caption{ContinualCRUST and ContinualCosineCRUST (Lines 17 and 20, respectively).}\label{alg:continualCRUST}
\begin{algorithmic}[1]
\State \textbf{Given}: 1. Neural Net $f_\theta$ with $L$ layers and trainable parameters $\theta_1,\ldots,\theta_L$.
\State \textbf{Input}: Dataset for the current experience $\mathcal{D}_i=(X,Y)$.
\State \Comment{$i$-th experience contains class $i$ only}
\State \textbf{Hyperparameters}: Learning rate $\eta$; Minimum Coreset size $n_A$; Num. iterations $m$ and $n$.
\State \textbf{Learns}: Per-Class Coreset dictionary $S$ (initial empty). 
\For{$i$ in \texttt{CIL Experiences}}
    \If{Coreset $S[i]$ is empty}
        \State $S[i] \gets \mathcal{D}_i$
        \REPEATN{ $m$ times}
            \State $f \gets$ SGD($\mathcal{L}, \theta, \eta, S$)
            \State \Comment{Update classifier using all Coresets $S$}
        \END
    \EndIf
    \REPEATN{ $n$ times}
        \State $G_i \gets \nabla(\mathcal{L}_{CE}(D_i), \theta_{L-1})$
        \State \Comment{Calculate Gradient wrt last layer params.}

        \State \textbf{For CRUST:} 
        \State \indent $S[i] \gets$ CRUST($D_i$, $G_i$)
        \State \Comment{Update Coreset $S[i]$}

        \State \textbf{For CosineCRUST:}
        \State \indent Cluster $G_i$ using Cosine Distance Eq. \ref{eq:cos_dist} 
        \State \indent Keep samples $(D_i^c, G_i^c)$ for clusters $c \in C$
        \State \indent of size greater than $n_A$
        \State \indent $S[i] \gets \bigcup\limits_{c \in C}^{\infty} $ CRUST($D_i^{c}$, $G_i^{c}$)
        \State \Comment{Update Coreset $S[i]$}
                
        \State Update $\theta$ of classifier using all Coresets $S$
    \END
\EndFor
\end{algorithmic}
\end{algorithm}


\section{Experiments and Results}
\label{sec:experiments}
To understand the effect of noisy data on continual learning algorithms and to understand the quality of the Coresets discovered by Continual CRUST and Continual CosineCRUST, we conduct a thorough set of studies related to the amount of label noise, the amount of perturbed data, and the purity of the Coresets. 

\subsection{Datasets}
\begin{figure}[t]
  \centering
   \includegraphics[width=0.9\linewidth]{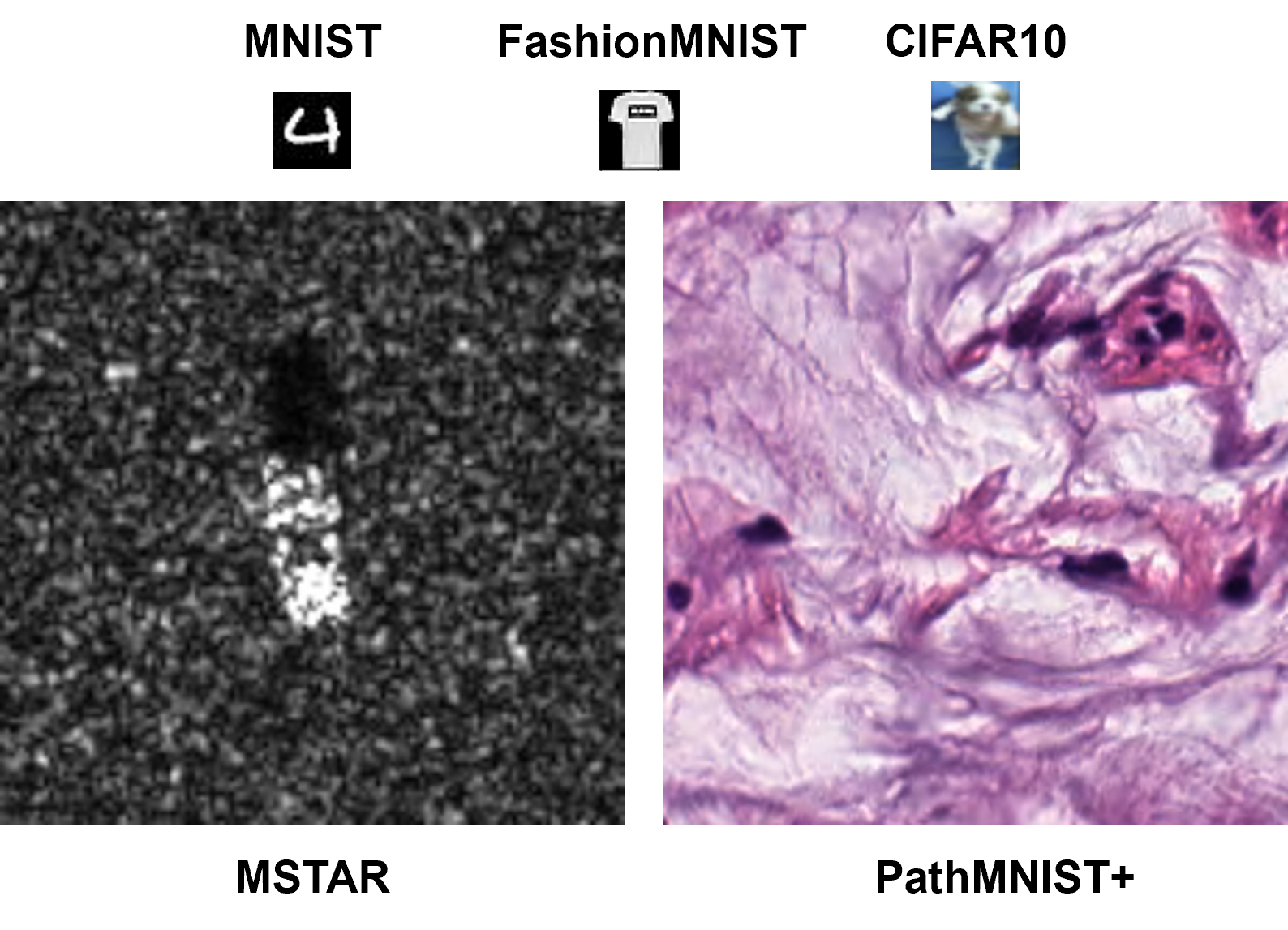}
   \caption{Representative samples from the datasets used for evaluation, which range in image size (28x28 up to 224x224), sample size (100s to 10,000s), color profile (grayscale, RGB), and modality (handwriting, natural images, SAR aerial imagery, pathology)}
   \label{fig:representative samples}
\end{figure}

These tests helped us evaluate how well our methods perform under different challenging (yet realistic) conditions. Example data samples are shown in Fig. \ref{fig:representative samples}.

\textbf{Toy Domains}: The \textbf{MNIST} dataset \cite{1571417126193283840} consists of 60,000 training and 10,000 test grayscale images which need to be classified into ten handwritten digits, each 28x28 pixels in size. The \textbf{FashionMNIST} dataset \cite{xiao2017fashion} is a collection of 60,000 training images and 10,000 test images, each 28x28 pixels in size. The dataset involves classifying ten categories of clothing items. Most experiments on the toy domains have been delegated to the supplementary section. 

\textbf{Challenge Domains}: The \textbf{CIFAR10} dataset \cite{krizhevsky2009learning} consists of 50,000 training and 10,000 test color images, where every image is 32x32 pixels, and the data is divided into ten generic object classes. The \textbf{MSTAR} dataset \cite{keydel1996mstar} consists of 5,173 Synthetic Aperture Radar (SAR) images, resized to 224x224 pixels. We follow the standard operating conditions, classifying ten vehicles from aerial SAR imagery where training and test sets are split based on elevation angle. We use a subset of the \textbf{PathMNIST+} dataset \cite{yang2023medmnist} consisting of 18,000 training and 2,700 test non-overlapping RGB patches from hematoxylin and eosin stained histological images, where every patch is 224x224 pixels. PathMNIST+ involves classifying nine types of tissue.

\begin{table*}[ht]
\centering
\resizebox{0.9\textwidth}{!}{
\begin{tabular}{c|cc|cc|cc|cc|cc|cc|}
\cline{2-11}
\textbf{} & \multicolumn{2}{c|}{\textbf{MNIST}} & \multicolumn{2}{||c|}{\textbf{FashionMNIST}} & \multicolumn{2}{||c|}{\textbf{CIFAR10}} & \multicolumn{2}{||c|}{\textbf{MSTAR}} & \multicolumn{2}{||c|}{\textbf{PathMNIST+}} &  \multicolumn{2}{|c}{} \\ \hline
\multicolumn{1}{|c|}{\textbf{Algorithm}} & \multicolumn{1}{c|}{\textit{\textbf{Acc}}} & \textit{\textbf{Forg.}} & \multicolumn{1}{||c|}{\textit{\textbf{Acc}}} & \textit{\textbf{Forg.}} & \multicolumn{1}{||c|}{\textit{\textbf{Acc}}} & \textit{\textbf{Forg.}} & \multicolumn{1}{||c|}{\textit{\textbf{Acc}}} & \textit{\textbf{Forg.}} & \multicolumn{1}{||c|}{\textit{\textbf{Acc}}} & \textit{\textbf{Forg.}} & \multicolumn{1}{|c|}{\textit{\textbf{Sig Acc?}}}
& \multicolumn{1}{|c|}{\textit{\textbf{Sig Forg?}}}\\ \hline
\multicolumn{1}{|c|}{\begin{tabular}[c]{@{}c@{}}Naive \\ Sequential \\ Learner\end{tabular}} & \multicolumn{1}{c|}{\begin{tabular}[c]{@{}c@{}}0.11\\ $\pm$0.0\end{tabular}} & \begin{tabular}[c]{@{}c@{}}1.0\\ $\pm$0.0\end{tabular} & \multicolumn{1}{||c|}{\begin{tabular}[c]{@{}c@{}}0.11\\ $\pm$0.0\end{tabular}} & \begin{tabular}[c]{@{}c@{}}1.0\\ $\pm$0.0\end{tabular} & \multicolumn{1}{||c|}{\begin{tabular}[c]{@{}c@{}}0.11\\ $\pm$0.0\end{tabular}} & \begin{tabular}[c]{@{}c@{}}0.96\\ $\pm$0.02\end{tabular} & \multicolumn{1}{||c|}{\begin{tabular}[c]{@{}c@{}}0.11\\ $\pm$0.0\end{tabular}} & \begin{tabular}[c]{@{}c@{}}0.93\\ $\pm$0.06\end{tabular} & \multicolumn{1}{||c|}{\begin{tabular}[c]{@{}c@{}}0.12\\ $\pm$0.0\end{tabular}} & \begin{tabular}[c]{@{}c@{}}0.99\\ $\pm$0.0\end{tabular} &  \multicolumn{1}{|c|}{$\uparrow$} & \multicolumn{1}{c|}{$\uparrow$} \\ \hline
\multicolumn{1}{|c|}{\begin{tabular}[c]{@{}c@{}}Joint \\ Learner\end{tabular}} & \multicolumn{1}{c|}{\begin{tabular}[c]{@{}c@{}}0.98\\ $\pm$0.0\end{tabular}} & \begin{tabular}[c]{@{}c@{}} N/A \end{tabular} & \multicolumn{1}{||c|}{\begin{tabular}[c]{@{}c@{}}0.87\\ $\pm$0.01\end{tabular}} & \begin{tabular}[c]{@{}c@{}} N/A \end{tabular} & \multicolumn{1}{||c|}{\begin{tabular}[c]{@{}c@{}}0.78\\ $\pm$0.01\end{tabular}} & \begin{tabular}[c]{@{}c@{}} N/A \end{tabular} & \multicolumn{1}{||c|}{\begin{tabular}[c]{@{}c@{}}0.58\\ $\pm$0.05\end{tabular}} & \begin{tabular}[c]{@{}c@{}} N/A \end{tabular} & \multicolumn{1}{||c|}{\begin{tabular}[c]{@{}c@{}}0.81\\ $\pm$0.02\end{tabular}} & \begin{tabular}[c]{@{}c@{}} N/A \end{tabular} &  \multicolumn{1}{|c|}{ } & \multicolumn{1}{c|}{N/A} \\ \hline
\multicolumn{1}{|c|}{\begin{tabular}[c]{@{}c@{}}Cumulative \\ Learner\end{tabular}} & \multicolumn{1}{c|}{\begin{tabular}[c]{@{}c@{}}0.97\\ $\pm$0.0\end{tabular}} & \begin{tabular}[c]{@{}c@{}}0.01\\ $\pm$0.0\end{tabular} & \multicolumn{1}{||c|}{\begin{tabular}[c]{@{}c@{}}0.87\\ $\pm$0.0\end{tabular}} & \begin{tabular}[c]{@{}c@{}}0.04\\ $\pm$0.02\end{tabular} & \multicolumn{1}{||c|}{\begin{tabular}[c]{@{}c@{}}0.71\\ $\pm$0.04\end{tabular}} & \begin{tabular}[c]{@{}c@{}}0.06\\ $\pm$0.03\end{tabular} & \multicolumn{1}{||c|}{\begin{tabular}[c]{@{}c@{}}0.8\\ $\pm$0.05\end{tabular}} & \begin{tabular}[c]{@{}c@{}}-0.03\\ $\pm$0.06\end{tabular} & \multicolumn{1}{||c|}{\begin{tabular}[c]{@{}c@{}}0.49\\ $\pm$0.03\end{tabular}} & \begin{tabular}[c]{@{}c@{}}0.22\\ $\pm$0.02\end{tabular} &  \multicolumn{1}{|c|}{ } & \multicolumn{1}{c|}{ } \\ \hline
\multicolumn{1}{|c|}{\begin{tabular}[c]{@{}c@{}}Random \\ Replay\end{tabular}} & \multicolumn{1}{c|}{\begin{tabular}[c]{@{}c@{}}0.16\\ $\pm$0.02\end{tabular}} & \begin{tabular}[c]{@{}c@{}}0.88\\ $\pm$0.03\end{tabular} & \multicolumn{1}{||c|}{\begin{tabular}[c]{@{}c@{}}0.17\\ $\pm$0.03\end{tabular}} & \begin{tabular}[c]{@{}c@{}}0.79\\ $\pm$0.03\end{tabular} & \multicolumn{1}{||c|}{\begin{tabular}[c]{@{}c@{}}0.12\\ $\pm$0.01\end{tabular}} & \begin{tabular}[c]{@{}c@{}}0.87\\ $\pm$0.03\end{tabular} & \multicolumn{1}{||c|}{\begin{tabular}[c]{@{}c@{}}0.36\\ $\pm$0.04\end{tabular}} & \begin{tabular}[c]{@{}c@{}}0.7\\ $\pm$0.07\end{tabular} & \multicolumn{1}{||c|}{\begin{tabular}[c]{@{}c@{}}0.45\\ $\pm$0.04\end{tabular}} & \begin{tabular}[c]{@{}c@{}}0.29\\ $\pm$0.08\end{tabular} &  \multicolumn{1}{|c|}{$\uparrow$} & \multicolumn{1}{c|}{$\uparrow$} \\ \hline
\multicolumn{1}{|c|}{\begin{tabular}[c]{@{}c@{}}Random \\ Replay \\ w/EWC\end{tabular}} & \multicolumn{1}{c|}{\begin{tabular}[c]{@{}c@{}}0.3\\ $\pm$0.02\end{tabular}} & \begin{tabular}[c]{@{}c@{}}0.63\\ $\pm$0.02\end{tabular} & \multicolumn{1}{||c|}{\begin{tabular}[c]{@{}c@{}}0.27\\ $\pm$0.02\end{tabular}} & \begin{tabular}[c]{@{}c@{}}0.63\\ $\pm$0.04\end{tabular} & \multicolumn{1}{||c|}{\begin{tabular}[c]{@{}c@{}}0.13\\ $\pm$0.01\end{tabular}} & \begin{tabular}[c]{@{}c@{}}0.85\\ $\pm$0.05\end{tabular} & \multicolumn{1}{||c|}{\begin{tabular}[c]{@{}c@{}}0.18\\ $\pm$0.09\end{tabular}} & \begin{tabular}[c]{@{}c@{}}0.64\\ $\pm$0.02\end{tabular} & \multicolumn{1}{||c|}{\begin{tabular}[c]{@{}c@{}}0.24\\ $\pm$0.07\end{tabular}} & \begin{tabular}[c]{@{}c@{}}0.68\\ $\pm$0.04\end{tabular} &  \multicolumn{1}{|c|}{$\uparrow$} & \multicolumn{1}{c|}{$\uparrow$} \\ \hline
\multicolumn{1}{|c|}{\begin{tabular}[c]{@{}c@{}}Dark ER\end{tabular}} & \multicolumn{1}{c|}{\begin{tabular}[c]{@{}c@{}}0.44\\ $\pm$0.02\end{tabular}} & \begin{tabular}[c]{@{}c@{}}0.46\\ $\pm$0.02\end{tabular} & \multicolumn{1}{||c|}{\begin{tabular}[c]{@{}c@{}}0.29\\ $\pm$0.02\end{tabular}} & \begin{tabular}[c]{@{}c@{}}0.67\\ $\pm$0.06\end{tabular} & \multicolumn{1}{||c|}{\begin{tabular}[c]{@{}c@{}}0.17\\ $\pm$0.02\end{tabular}} & \begin{tabular}[c]{@{}c@{}}0.78\\ $\pm$0.04\end{tabular} & \multicolumn{1}{||c|}{\begin{tabular}[c]{@{}c@{}}0.38\\ $\pm$0.01\end{tabular}} & \begin{tabular}[c]{@{}c@{}}0.64\\ $\pm$0.08\end{tabular} & \multicolumn{1}{||c|}{\begin{tabular}[c]{@{}c@{}}0.32\\ $\pm$0.07\end{tabular}} & \begin{tabular}[c]{@{}c@{}}0.61\\ $\pm$0.07\end{tabular} &  \multicolumn{1}{|c|}{$\uparrow$} & \multicolumn{1}{c|}{$\uparrow$} \\ \hline
\multicolumn{1}{|c|}{\begin{tabular}[c]{@{}c@{}}iCaRL\end{tabular}} & \multicolumn{1}{c|}{\begin{tabular}[c]{@{}c@{}}0.3\\ $\pm$0.15\end{tabular}} & \begin{tabular}[c]{@{}c@{}}0.41\\ $\pm$0.04\end{tabular} & \multicolumn{1}{||c|}{\begin{tabular}[c]{@{}c@{}}0.27\\ $\pm$0.06\end{tabular}} & \begin{tabular}[c]{@{}c@{}}0.56\\ $\pm$0.15\end{tabular} & \multicolumn{1}{||c|}{\begin{tabular}[c]{@{}c@{}}0.25\\ $\pm$0.02\end{tabular}} & \begin{tabular}[c]{@{}c@{}}0.44\\ $\pm$0.03\end{tabular} & \multicolumn{1}{||c|}{\begin{tabular}[c]{@{}c@{}}0.25\\ $\pm$0.02\end{tabular}} & \begin{tabular}[c]{@{}c@{}}0.59\\ $\pm$0.08\end{tabular} & \multicolumn{1}{||c|}{\begin{tabular}[c]{@{}c@{}}0.19\\ $\pm$0.03\end{tabular}} & \begin{tabular}[c]{@{}c@{}}0.76\\ $\pm$0.05\end{tabular} &  \multicolumn{1}{|c|}{$\uparrow$} & \multicolumn{1}{c|}{$\uparrow$} \\ \hline
\multicolumn{1}{|c|}{\begin{tabular}[c]{@{}c@{}}Continual \\ CRUST\end{tabular}} & \multicolumn{1}{c|}{\begin{tabular}[c]{@{}c@{}}0.86\\ $\pm$0.08\end{tabular}} & \begin{tabular}[c]{@{}c@{}}0.03\\ $\pm$0.03\end{tabular} & \multicolumn{1}{||c|}{\begin{tabular}[c]{@{}c@{}}\textbf{0.79}\\ $\pm$\textbf{0.02}\end{tabular}} & \begin{tabular}[c]{@{}c@{}}\textbf{0.04}\\ $\pm$\textbf{0.01}\end{tabular} & \multicolumn{1}{||c|}{\begin{tabular}[c]{@{}c@{}}0.36\\ $\pm$0.02\end{tabular}} & \begin{tabular}[c]{@{}c@{}}\textbf{0.21}\\ $\pm$\textbf{0.03}\end{tabular} & \multicolumn{1}{||c|}{\begin{tabular}[c]{@{}c@{}}0.71\\ $\pm$0.02\end{tabular}} & \begin{tabular}[c]{@{}c@{}}0.05\\ $\pm$0.05\end{tabular} & \multicolumn{1}{||c|}{\begin{tabular}[c]{@{}c@{}}\textbf{0.65}\\ $\pm$\textbf{0.08}\end{tabular}} & \begin{tabular}[c]{@{}c@{}}0.16\\ $\pm$0.05\end{tabular} &  \multicolumn{1}{|c|}{ } & \multicolumn{1}{c|}{ } \\ \hline
\multicolumn{1}{|c|}{\begin{tabular}[c]{@{}c@{}}Continual \\ CosineCRUST\end{tabular}} & \multicolumn{1}{c|}{\begin{tabular}[c]{@{}c@{}}\textbf{0.9}\\ $\pm$\textbf{0.02}\end{tabular}} & \begin{tabular}[c]{@{}c@{}}\textbf{0.02}\\ $\pm$\textbf{0.02}\end{tabular} & \multicolumn{1}{||c|}{\begin{tabular}[c]{@{}c@{}}0.73\\ $\pm$0.02\end{tabular}} & \begin{tabular}[c]{@{}c@{}}0.12\\ $\pm$0.03\end{tabular} & \multicolumn{1}{||c|}{\begin{tabular}[c]{@{}c@{}}\textbf{0.4}\\ $\pm$\textbf{0.01}\end{tabular}} & \begin{tabular}[c]{@{}c@{}}0.22\\ $\pm$0.04\end{tabular} & \multicolumn{1}{||c|}{\begin{tabular}[c]{@{}c@{}}\textbf{0.73}\\ $\pm$\textbf{0.02}\end{tabular}} & \begin{tabular}[c]{@{}c@{}}\textbf{0.0}\\ $\pm$\textbf{0.05}\end{tabular} & \multicolumn{1}{||c|}{\begin{tabular}[c]{@{}c@{}}\textbf{0.65}\\ $\pm$\textbf{0.03}\end{tabular}} & \begin{tabular}[c]{@{}c@{}}\textbf{0.12}\\ $\pm$\textbf{0.03}\end{tabular} &  \multicolumn{1}{|c|}{ } & \multicolumn{1}{c|}{ } \\ \hline
\end{tabular}}
\caption{Comparing the final accuracy and forgetting metrics across five datasets under \textbf{high label noise} (flipping probability=0.5) averaged over five random seeds. We also perform a Friedman test with posthoc Conover tests with the Holm step down procedure with a p-value of 0.05 to determine whether Continual CosineCRUST statistically significantly outperforms ($\uparrow$) or underperforms ($\downarrow$) the other approaches on average in terms of final accuracy and forgetting. Additional results appear in the supplemental materials.}
\label{tab:main_results}
\end{table*}

\subsection{Experimental Procedure}
We follow the standard protocol used in prior work \cite{new2022lifelong} for evaluating CIL with alternating training and evaluation experiences. The first experience always consists of two classes, and every subsequent experience consists of adding one class. During each training experience, the model is shown only data labeled with new classes, and has no external access to past experiences. 
In the evaluation phase, the classifier is applied to a held-out test dataset for each past, current, and future class. We use the Avalanche continual learning library \cite{lomonaco2021avalanche,carta2023avalanche} to manage experiments and for baseline comparison algorithms. 
Hyperparameters for our experiments are reported in the supplemental materials. We run five seeds per experiment. In general, Coreset sizes range from 60 to 300 samples per class.

\begin{figure*}[t]
  \centering
   \includegraphics[width=0.32\linewidth]{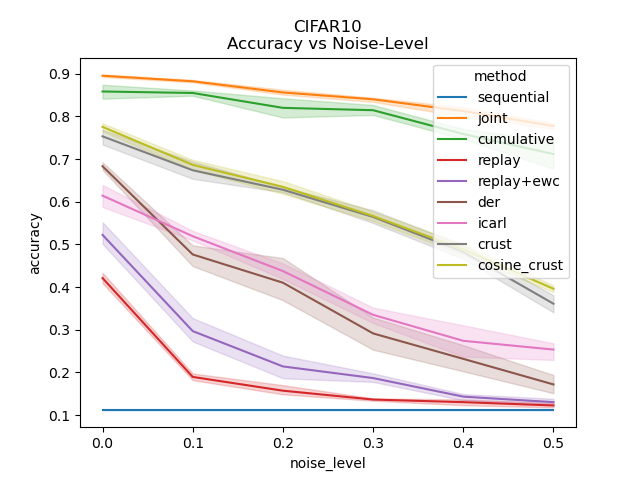}
   \includegraphics[width=0.32\linewidth]{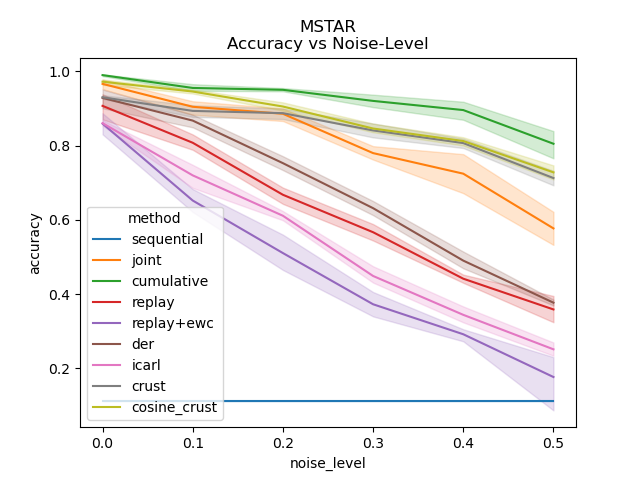}
   \includegraphics[width=0.32\linewidth]{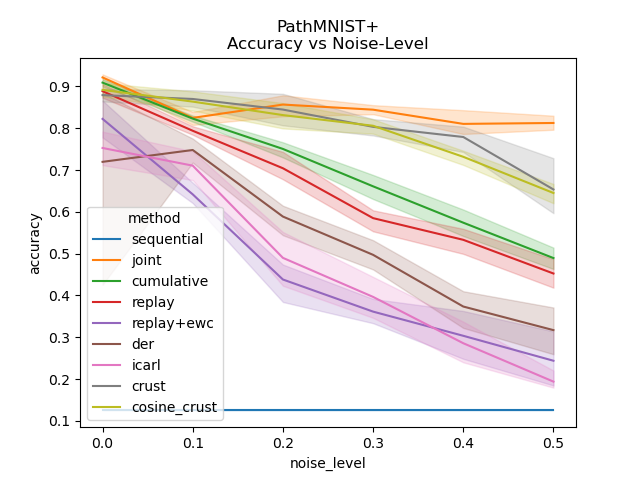}
   \caption{Looking at the final accuracy at different label flipping noise levels (0.0-0.5) for five benchmark datasets for different strategies; additional results and similar plots for the forgetting metric are included in the supplemental materials.}
   \label{fig:all_noise}
\end{figure*}

\subsection{Metrics}
We consider three metrics. Let $R_i,j$ be the test accuracy on class $j$ after training on class $i$. 
\begin{itemize}
\item \textbf{Average Final Accuracy (Acc)} \cite{diaz2018don,lopez2017gradient}: Let $T$ be the final experience. After training on all experiences, $afa = \frac{1}{T}\sum_{j=1..T} R_{T,j}$.
\item \textbf{Forgetting} \cite{lopez2017gradient}: Forgetting measures the loss of performance on previous experiences after learning new experiences. We define $forg = \frac{1}{|C|}\sum_{i=j+1..T}\sum_{j=1..T}R{j,j} - R_{i,j}$ where $|C|$ is the number of pairwise comparisons in the double summation. If $forg > 0$, the model has lost performance.
\item \textbf{Coreset purity}: Purity represents the proportion of clean samples selected per class Coreset, averaged over all classes, where $purity=1$ implies noiseless Coresets. 
\end{itemize}
\subsection{Compared Algorithms}
\textbf{Baselines}: \textbf{Replay} uniformly randomly selects samples at the end of each learning experience. It adds these samples to the replay buffer, which is used for data augmentation; \textbf{Naive}: sequentially trains on each experience without any machinery to mitigate forgetting and serves as a lower bound on performance; \textbf{Cumulative}: trains progressively, incorporating new data by including \textit{all} past data and serving as an approximate upper bound; and \textbf{Joint} training: provides an approximate upper bound by learning all tasks simultaneously aka multi-task learning.

\textbf{Prior Work}: \textbf{iCaRL} \cite{rebuffi2017icarl} combines classification and representation learning, using clustering-based exemplars to preserve past information; \textbf{Dark Experience Replay (Dark ER)} \cite{buzzega2020dark} leverages past model states to generate pseudo-examples, enhancing knowledge retention without storing raw data; \textbf{Replay+Elastic Weight Consolidation (EWC)} employs EWC regularization \cite{kirkpatrick2017overcoming} on top of replay. These are all well-tested and well-adopted memory-based strategies used in the CL community. Thus, we conduct comprehensive comparisons against a representative set of memory-based algorithms for CIL. 

\subsection{Effect of Label Noise}
Our first experiment focuses on the effect of mislabeling during data collection/training on CL performance. Labels in the training set are uniformly selected and changed to a different randomly selected label with equal probability. In practice, this can occur naturally when collecting a batch of data and labeling the entire batch based on the majority label. In this experiment, there may be mislabeled samples from both past and future experiences. If a continual learning agent could learn despite the label noise, it would significantly reduce the cost of labeling data. We study the effect of label noise by varying the label flipping probabilities in the range 0.0 to 0.5 (in steps of 0.1).

We show performance degradation profiles as a function of label noise for different strategies in Fig. \ref{fig:all_noise}.
Across all datasets, we see that the prior work on CL work well in the zero-noise setting, but fail to learn when there are moderate (noise$\geq$0.2) to high levels of label noise, showing rapid degradation in final accuracy (e.g.,\! iCaRL drops from 75\% at 0.0 noise to 19\% at 0.5 noise for the PathMNIST+ dataset) and increased forgetting. So learning under label noise can exacerbate forgetting in the CIL setting, potentially due to overfitting to noisy samples. In contrast, Continual CRUST and Continual CosineCRUST are significantly more robust (e.g.,\! 89\% at 0.0 noise down to 65\% at 0.5 noise for the PathMNIST+ dataset) and show a more graceful degradation in accuracy with reduced forgetting as data collection becomes noisier.

We show results over all datasets in Table \ref{tab:main_results} for flipping probability of 0.5 (highest noise). We report the Acc and Forg. metrics. Further, for meta-analysis across datasets, we also perform the Friedman test with posthoc Conover tests with the Holm step down procedure with a p-value of 0.05 to determine whether Continual CosineCRUST statistically significantly outperforms (↑) or underperforms (↓) the other approaches. For example, compared to iCARL, Continual CosineCRUST has statistically significant higher accuracy and lower forgetting over all datasets.
Results for all settings are reported in the supplemental. \textbf{In all datasets considered, our proposed methods have statistically significant best performance compared to prior work}. However, there is still a large gap with the upper bound multi-task learner in CIFAR10, so there is still room to improve. Interestingly, our method outperforms the joint learner in the MSTAR dataset (73\% for Continual CosineCrust vs 58\% for the joint learner).
\begin{table}[h!]
\centering
\resizebox{0.49\textwidth}{!}{
\begin{tabular}{c|cc|cc|cc|cc|}
\cline{2-9}
\textbf{} & \multicolumn{2}{c|}{\textbf{Noise=0.2}} & \multicolumn{2}{||c|}{\textbf{Noise=0.4}} & \multicolumn{2}{||c|}{\textbf{Noise=0.6}} & \multicolumn{2}{||c|}{\textbf{Noise=0.8}} \\ \hline
\multicolumn{1}{|c|}{\textbf{Algorithm}} & \multicolumn{1}{c|}{\textit{\textbf{Acc}}} & \textit{\textbf{Forg.}} & \multicolumn{1}{||c|}{\textit{\textbf{Acc}}} & \textit{\textbf{Forg.}} & \multicolumn{1}{||c|}{\textit{\textbf{Acc}}} & \textit{\textbf{Forg.}} & \multicolumn{1}{||c|}{\textit{\textbf{Acc}}} & \textit{\textbf{Forg.}} \\ \hline
\multicolumn{1}{|c|}{\begin{tabular}[c]{@{}c@{}}Naive \\ Sequential \\ Learner\end{tabular}} & \multicolumn{1}{c|}{\begin{tabular}[c]{@{}c@{}}0.11\\ $\pm$0.0\end{tabular}} & \begin{tabular}[c]{@{}c@{}}1.0\\ $\pm$0.0\end{tabular} & \multicolumn{1}{||c|}{\begin{tabular}[c]{@{}c@{}}0.11\\ $\pm$0.0\end{tabular}} & \begin{tabular}[c]{@{}c@{}}1.0\\ $\pm$0.0\end{tabular} & \multicolumn{1}{||c|}{\begin{tabular}[c]{@{}c@{}}0.11\\ $\pm$0.0\end{tabular}} & \begin{tabular}[c]{@{}c@{}}1.0\\ $\pm$0.0\end{tabular} & \multicolumn{1}{||c|}{\begin{tabular}[c]{@{}c@{}}0.11\\ $\pm$0.0\end{tabular}} & \begin{tabular}[c]{@{}c@{}}1.0\\ $\pm$0.0\end{tabular} \\ \hline
\multicolumn{1}{|c|}{\begin{tabular}[c]{@{}c@{}}Joint \\ Learner\end{tabular}} & \multicolumn{1}{c|}{\begin{tabular}[c]{@{}c@{}}0.93\\ $\pm$0.0\end{tabular}} & \begin{tabular}[c]{@{}c@{}} N/A \end{tabular} & \multicolumn{1}{||c|}{\begin{tabular}[c]{@{}c@{}}0.93\\ $\pm$0.0\end{tabular}} & \begin{tabular}[c]{@{}c@{}} N/A \end{tabular} & \multicolumn{1}{||c|}{\begin{tabular}[c]{@{}c@{}}0.93\\ $\pm$0.0\end{tabular}} & \begin{tabular}[c]{@{}c@{}} N/A \end{tabular} & \multicolumn{1}{||c|}{\begin{tabular}[c]{@{}c@{}}0.92\\ $\pm$0.0\end{tabular}} & \begin{tabular}[c]{@{}c@{}} N/A \end{tabular} \\ \hline
\multicolumn{1}{|c|}{\begin{tabular}[c]{@{}c@{}}Cumulative \\ Learner\end{tabular}} & \multicolumn{1}{c|}{\begin{tabular}[c]{@{}c@{}}0.93\\ $\pm$0.0\end{tabular}} & \begin{tabular}[c]{@{}c@{}}0.02\\ $\pm$0.01\end{tabular} & \multicolumn{1}{||c|}{\begin{tabular}[c]{@{}c@{}}0.92\\ $\pm$0.01\end{tabular}} & \begin{tabular}[c]{@{}c@{}}0.03\\ $\pm$0.01\end{tabular} & \multicolumn{1}{||c|}{\begin{tabular}[c]{@{}c@{}}0.92\\ $\pm$0.01\end{tabular}} & \begin{tabular}[c]{@{}c@{}}0.02\\ $\pm$0.01\end{tabular} & \multicolumn{1}{||c|}{\begin{tabular}[c]{@{}c@{}}0.91\\ $\pm$0.01\end{tabular}} & \begin{tabular}[c]{@{}c@{}}0.03\\ $\pm$0.01\end{tabular} \\ \hline
\multicolumn{1}{|c|}{\begin{tabular}[c]{@{}c@{}}Random \\ Replay\end{tabular}} & \multicolumn{1}{c|}{\begin{tabular}[c]{@{}c@{}}0.74\\ $\pm$0.05\end{tabular}} & \begin{tabular}[c]{@{}c@{}}0.23\\ $\pm$0.03\end{tabular} & \multicolumn{1}{||c|}{\begin{tabular}[c]{@{}c@{}}0.75\\ $\pm$0.02\end{tabular}} & \begin{tabular}[c]{@{}c@{}}0.25\\ $\pm$0.02\end{tabular} & \multicolumn{1}{||c|}{\begin{tabular}[c]{@{}c@{}}0.72\\ $\pm$0.03\end{tabular}} & \begin{tabular}[c]{@{}c@{}}0.26\\ $\pm$0.02\end{tabular} & \multicolumn{1}{||c|}{\begin{tabular}[c]{@{}c@{}}0.72\\ $\pm$0.03\end{tabular}} & \begin{tabular}[c]{@{}c@{}}0.28\\ $\pm$0.05\end{tabular} \\ \hline
\multicolumn{1}{|c|}{\begin{tabular}[c]{@{}c@{}}Random \\ Replay \\ w/EWC\end{tabular}} & \multicolumn{1}{c|}{\begin{tabular}[c]{@{}c@{}}0.74\\ $\pm$0.02\end{tabular}} & \begin{tabular}[c]{@{}c@{}}0.28\\ $\pm$0.04\end{tabular} & \multicolumn{1}{||c|}{\begin{tabular}[c]{@{}c@{}}0.72\\ $\pm$0.03\end{tabular}} & \begin{tabular}[c]{@{}c@{}}0.29\\ $\pm$0.02\end{tabular} & \multicolumn{1}{||c|}{\begin{tabular}[c]{@{}c@{}}0.69\\ $\pm$0.02\end{tabular}} & \begin{tabular}[c]{@{}c@{}}0.29\\ $\pm$0.02\end{tabular} & \multicolumn{1}{||c|}{\begin{tabular}[c]{@{}c@{}}0.66\\ $\pm$0.02\end{tabular}} & \begin{tabular}[c]{@{}c@{}}0.31\\ $\pm$0.02\end{tabular} \\ \hline
\multicolumn{1}{|c|}{\begin{tabular}[c]{@{}c@{}}Dark ER\end{tabular}} & \multicolumn{1}{c|}{\begin{tabular}[c]{@{}c@{}}0.83\\ $\pm$0.02\end{tabular}} & \begin{tabular}[c]{@{}c@{}}0.16\\ $\pm$0.02\end{tabular} & \multicolumn{1}{||c|}{\begin{tabular}[c]{@{}c@{}}0.8\\ $\pm$0.02\end{tabular}} & \begin{tabular}[c]{@{}c@{}}0.2\\ $\pm$0.03\end{tabular} & \multicolumn{1}{||c|}{\begin{tabular}[c]{@{}c@{}}0.81\\ $\pm$0.01\end{tabular}} & \begin{tabular}[c]{@{}c@{}}0.2\\ $\pm$0.02\end{tabular} & \multicolumn{1}{||c|}{\begin{tabular}[c]{@{}c@{}}0.78\\ $\pm$0.02\end{tabular}} & \begin{tabular}[c]{@{}c@{}}0.2\\ $\pm$0.03\end{tabular} \\ \hline
\multicolumn{1}{|c|}{\begin{tabular}[c]{@{}c@{}}iCaRL\end{tabular}} & \multicolumn{1}{c|}{\begin{tabular}[c]{@{}c@{}}0.55\\ $\pm$0.02\end{tabular}} & \begin{tabular}[c]{@{}c@{}}0.37\\ $\pm$0.03\end{tabular} & \multicolumn{1}{||c|}{\begin{tabular}[c]{@{}c@{}}0.5\\ $\pm$0.02\end{tabular}} & \begin{tabular}[c]{@{}c@{}}0.4\\ $\pm$0.05\end{tabular} & \multicolumn{1}{||c|}{\begin{tabular}[c]{@{}c@{}}0.49\\ $\pm$0.01\end{tabular}} & \begin{tabular}[c]{@{}c@{}}0.42\\ $\pm$0.05\end{tabular} & \multicolumn{1}{||c|}{\begin{tabular}[c]{@{}c@{}}0.48\\ $\pm$0.02\end{tabular}} & \begin{tabular}[c]{@{}c@{}}0.39\\ $\pm$0.05\end{tabular} \\ \hline
\multicolumn{1}{|c|}{\begin{tabular}[c]{@{}c@{}}Continual \\ CRUST\end{tabular}} & \multicolumn{1}{c|}{\begin{tabular}[c]{@{}c@{}}0.83\\ $\pm$0.0\end{tabular}} & \begin{tabular}[c]{@{}c@{}}0.07\\ $\pm$0.02\end{tabular} & \multicolumn{1}{||c|}{\begin{tabular}[c]{@{}c@{}}0.83\\ $\pm$0.0\end{tabular}} & \begin{tabular}[c]{@{}c@{}}0.08\\ $\pm$0.01\end{tabular} & \multicolumn{1}{||c|}{\begin{tabular}[c]{@{}c@{}}0.82\\ $\pm$0.01\end{tabular}} & \begin{tabular}[c]{@{}c@{}}0.08\\ $\pm$0.02\end{tabular} & \multicolumn{1}{||c|}{\begin{tabular}[c]{@{}c@{}}0.81\\ $\pm$0.01\end{tabular}} & \begin{tabular}[c]{@{}c@{}}0.1\\ $\pm$0.03\end{tabular} \\ \hline
\multicolumn{1}{|c|}{\begin{tabular}[c]{@{}c@{}}Continual \\ CosineCRUST\end{tabular}} & \multicolumn{1}{c|}{\begin{tabular}[c]{@{}c@{}}0.85\\ $\pm$0.0\end{tabular}} & \begin{tabular}[c]{@{}c@{}}0.04\\ $\pm$0.01\end{tabular} & \multicolumn{1}{||c|}{\begin{tabular}[c]{@{}c@{}}0.85\\ $\pm$0.01\end{tabular}} & \begin{tabular}[c]{@{}c@{}}0.04\\ $\pm$0.01\end{tabular} & \multicolumn{1}{||c|}{\begin{tabular}[c]{@{}c@{}}0.84\\ $\pm$0.0\end{tabular}} & \begin{tabular}[c]{@{}c@{}}0.06\\ $\pm$0.02\end{tabular} & \multicolumn{1}{||c|}{\begin{tabular}[c]{@{}c@{}}0.82\\ $\pm$0.01\end{tabular}} & \begin{tabular}[c]{@{}c@{}}0.08\\ $\pm$0.01\end{tabular} \\ \hline
\end{tabular} }
\caption{Comparing the final accuracy and forgetting metrics 
with varying Salt-and-Pepper instance noise for FashionMNIST with Coreset size=128. Results for other datasets in the supplemental.}
\label{tab:data_noise_fmnist}
\end{table}

\subsection{Effect of Instance Noise}
We study the effect on accuracy vs the number of noisy samples at a given amplitude of salt-and-pepper noise. In practice, this could occur if data is being collected from a failing sensor, which may stochastically capture either clean or perturbed images. The perturbation adds a high level of salt-and-pepper noise (=90\% perturbed pixels) and vary the proportion of noisy samples in the training data in the range 0.2 to 0.8 (in steps of 0.2). Table \ref{tab:data_noise_fmnist} shows that accuracy and forgetting metrics vary with the fraction of noisy samples of FashionMNIST. 
The prior work on CL shows only a small degradations in performance with noise proportion. This is expected because Convolutional NNs tend to be less sensitive to salt-and-pepper noise than  label noise. \textbf{However, the forgetting metrics reveals that Continual CRUST and Continual CosineCRUST significantly outperform the prior work.} For example, our methods achieve a forgetting metric less than 0.1 at all noise levels, compared to the roughly average 0.2 achieved by DER. The method iCARL seems to be much more sensitive to salt-and-pepper noise, with almost 4-5x the forgetting metric.
Additional experiments are reported in the supplementary materials.

\begin{figure*}[h!]
\centering
\includegraphics[width=0.69\linewidth]{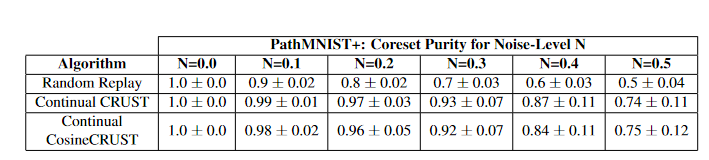}
\includegraphics[width=0.3\linewidth]{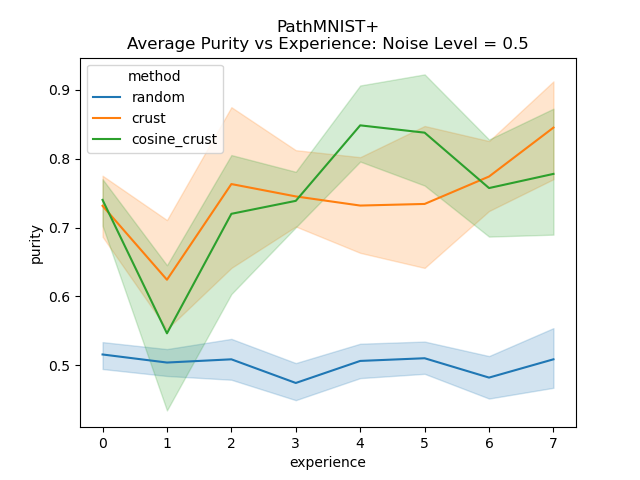}
\caption{Left: Average Coreset purity over all experiences on PathMNIST+ dataset for different levels of label noise. Right: Examining purity of selected Coresets for each learning experience for PathMNIST+ for label-flipping noise level of 0.5}
\label{fig:purity_pathmnist}
\end{figure*}

\begin{table*}[h!]
\centering
\resizebox{0.65\textwidth}{!}{
\begin{tabular}{c|cccccc|}
\cline{2-7}
\multicolumn{1}{l|}{} & \multicolumn{6}{c|}{\textbf{FashionMNIST: Coreset Purity for Salt and Pepper at Noise-Level N}} \\ \hline
\multicolumn{1}{|c|}{\textbf{Algorithm}} & \multicolumn{1}{c|}{\textbf{N=0.0}} & \multicolumn{1}{c|}{\textbf{N=0.1}} & \multicolumn{1}{c|}{\textbf{N=0.2}} & \multicolumn{1}{c|}{\textbf{N=0.3}} & \multicolumn{1}{c|}{\textbf{N=0.4}} & \textbf{N=0.5} \\ \hline
\multicolumn{1}{|c|}{Random Replay} & \multicolumn{1}{c|}{1.0 $\pm$ 0.0} & \multicolumn{1}{c|}{0.89 $\pm$ 0.02} & \multicolumn{1}{c|}{0.79 $\pm$ 0.03} & \multicolumn{1}{c|}{0.69 $\pm$ 0.04} & \multicolumn{1}{c|}{0.59 $\pm$ 0.04} & \multicolumn{1}{c|}{0.49 $\pm$ 0.04} \\ \hline
\multicolumn{1}{|c|}{Continual CRUST} & \multicolumn{1}{c|}{1.0 $\pm$ 0.0} & \multicolumn{1}{c|}{0.94 $\pm$ 0.05} & \multicolumn{1}{c|}{0.84 $\pm$ 0.08} & \multicolumn{1}{c|}{0.74 $\pm$ 0.07} & \multicolumn{1}{c|}{0.6 $\pm$ 0.08} & \multicolumn{1}{c|}{0.51 $\pm$ 0.08} \\ \hline
\multicolumn{1}{|c|}{Continual CosineCRUST} & \multicolumn{1}{c|}{1.0 $\pm$ 0.0} & \multicolumn{1}{c|}{0.93 $\pm$ 0.03} & \multicolumn{1}{c|}{0.83 $\pm$ 0.04} & \multicolumn{1}{c|}{0.74 $\pm$ 0.05} & \multicolumn{1}{c|}{0.61 $\pm$ 0.06} & \multicolumn{1}{c|}{0.52 $\pm$ 0.05} \\ \hline
\end{tabular}}
\caption{Coreset purity for FashionMNIST with perturbed data for salt and pepper noise (top) and samples consisting of uniform random noise (complete sample corruption) (bottom). }
\label{tab:data_purity_fashionmnist}
\end{table*}


\subsection{Purity of Coresets}
\label{sec:purity_main}
We evaluate Coreset purity to check if Continual CRUST and Continual CosineCRUST are selecting significantly fewer noisy samples than random selection. In Fig. \ref{fig:purity_pathmnist}, we show Coreset purity vs amount of label noise on the PathMNIST+ dataset.
we want to understand how purity of Coresets is affected by amount of label noise; 
and we want to understand how the behavior of the Coreset selection changes as the model encounters new tasks. 

We see that as label noise increases, unsurprisingly, purity decreases; however, \textbf{Continual CRUST and Continual CosineCRUST maintain more than 90\% purity until noise level 0.3 and 75\% purity at noise level 0.5}. The requirement from Theorem \ref{thm:crust_convergence} is satisfied at 90\% purity i.e.\! the Coreset contains $\rho=0.1$ and label margin $\delta \geq 1 \geq 8\rho$ in these datasets. 
In Fig. \ref{fig:purity_pathmnist}, we see interesting behavior w.r.t. Coreset purity over time or learning experiences. Coreset purity actually tends to increase in this experiment as the model encounters additional classes and seems to develop a better understanding of the total data space.

We also look at Coreset purity under instance noise on FashionMNIST in Table \ref{tab:data_purity_fashionmnist}. When data is perturbed with salt-and pepper-noise, we see robustness till 30\% of the data is noisy, 
but as the clean-to-noisy sample ratio approaches 0.5, Continual CRUST and Continual CosineCRUST select Coresets with purity matching the noise level. 
We hypothesize 
that unlike in the label-flapping scenario, the "noisy" instances are still informative for the classification task. Thus, it may be useful to keep them in the Coreset. Additional studies are included in the supplemental materials exploring purity under more extreme noise settings.

\section{Related Work}
\label{sec:related}

\subsection{Overview of Continual and Lifelong Learning}
\label{sec:lifelong} 

In CL \cite{de2021continual,van2022three}, an agent is trained on a sequence of tasks, and the agent must balance plasticity vs stability. There are three main scenarios in CL \cite{van2022three}: task- (TIL), domain- (DIL), and class-incremental learning (CIL). In TIL, an agent solves sequences of tasks with explicit knowledge of the identity of the current task-of-interest. In DIL, the current task-of-interest is not known by the agent, but the structure of the problem does not change between tasks as input distributions shift. In CIL, new class sets are added over time. This work focuses on the CIL setting.

There are three general classes of strategies for CL \cite{chen2018lifelong} in the literature. Memory-based methods (e.g., \cite{rolnick2019experience,isele2018selective,lopez2017gradient,buzzega2020dark,rebuffi2017icarl,aljundi2019gradient}) save data samples (or derived representations) from previous tasks and periodically ``replay'' the samples to minimize forgetting. Regularization-based methods (e.g., \cite{kirkpatrick2017overcoming,zenke2017continual,aljundi2018memory}) constrain the learning behavior of neural networks to prevent overfitting to the new task. Architecture-based methods (e.g. \cite{rusu2016progressive,fayek2020progressive,yoon2017lifelong,hung2019compacting,li2019learn}) adapt by intelligently growing and pruning the model architecture. This work focuses on understanding how memory-based approaches are effected by corrupted data collection.

\textbf{Memory-Based Approaches for Continual Learning}: 
Often, CL must run on hardware with a fixed small size memory available. 
Prioritized replay \cite{schaul2015prioritized,isele2018selective,wang2024prioritized} uses intelligent heuristics to construct the replay buffer. Because Coresets are effective at summarizing large datasets with few data points, they have been used to construct replay buffers for continual learning \cite{borsos2020coresets,yoononline,tiwari2022gcr}. Using gradient information to construct the replay buffer is another popular approach \cite{lopez2017gradient,aljundi2019gradient}. Other methods derive exemplars from the training data (e.g., using class means in iCARL \cite{rebuffi2017icarl}) to construct representative buffers. Intermediate representations (latent replay \cite{pellegrini2020latent}) or compressed representations \cite{hayes2020remind} can also be saved in place of real data samples. Generative modeling (``generative replay'' \cite{shin2017continual}) has been used to try to improve the diversity of replay buffers. This work proposes a noise-tolerant gradient-based Coreset selection strategy for generating the replay buffer, extending CRUST \cite{mirzasoleiman2020coresets} from the static learning setting to the CL setting, and extending label noise to include instance noise. 

\textbf{Robust Continual Learning}
CL exposes a unique attack surface for introducing corrupted data during training. Recent work has examined attacks for continual learning designed to induce forgetting, prevent the learning of new tasks, and inject backdoor triggers \cite{li2022targeted,li2023pacol,abbasi2024brainwash,guo2024persistent,kang2023poisoning,umer2021adversarial,kangcontinual}. Li et al. \cite{li2023pacol} showed that continual learners are susceptible to label flipping attacks. Approaches have been proposed to improve continual learner's robustness (generally robustness to adversarial attacks) \cite{khan2022adversarially,bai2023towards,wang2023metamix,umer2024adversary,jia2023robustness,ru2024maintaining}. Our approach considers more generic label and instance noise. Existing approaches generally rely on empirical validation as evidence of robustness. Our approach builds on theoretical foundations in addition rigorous empirical validation.
\section{Conclusions and Future Work}
\label{sec:conclusions}
\textbf{Summary of Findings}: 
We introduced Continual CRUST and Continual CosineCRUST: Coreset-based replay strategies that are theoretically-motivated to be noise-tolerant for both label and instance noise during continual learning. These approaches were rigorously validated on five diverse benchmark datasets. We demonstrated the robustness of our memory-based continual learner, whereas traditional strategies failed under moderate amounts of noise. 

\textbf{Weaknesses and Future Work}: 
Continual CosineCRUST generally exceeds the performance of CRUST in CIL. However, in the worst case scenario (high fraction of noisy samples), the clustering step of CosineCRUST may identify two dense clusters, one of clean samples and one of noisy samples. Performing submodular optimization on the noisy cluster will pollute the global Coreset. Theoretical bounds developed would not apply in this scenario. This requires further exploration and may lead to better methods with improved bounds. 
We observed that Continual(Cosine)CRUST works better for well-structured data and struggles with unstructured low-information data (e.g., CIFAR10). Fully understanding this issue requires additional exploration. 
We have not yet explored Continual (Cosine)CRUST's susceptibility to adversarial attacks (e.g., backdoor triggers), which purposely strengthen spurious patterns in the data. 
It should also be noted that CosineCRUST involves a clustering step, resulting in worse asymptotic complexity than CRUST.
Finally, we only examined noise-tolerant learning in the context of class-incremental learning. Further experimentation is needed to validate that the approach generalizes to continual learning beyond the CIL classification problem.

{
    \small
    \bibliographystyle{ieeenat_fullname}
    \bibliography{main}

\begin{thebibliography}{61}
\providecommand{\natexlab}[1]{#1}
\providecommand{\url}[1]{\texttt{#1}}
\expandafter\ifx\csname urlstyle\endcsname\relax
  \providecommand{\doi}[1]{doi: #1}\else
  \providecommand{\doi}{doi: \begingroup \urlstyle{rm}\Url}\fi

\bibitem[Abbasi et~al.(2024)Abbasi, Nooralinejad, Pirsiavash, and Kolouri]{abbasi2024brainwash}
Ali Abbasi, Parsa Nooralinejad, Hamed Pirsiavash, and Soheil Kolouri.
\newblock Brainwash: A poisoning attack to forget in continual learning.
\newblock In \emph{Proceedings of the IEEE/CVF Conference on Computer Vision and Pattern Recognition}, pages 24057--24067, 2024.

\bibitem[Aljundi et~al.(2018)Aljundi, Babiloni, Elhoseiny, Rohrbach, and Tuytelaars]{aljundi2018memory}
Rahaf Aljundi, Francesca Babiloni, Mohamed Elhoseiny, Marcus Rohrbach, and Tinne Tuytelaars.
\newblock Memory aware synapses: Learning what (not) to forget.
\newblock In \emph{Proceedings of the European conference on computer vision (ECCV)}, pages 139--154, 2018.

\bibitem[Aljundi et~al.(2019)Aljundi, Lin, Goujaud, and Bengio]{aljundi2019gradient}
Rahaf Aljundi, Min Lin, Baptiste Goujaud, and Yoshua Bengio.
\newblock Gradient based sample selection for online continual learning.
\newblock \emph{Advances in neural information processing systems}, 32, 2019.

\bibitem[Asadi et~al.(2022)Asadi, Mudur, and Belilovsky]{asadi2022tackling}
Nader Asadi, Sudhir Mudur, and Eugene Belilovsky.
\newblock Tackling online one-class incremental learning by removing negative contrasts.
\newblock \emph{arXiv preprint arXiv:2203.13307}, 2022.

\bibitem[Bai et~al.(2023)Bai, Chen, Lyu, Zhao, and Wen]{bai2023towards}
Tao Bai, Chen Chen, Lingjuan Lyu, Jun Zhao, and Bihan Wen.
\newblock Towards adversarially robust continual learning.
\newblock In \emph{ICASSP 2023-2023 IEEE International Conference on Acoustics, Speech and Signal Processing (ICASSP)}, pages 1--5. IEEE, 2023.

\bibitem[Borsos et~al.(2020)Borsos, Mutny, and Krause]{borsos2020coresets}
Zal{\'a}n Borsos, Mojmir Mutny, and Andreas Krause.
\newblock Coresets via bilevel optimization for continual learning and streaming.
\newblock \emph{Advances in neural information processing systems}, 33:\penalty0 14879--14890, 2020.

\bibitem[Buzzega et~al.(2020)Buzzega, Boschini, Porrello, Abati, and Calderara]{buzzega2020dark}
Pietro Buzzega, Matteo Boschini, Angelo Porrello, Davide Abati, and Simone Calderara.
\newblock Dark experience for general continual learning: a strong, simple baseline.
\newblock \emph{Advances in neural information processing systems}, 33:\penalty0 15920--15930, 2020.

\bibitem[Carta et~al.(2023)Carta, Pellegrini, Cossu, Hemati, and Lomonaco]{carta2023avalanche}
Antonio Carta, Lorenzo Pellegrini, Andrea Cossu, Hamed Hemati, and Vincenzo Lomonaco.
\newblock Avalanche: A pytorch library for deep continual learning.
\newblock \emph{Journal of Machine Learning Research}, 24\penalty0 (363):\penalty0 1--6, 2023.

\bibitem[Chen and Liu(2018)]{chen2018lifelong}
Zhiyuan Chen and Bing Liu.
\newblock Lifelong machine learning.
\newblock \emph{Synthesis Lectures on AI and ML}, 2018.

\bibitem[Chen and Liu(2022)]{chen2022lifelong}
Zhiyuan Chen and Bing Liu.
\newblock \emph{Lifelong machine learning}.
\newblock Springer Nature, 2022.

\bibitem[De~Lange et~al.(2021)De~Lange, Aljundi, Masana, Parisot, Jia, Leonardis, Slabaugh, and Tuytelaars]{de2021continual}
Matthias De~Lange, Rahaf Aljundi, Marc Masana, Sarah Parisot, Xu Jia, Ale{\v{s}} Leonardis, Gregory Slabaugh, and Tinne Tuytelaars.
\newblock A continual learning survey: Defying forgetting in classification tasks.
\newblock \emph{IEEE TPAMI}, 2021.

\bibitem[D{\'\i}az-Rodr{\'\i}guez et~al.(2018)D{\'\i}az-Rodr{\'\i}guez, Lomonaco, Filliat, and Maltoni]{diaz2018don}
Natalia D{\'\i}az-Rodr{\'\i}guez, Vincenzo Lomonaco, David Filliat, and Davide Maltoni.
\newblock Don't forget, there is more than forgetting: new metrics for continual learning.
\newblock \emph{arXiv preprint arXiv:1810.13166}, 2018.

\bibitem[Fayek et~al.(2020)Fayek, Cavedon, and Wu]{fayek2020progressive}
Haytham~M Fayek, Lawrence Cavedon, and Hong~Ren Wu.
\newblock Progressive learning: A deep learning framework for continual learning.
\newblock \emph{Neural Networks}, 128, 2020.

\bibitem[Guo et~al.(2024)Guo, Kumar, and Tourani]{guo2024persistent}
Zhen Guo, Abhinav Kumar, and Reza Tourani.
\newblock Persistent backdoor attacks in continual learning.
\newblock \emph{arXiv preprint arXiv:2409.13864}, 2024.

\bibitem[Hayes et~al.(2020)Hayes, Kafle, Shrestha, Acharya, and Kanan]{hayes2020remind}
Tyler~L Hayes, Kushal Kafle, Robik Shrestha, Manoj Acharya, and Christopher Kanan.
\newblock Remind your neural network to prevent catastrophic forgetting.
\newblock In \emph{European conference on computer vision}, pages 466--483. Springer, 2020.

\bibitem[He et~al.(2016)He, Zhang, Ren, and Sun]{he2016deep}
Kaiming He, Xiangyu Zhang, Shaoqing Ren, and Jian Sun.
\newblock Deep residual learning for image recognition.
\newblock In \emph{Proceedings of the IEEE conference on computer vision and pattern recognition}, pages 770--778, 2016.

\bibitem[Hung et~al.(2019)Hung, Tu, Wu, Chen, Chan, and Chen]{hung2019compacting}
Ching-Yi Hung, Cheng-Hao Tu, Cheng-En Wu, Chien-Hung Chen, Yi-Ming Chan, and Chu-Song Chen.
\newblock Compacting, picking and growing for unforgetting continual learning.
\newblock \emph{NeurIPS}, 2019.

\bibitem[Isele and Cosgun(2018)]{isele2018selective}
David Isele and Akansel Cosgun.
\newblock Selective experience replay for lifelong learning.
\newblock In \emph{Proceedings of the AAAI Conference on Artificial Intelligence}, 2018.

\bibitem[Jia et~al.(2023)Jia, Zhang, Song, Liu, and Hero]{jia2023robustness}
Jinghan Jia, Yihua Zhang, Dogyoon Song, Sijia Liu, and Alfred Hero.
\newblock Robustness-preserving lifelong learning via dataset condensation.
\newblock In \emph{ICASSP 2023-2023 IEEE International Conference on Acoustics, Speech and Signal Processing (ICASSP)}, pages 1--5. IEEE, 2023.

\bibitem[Kang et~al.(2022)Kang, Shi, and Zhang]{kangcontinual}
Siteng Kang, Zhan Shi, and Xinhua Zhang.
\newblock Continual poisoning of generative models to promote catastrophic forgetting.
\newblock In \emph{NeurIPS ML Safety Workshop}, 2022.

\bibitem[Kang et~al.(2023)Kang, Shi, and Zhang]{kang2023poisoning}
Siteng Kang, Zhan Shi, and Xinhua Zhang.
\newblock Poisoning generative replay in continual learning to promote forgetting.
\newblock In \emph{International Conference on Machine Learning}, pages 15769--15785. PMLR, 2023.

\bibitem[Katharopoulos and Fleuret(2018)]{katharopoulos2018not}
Angelos Katharopoulos and Fran{\c{c}}ois Fleuret.
\newblock Not all samples are created equal: Deep learning with importance sampling.
\newblock In \emph{International conference on machine learning}, pages 2525--2534. PMLR, 2018.

\bibitem[Keydel et~al.(1996)Keydel, Lee, and Moore]{keydel1996mstar}
Eric~R Keydel, Shung~Wu Lee, and John~T Moore.
\newblock Mstar extended operating conditions: A tutorial.
\newblock \emph{Algorithms for Synthetic Aperture Radar Imagery III}, 2757:\penalty0 228--242, 1996.

\bibitem[Khan et~al.(2022)Khan, Bouaynaya, and Rasool]{khan2022adversarially}
Hikmat Khan, Nidhal~Carla Bouaynaya, and Ghulam Rasool.
\newblock Adversarially robust continual learning.
\newblock In \emph{2022 International Joint Conference on Neural Networks (IJCNN)}, pages 1--8. IEEE, 2022.

\bibitem[Khodaee et~al.(2024)Khodaee, Viktor, and Michalowski]{khodaee2024knowledge}
Pouya Khodaee, Herna~L Viktor, and Wojtek Michalowski.
\newblock Knowledge transfer in lifelong machine learning: a systematic literature review.
\newblock \emph{Artificial Intelligence Review}, 57\penalty0 (8):\penalty0 217, 2024.

\bibitem[Kingma and Ba(2014)]{kingma2014adam}
Diederik~P Kingma and Jimmy Ba.
\newblock Adam: A method for stochastic optimization.
\newblock \emph{arXiv:1412.6980}, 2014.

\bibitem[Kirkpatrick et~al.(2017)Kirkpatrick, Pascanu, Rabinowitz, Veness, Desjardins, Rusu, Milan, Quan, Ramalho, Grabska-Barwinska, et~al.]{kirkpatrick2017overcoming}
James Kirkpatrick, Razvan Pascanu, Neil Rabinowitz, Joel Veness, Guillaume Desjardins, Andrei~A Rusu, Kieran Milan, John Quan, Tiago Ramalho, Agnieszka Grabska-Barwinska, et~al.
\newblock Overcoming catastrophic forgetting in neural networks.
\newblock \emph{Proceedings of the national academy of sciences}, 114\penalty0 (13):\penalty0 3521--3526, 2017.

\bibitem[Krizhevsky et~al.(2009)Krizhevsky, Hinton, et~al.]{krizhevsky2009learning}
Alex Krizhevsky, Geoffrey Hinton, et~al.
\newblock Learning multiple layers of features from tiny images.
\newblock 2009.

\bibitem[LECUN()]{1571417126193283840}
Y. LECUN.
\newblock The mnist database of handwritten digits.
\newblock \emph{http://yann.lecun.com/exdb/mnist/}.

\bibitem[Li and Ditzler(2022)]{li2022targeted}
Huayu Li and Gregory Ditzler.
\newblock Targeted data poisoning attacks against continual learning neural networks.
\newblock In \emph{2022 International Joint Conference on Neural Networks (IJCNN)}, pages 1--8. IEEE, 2022.

\bibitem[Li and Ditzler(2023)]{li2023pacol}
Huayu Li and Gregory Ditzler.
\newblock Pacol: Poisoning attacks against continual learners.
\newblock \emph{arXiv preprint arXiv:2311.10919}, 2023.

\bibitem[Li et~al.(2020)Li, Soltanolkotabi, and Oymak]{li2020gradient}
Mingchen Li, Mahdi Soltanolkotabi, and Samet Oymak.
\newblock Gradient descent with early stopping is provably robust to label noise for overparameterized neural networks.
\newblock In \emph{International conference on artificial intelligence and statistics}, pages 4313--4324. PMLR, 2020.

\bibitem[Li et~al.(2019)Li, Zhou, Wu, Socher, and Xiong]{li2019learn}
Xilai Li, Yingbo Zhou, Tianfu Wu, Richard Socher, and Caiming Xiong.
\newblock Learn to grow: A continual structure learning framework for overcoming catastrophic forgetting.
\newblock In \emph{ICML}. PMLR, 2019.

\bibitem[Lomonaco et~al.(2021)Lomonaco, Pellegrini, Cossu, Carta, Graffieti, Hayes, De~Lange, Masana, Pomponi, Van~de Ven, et~al.]{lomonaco2021avalanche}
Vincenzo Lomonaco, Lorenzo Pellegrini, Andrea Cossu, Antonio Carta, Gabriele Graffieti, Tyler~L Hayes, Matthias De~Lange, Marc Masana, Jary Pomponi, Gido~M Van~de Ven, et~al.
\newblock Avalanche: an end-to-end library for continual learning.
\newblock In \emph{Proceedings of the IEEE/CVF Conference on Computer Vision and Pattern Recognition}, pages 3600--3610, 2021.

\bibitem[Lopez-Paz and Ranzato(2017)]{lopez2017gradient}
David Lopez-Paz and Marc'Aurelio Ranzato.
\newblock Gradient episodic memory for continual learning.
\newblock \emph{Advances in neural information processing systems}, 30, 2017.

\bibitem[Mirzasoleiman et~al.(2020)Mirzasoleiman, Cao, and Leskovec]{mirzasoleiman2020coresets}
Baharan Mirzasoleiman, Kaidi Cao, and Jure Leskovec.
\newblock Coresets for robust training of deep neural networks against noisy labels.
\newblock \emph{Advances in Neural Information Processing Systems}, 33:\penalty0 11465--11477, 2020.

\bibitem[Munteanu and Schwiegelshohn(2018)]{munteanu2018coresets}
Alexander Munteanu and Chris Schwiegelshohn.
\newblock Coresets-methods and history: A theoreticians design pattern for approximation and streaming algorithms.
\newblock \emph{KI-K{\"u}nstliche Intelligenz}, 32:\penalty0 37--53, 2018.

\bibitem[New et~al.(2022)New, Baker, Nguyen, and Vallabha]{new2022lifelong}
Alexander New, Megan Baker, Eric Nguyen, and Gautam Vallabha.
\newblock Lifelong learning metrics.
\newblock \emph{arXiv preprint arXiv:2201.08278}, 2022.

\bibitem[Pellegrini et~al.(2020)Pellegrini, Graffieti, Lomonaco, and Maltoni]{pellegrini2020latent}
Lorenzo Pellegrini, Gabriele Graffieti, Vincenzo Lomonaco, and Davide Maltoni.
\newblock Latent replay for real-time continual learning.
\newblock In \emph{2020 IEEE/RSJ International Conference on Intelligent Robots and Systems (IROS)}, pages 10203--10209. IEEE, 2020.

\bibitem[Rebuffi et~al.(2017)Rebuffi, Kolesnikov, Sperl, and Lampert]{rebuffi2017icarl}
Sylvestre-Alvise Rebuffi, Alexander Kolesnikov, Georg Sperl, and Christoph~H Lampert.
\newblock icarl: Incremental classifier and representation learning.
\newblock In \emph{Proceedings of the IEEE conference on Computer Vision and Pattern Recognition}, pages 2001--2010, 2017.

\bibitem[Rolnick et~al.(2019)Rolnick, Ahuja, Schwarz, Lillicrap, and Wayne]{rolnick2019experience}
David Rolnick, Arun Ahuja, Jonathan Schwarz, Timothy Lillicrap, and Gregory Wayne.
\newblock Experience replay for continual learning.
\newblock \emph{Advances in neural information processing systems}, 32, 2019.

\bibitem[Ru et~al.(2024)Ru, Cao, Liu, Moore, Zhang, Zhu, Wei, and Yan]{ru2024maintaining}
Xiaolei Ru, Xiaowei Cao, Zijia Liu, Jack~Murdoch Moore, Xin-Ya Zhang, Xia Zhu, Wenjia Wei, and Gang Yan.
\newblock Maintaining adversarial robustness in continuous learning.
\newblock \emph{arXiv preprint arXiv:2402.11196}, 2024.

\bibitem[Rusu et~al.(2016)Rusu, Rabinowitz, Desjardins, Soyer, Kirkpatrick, Kavukcuoglu, Pascanu, and Hadsell]{rusu2016progressive}
Andrei~A Rusu, Neil~C Rabinowitz, Guillaume Desjardins, Hubert Soyer, James Kirkpatrick, Koray Kavukcuoglu, Razvan Pascanu, and Raia Hadsell.
\newblock Progressive neural networks.
\newblock \emph{arXiv:1606.04671}, 2016.

\bibitem[Saha et~al.(2020)Saha, Subramanya, and Pirsiavash]{saha2020hidden}
Aniruddha Saha, Akshayvarun Subramanya, and Hamed Pirsiavash.
\newblock Hidden trigger backdoor attacks.
\newblock In \emph{Proceedings of the AAAI conference on artificial intelligence}, pages 11957--11965, 2020.

\bibitem[Schaul(2015)]{schaul2015prioritized}
Tom Schaul.
\newblock Prioritized experience replay.
\newblock \emph{arXiv preprint arXiv:1511.05952}, 2015.

\bibitem[Shin et~al.(2017)Shin, Lee, Kim, and Kim]{shin2017continual}
Hanul Shin, Jung~Kwon Lee, Jaehong Kim, and Jiwon Kim.
\newblock Continual learning with deep generative replay.
\newblock \emph{Advances in neural information processing systems}, 30, 2017.

\bibitem[Tan and Le(2019)]{tan2019efficientnet}
Mingxing Tan and Quoc Le.
\newblock Efficientnet: Rethinking model scaling for convolutional neural networks.
\newblock In \emph{ICML}. PMLR, 2019.

\bibitem[Thrun(1998)]{thrun1998lifelong}
Sebastian Thrun.
\newblock Lifelong learning algorithms.
\newblock In \emph{Learning to learn}, pages 181--209. Springer, 1998.

\bibitem[Tiwari et~al.(2022)Tiwari, Killamsetty, Iyer, and Shenoy]{tiwari2022gcr}
Rishabh Tiwari, Krishnateja Killamsetty, Rishabh Iyer, and Pradeep Shenoy.
\newblock Gcr: Gradient coreset based replay buffer selection for continual learning.
\newblock In \emph{Proceedings of the IEEE/CVF Conference on Computer Vision and Pattern Recognition}, pages 99--108, 2022.

\bibitem[Umer and Polikar(2021)]{umer2021adversarial}
Muhammad Umer and Robi Polikar.
\newblock Adversarial targeted forgetting in regularization and generative based continual learning models.
\newblock In \emph{2021 International Joint Conference on Neural Networks (IJCNN)}, pages 1--8. IEEE, 2021.

\bibitem[Umer and Polikar(2024)]{umer2024adversary}
Muhammad Umer and Robi Polikar.
\newblock Adversary aware continual learning.
\newblock \emph{IEEE Access}, 2024.

\bibitem[Umer et~al.(2020)Umer, Dawson, and Polikar]{umer2020targeted}
Muhammad Umer, Glenn Dawson, and Robi Polikar.
\newblock Targeted forgetting and false memory formation in continual learners through adversarial backdoor attacks.
\newblock In \emph{2020 International Joint Conference on Neural Networks (IJCNN)}, pages 1--8. IEEE, 2020.

\bibitem[Van~de Ven and Tolias(2019)]{van2019three}
Gido~M Van~de Ven and Andreas~S Tolias.
\newblock Three scenarios for continual learning.
\newblock \emph{arXiv preprint arXiv:1904.07734}, 2019.

\bibitem[van~de Ven et~al.(2022)van~de Ven, Tuytelaars, and Tolias]{van2022three}
Gido~M van~de Ven, Tinne Tuytelaars, and Andreas~S Tolias.
\newblock Three types of incremental learning.
\newblock \emph{Nature Machine Intelligence}, 2022.

\bibitem[Wang et~al.(2024)Wang, Frans, Abbeel, Levine, and Efros]{wang2024prioritized}
Renhao Wang, Kevin Frans, Pieter Abbeel, Sergey Levine, and Alexei~A Efros.
\newblock Prioritized generative replay.
\newblock \emph{arXiv preprint arXiv:2410.18082}, 2024.

\bibitem[Wang et~al.(2023)Wang, Shen, Zhan, Suo, Zhu, Duan, and Gao]{wang2023metamix}
Zhenyi Wang, Li Shen, Donglin Zhan, Qiuling Suo, Yanjun Zhu, Tiehang Duan, and Mingchen Gao.
\newblock Metamix: Towards corruption-robust continual learning with temporally self-adaptive data transformation.
\newblock In \emph{Proceedings of the IEEE/CVF Conference on Computer Vision and Pattern Recognition}, pages 24521--24531, 2023.

\bibitem[Xiao et~al.(2017)Xiao, Rasul, and Vollgraf]{xiao2017fashion}
Han Xiao, Kashif Rasul, and Roland Vollgraf.
\newblock Fashion-mnist: a novel image dataset for benchmarking machine learning algorithms.
\newblock \emph{arXiv preprint arXiv:1708.07747}, 2017.

\bibitem[Yang et~al.(2023)Yang, Shi, Wei, Liu, Zhao, Ke, Pfister, and Ni]{yang2023medmnist}
Jiancheng Yang, Rui Shi, Donglai Wei, Zequan Liu, Lin Zhao, Bilian Ke, Hanspeter Pfister, and Bingbing Ni.
\newblock Medmnist v2-a large-scale lightweight benchmark for 2d and 3d biomedical image classification.
\newblock \emph{Scientific Data}, 10\penalty0 (1):\penalty0 41, 2023.

\bibitem[Yoon et~al.(2017)Yoon, Yang, Lee, and Hwang]{yoon2017lifelong}
Jaehong Yoon, Eunho Yang, Jeongtae Lee, and Sung~Ju Hwang.
\newblock Lifelong learning with dynamically expandable networks.
\newblock \emph{arXiv:1708.01547}, 2017.

\bibitem[Yoon et~al.(2021)Yoon, Madaan, Yang, and Hwang]{yoononline}
Jaehong Yoon, Divyam Madaan, Eunho Yang, and Sung~Ju Hwang.
\newblock Online coreset selection for rehearsal-based continual learning.
\newblock In \emph{International Conference on Learning Representations}, 2021.

\bibitem[Zenke et~al.(2017)Zenke, Poole, and Ganguli]{zenke2017continual}
Friedemann Zenke, Ben Poole, and Surya Ganguli.
\newblock Continual learning through synaptic intelligence.
\newblock In \emph{International conference on machine learning}, pages 3987--3995. PMLR, 2017.

\end{thebibliography}
}

\clearpage
\setcounter{page}{1}

\maketitlesupplementary

\raggedbottom

\section{Proof of Theorem \ref{thm:random_noise}}

As it is mentioned in \cite{mirzasoleiman2020coresets}, the Jacobian spectrum can be split into information space $\mathcal{I}$ and nuisance space $\mathcal{N}$, associated with the large and small singular values. We first need to define subspaces $S_+$ and $S_-$.\\

First we need: 

\begin{definition}\label{df:lipschitz}
    In our following analysis, the Jacobian matrix is going to be L-smooth, which means that for any $W_1$ and $W_2$ learnable parameter matrices and $L>0$ we have that 

    $$\|\mathcal{J}(W_1) - \mathcal{J} (W_2)\|_2 \leq L \|W_1 - W_2\|_2$$
\end{definition}

\begin{definition}\label{df:subsetS}
Let $f: \mathbb{R}^p \longrightarrow \mathbb{R}^n$ be a nonlinear mapping (i.e. neural network). Then if we have $S_+, S_- \subseteq \mathbb{R}^p$ and for all unit vectors $w_+ \in S_+$ and $w_- \in S_-$ and scalars $0 \leq \mu \ll \alpha \leq \beta$ we have:

$$\alpha \leq \|\mathcal{J}^T (W, X)w_+\|_2 \leq \beta \ \ \text{and } \ \ \|\mathcal{J}^T (W, X)w_-\|_2 \leq \mu$$

Where, $W$ are the learnable parameters and $X$ is the data.
\end{definition}

To have a better idea from Definition \ref{df:subsetS}, $S_+$ is the subset consisting of the large singular values and $S_-$ is the subset consisting of the smaller singular values. It will be better for us for the future proofs to have bounds for the Jacobian $\mathcal{J}$ and so we get the following Lemma:\\

\begin{lemma}\label{lm:alphabetabounds}
Let $W$ be the learnable parameters that satisfy $y=f(W,x)$ and let $\mathcal{J}$ be the Jacobian matrix. Define the matrices, $\mathcal{J}_+ = \Pi_{S_+}(\mathcal{J})$ and $\mathcal{J}_- = \Pi_{S_-}(\mathcal{J})$, where $\pi_{S_+}(\mathcal{J})$ and $\pi_{S_-}(\mathcal{J})$ are the projections of $\mathcal{J}$ onto $S_+$ and $S_-$, respectively. Then from Definition \ref{df:subsetS} we have:

$$\alpha \leq \|\mathcal{J}_+\|_2 \leq \beta \ \ \ , \|\mathcal{J}_-\|_2\leq \mu$$

\end{lemma}
\begin{proof} By just using the definition of the norm of a matrix we obtain,

$$\alpha \leq \|\mathcal{J}_+\|_2 = \sup _{w_+\in S_+, \|w_+\|_2 = 1} \|\mathcal{J}^Tw_+\|_2 \leq \beta \ \ \text{and} \ \|\mathcal{J}_-\|_2 = \sup _{w_-\in S_-, \|w_-\|_2 = 1} \|\mathcal{J}^Tw_-\|_2 \leq \mu$$

\end{proof}

Prior to proceeding with the proofs, we must define our learning algorithm (in this case, a simple Gradient Descent) and the applied loss function.

\begin{equation}\label{eq:GradientDescent}
    \hat{W} = W - \eta \nabla \mathcal{L} (W)
\end{equation}

\begin{equation}\label{eq:MSE}
    \mathcal{L} (W) = \frac{1}{2} \sum_{i \in V} (y_i - f(W, x_i))^2 \ , \ \text{V} = \{1, 2, .., n\}
\end{equation}

Combining the two previous equations, we obtain the following expression for iteration $t+1$:

\begin{equation}\label{eq:totalGD}
    W_{t+1} = W_t - \eta \nabla \mathcal{L} (W_t) = W_t - \eta \mathcal{J}(W_t)^T (f(W_t, x) - y)
\end{equation}

where $\eta$ is the learning rate, $\nabla \mathcal{L} (W)$ is the gradient at $W$, $\mathcal{L}$ is the MSE and $\mathcal{J}$ is the Jacobian matrix.\\

Furthermore, we define the average Jacobian definition as it will be used throughout our analysis.

\begin{definition}\label{df:AverageJacobian} (Average Jacobian) \cite{li2020gradient}
    We define the $Average \ Jacobian$ along the path connecting two points $u, v \in \mathbb{R}^p$ as 

    $$\bar{\mathcal{J}}(v,u) = \int_0^1 \mathcal{J} (u+\alpha (v-u))d \alpha$$
\end{definition}

\begin{lemma}\label{lm:LinearizationResidual} \cite{li2020gradient}
    Given the Gradient Descent in Equation \ref{eq:GradientDescent} we define 

    $$C(W) = \bar{\mathcal{J}} (\hat{W}, W) \mathcal{J}(W)^T$$

    By using the residual formulas $\hat{r} = f(\hat{W}) - y$ and $r = f(W) - y$ we obtain the following equation

    $$\hat{r} = (I - \eta C(W) )r$$
\end{lemma}
\begin{proof}
    By using the residual formulas we get
\begin{align}
    \hat{r} &= f(\hat{W}) - y \\
    &\overset{\text{(a)}}{=} f(\hat{W}) + r - f(W) \\
    &\overset{\text{(b)}}{=} r + \bar{\mathcal{J}}(\hat{W}, W) (\hat{W} - W) \\
    &\overset{\text{(c)}}{=} r - \eta \bar{\mathcal{J}} (\hat{W}, W) \mathcal{J} (W)^T r \\
    &\overset{\text{(d)}}{=} (I - \eta C(W)) r
\end{align}

Now, let us prove each of the previously stated equalities. For (a), we can establish its validity using the residual formulas, leading to the expression $-y = r - f(W)$. (b) This part requires more work. First we need to prove that $f(\hat{W}) - f(W) = \bar{\mathcal{J}}(\hat{W}, W) (\hat{W} - W)$. Consider Definition \ref{df:AverageJacobian}. Let us introduce the following variable substitutions: $\hat{W} = v$ and $W = u$. We perform a change of variables $t = u + \alpha(v-u)$, which yields $d\alpha = \frac{dt}{v-u}$ with bounds $t \in [u,v]$ and thus we get:
\begin{equation}\label{eq:integral}
    \bar{\mathcal{J}}(v, u) = \int_u^v \mathcal{J} (t) \frac{dt}{v-u} = \frac{1}{v-u} \int_u^v \mathcal{J} (t) dt = \frac{1}{v-u} f(t)|_u^v = \frac{f(v) - f(u)}{v-u}
\end{equation}

The last two equations at the previous equalities are true due to the fact that $\mathcal{J}$ is the Jacobian. This proves (b) since by changing $v$ and $u$ back to $\hat{W}$ and $W$ we get $f(\hat{W}) - f(W) = \bar{\mathcal{J}} (\hat{W}, W) (\hat{W} - W)$. (c) is true from Equation \ref{eq:GradientDescent} and we get $\hat{W} - W = - \eta \nabla \mathcal{L} (W)$. (d) holds true because we can factor $r$, and we have defined $C(W) = \bar{\mathcal{J}}(\hat{W}, W) \mathcal{J}(W)^T$.
    
\end{proof}

\begin{lemma}\label{lm:PerturbedData} (Perturbed dataset)\\

    If we follow the previous Lemma \ref{lm:LinearizationResidual} but applying perturbed data $X + E_X$ instead of $X$ then 
    
    $$\hat{r} \approx (I - \eta (C(W) + E_{J_2} \mathcal{J}(W)^T + E_{J_1} \bar{\mathcal{J}} (\hat{W}, W))r$$

    where $E_{J_2}  := \bar{\mathcal{J}}(\hat{W}, W, E_X)$ and $E_{J_1} := \mathcal{J} (W, E_X)^T$. 
     
\end{lemma}
\begin{proof}
    We continue in a similar way as in Lemma \ref{lm:LinearizationResidual} but the Jacobian matrices change as follows $\bar{\mathcal{J}} (\hat{W}, W, X+E_X) = \bar{\mathcal{J}} (\hat{W}, W, X) + \bar{\mathcal{J}} (\hat{W}, W, E_X)$. And we write, $\bar{\mathcal{J}} (\hat{W}, W, E_X) = E_{J_2}$. In a similar way, $\mathcal{J}(W, X + E_X)^T = \mathcal{J} (W, X)^T + \mathcal{J} (W, E_X)^T$ and we write, $\mathcal{J} (W, E_X)^T = E_{J_1}$. Following the equalities in Lemma \ref{lm:LinearizationResidual} with the corresponding changes to the perturbed images $X + E_X$ we obtain the equality:

    $$\hat{r} \approx (I - \eta (C(W) + E_{J_2} \mathcal{J}(W)^T + \bar{\mathcal{J}} (\hat{W}, W) E_{J_1}))r$$

    (Numerical analysis suggests we can ignore the second order error term $E_{J_2}E_{J_1}$).
\end{proof}

We require the following Lemma:

\begin{lemma}\label{lm:firstboundsJacobian}

    If we have that the Jacobian is L-smooth as in Definition \ref{df:lipschitz}, then we get:

    $$\|\bar{\mathcal{J}}(W_2, W_1) - \mathcal{J}(W_1)\|_2 \lessapprox \frac{\alpha (\beta + E_{J_1})}{2 \beta} + \|E_{J_2} - E_{J_1}\|_2$$

    where $\alpha$ and $\beta$ are given as in Definition \ref{df:subsetS}, Jacobian and Average Jacobian are applied to $X+E_X$ and $E_{J_1}$ and $E_{J_2}$ are the errors from perturbed data in Jacobian and Average Jacobian.
\end{lemma}
\begin{proof} We will first give all the equalities and inequalities and then explain in details. 

{\large
\begin{align}
    \|\bar{\mathcal{J}} (W_2, W_1) - \mathcal{J}(W_1)\|_2 &\overset{\text{(a)}}{\approx} \|\int_0^1 \mathcal{J} (W_1 + t(W_2-W_1))dt + E_{J_2} - \mathcal{J}(W_1)  - E_{J_1}\|_2 \\
    &\overset{\text{(b)}}{=} \|\int_0^1 (\mathcal{J} (W_1 + t(W_2-W_1)) + E_{J_2} - \mathcal{J}(W_1)  - E_{J_1})dt\|_2\\
    &\overset{\text{(b)}}{\leq} \int_0^1 \|(\mathcal{J} (W_1 + t(W_2-W_1)) - \mathcal{J}(W_1)) \|_2 dt + \|E_{J_2} - E_{J_1}\|_2\\
    &\overset{\text{(d)}}{\leq} \int_0^1 L\|W_1 + t (W_2 - W_1) -W_1\|_2 dt + \|E_{J_2} - E_{J_1}\|_2 \\
    &= \int_0^1 Lt\|W_2 - W_1\|_2 dt + \|E_{J_2} - E_{J_1}\|_2\\
    &= L\|W_2 - W_1\|_2 \frac{t^2}{2}|_0^1 + \|E_{J_2} - E_{J_1}\|_2\\
    &= \frac{L}{2} \|W_2 - W_1\|_2 + \|E_{J_2} - E_{J_1}\|_2
\end{align}
}

In part (a), we use the Average Jacobian formula from Definition \ref{df:AverageJacobian}. At part (b) we use the fact that $\int_0^1 (E_{J_2} - \mathcal{J}(W_1)  - E_{J_1}) \, dt = (E_{J_2} - \mathcal{J}(W_1)  - E_{J_1}) \int_0^1 dt = (E_{J_2} - \mathcal{J}(W_1)  - E_{J_1})$, where $E_{J_2}$ is the error from the perturbed data using the Average Jacobian and $E_{J_1}$ is the corresponding error for the Jacobian. At (c), we use the properties of norms in the integral and note that $E_{J_2}$ and $E_{J_1}$ do not depend on $t$, which gives us $\int_0^1 \|E_{J_2} - E_{J_1}\|_2 dt = \|E_{J_2} - E_{J_1}\|_2 t\mid_0^1 = \|E_{J_2} - E_{J_1}\|_2$.since they are constants. At (d) we use the fact that the Jacobian is L-smooth. Then, we simplify the formulas and find the integral.\\

For $\eta \leq \frac{\alpha}{L \beta \|r_0\|_2}$, where $r_0$ is the initial residual and $\alpha, \beta$ are defined as in Definition \ref{df:subsetS}, and $L$ is the Lipschitz constant, we also have 

$$W_{\mathcal{T}+1} = W_\mathcal{T} - \eta \nabla \mathcal{L}(W_\mathcal{T}) = W_\mathcal{T} - \eta \mathcal{J}(W_\mathcal{T})^T f(W_\mathcal{T}, x).$$ 

From our previous work in this lemma, we obtain

$$\|\bar{\mathcal{J}}(W_{\mathcal{T}+1}, W_{\mathcal{T}}) - \mathcal{J}(W_{\mathcal{T}})\|_2 \leq \frac{L}{2} \|W_{\mathcal{T}+1} - W_{\mathcal{T}}\|_2 + \|E_{J_2} - E_{J_1}\|_2.$$

Furthermore we get the equations, 

\begin{align}
    \|\bar{\mathcal{J}} (W_{\mathcal{T}+1}, W_{\mathcal{T}}) - \mathcal{J} (W_{\mathcal{T}})\|_2 &\overset{\text{(a)}}{\lessapprox} \frac{\eta L}{2} \|(\mathcal{J}(W_\mathcal{T})^T + E_{J_1}) (f(W_\mathcal{T}) - y)\|_2 + \|E_{J_2} - E_{J_1}\|_2\\
    &\overset{\text{(b)}}{\leq} \frac{\eta (\beta + E_{J_1}) L}{2} \|r_{\mathcal{T}}\|_2 + \|E_{J_2} - E_{J_1}\|_2\\
    &\leq \frac{\eta (\beta + E_{J_1}) L}{2} \|r_{0}\|_2 \leq \frac{\alpha (\beta + E_{J_1})}{2 \beta} + \|E_{J_2} - E_{J_1}\|_2
\end{align}

At (a), we utilized the equation $W_{\mathcal{T}+1} = W_\mathcal{T} - \eta \mathcal{J}(W_\mathcal{T})^T f(W_\mathcal{T}, x)$, which leads to the expression $- \eta \mathcal{J}(W_\mathcal{T})^T f(W_\mathcal{T}, x) = W_{\mathcal{T}+1} - W_\mathcal{T}$. Then, in (b), we applied Lemma \ref{lm:alphabetabounds}, which gives us $\|\mathcal{J}(W_\mathcal{T})^T\|_2 \leq \beta$. Finally, we used the fact that $\eta \leq \frac{\alpha}{L \beta \|r_0\|_2}$ in the last inequality.

\end{proof}

To apply it in other proofs, we also need the following lemma:

\begin{lemma}\label{lm:general} \cite{li2020gradient}\\

Let matrices $A, C \subset \mathbb{R}^{n \times p}$ where $\|A-C\|_2 \leq \frac{\alpha}{2}$. Furthermore, if we assume $A$ and $C$ lie in $S_+$ then we get:

$$AC^T \geq \frac{CC^T}{2}$$    
\end{lemma}
\begin{proof} For every $u \in S_+$ with unit Euclidean norm we get 

\begin{align}
    u^T AC^Tu &= u^T(C + (A-C))C^Tu = u^T CC^Tu + u^T(A-C)C^Tu\\
    &= \|C^Tu\|_2^2 + u^T (A-C)C^Tu \overset{\text{(a)}}{\geq} \|C^Tu\|_2^2 - \|u^T(A-C)\|_2\|C^Tu\|_2\\
    &\overset{\text{(b)}}{=} (\|C^Tu\|_2 - \|u^T(A-C)\|_2) \|C^Tu\|_2 \overset{\text{(c)}}{\geq} (\|C^Tu\|_2 - \frac{\alpha}{2} \|C^Tu\|_2\\
    &\overset{\text{(d)}}{\geq} \frac{\|C^Tu\|_2^2}{2} = \frac{u^TCC^Tu}{2}
\end{align}

First, we need to clarify the equalities and inequalities from the previous equations. In the first row, we rewrite $A = C + (A - C)$ and then expand. In the second row, we start with the fact that $\|C^T u\|_2^2 = u^T C C^T u$, and at (a), we apply the product norm and change of signs. In (b), we factor out $\|C^T u\|_2$, and in (c), we use the lemma stating that $\|A - C\|_2 \leq \frac{\alpha}{2}$, which implies $-\|A - C\|_2 \geq -\frac{\alpha}{2}$. In (d), we note that $\alpha > 0$ because we have $0 \leq \mu \ll \alpha \leq \beta$. Finally, using the fact that $\|u\|_2 = 1$ since they are unit vectors, we obtain 

$$AC^T \geq \frac{CC^T}{2}.$$
    
\end{proof}

We can use Lemma \ref{lm:general} even when the constant is $\frac{\alpha (\beta + E)}{2 \beta} + \frac{|E - E_{av}|}{2}$ instead of $\frac{\alpha}{2}$ and the proof is in a similar way since $\frac{\alpha (\beta + E)}{2 \beta} + \frac{|E - E_{av}|}{2}$ is also positive.\\

\begin{lemma}\label{lm:paperbefore}\cite{li2020gradient}
    By using the results from Lemma \ref{lm:LinearizationResidual}, prove that 

    $$\|r_{\mathcal{T}+1}\|_2 \leq (1 - \frac{\eta \alpha^2}{2})\|r_{\mathcal{T}}\|_2$$
\end{lemma}
\begin{proof}
    We will first present all the equations and inequalities, and then provide an explanation for each part.

    \begin{align}
        \|r_{\mathcal{T}+1}\|_2^2 = \|(I-\eta C(W_{\mathcal{T}}))r_{\mathcal{T}}\|_2^2 &\overset{\text{(a)}}{=} \|r_{\mathcal{T}}\|_2 -2 \eta r_{\mathcal{T}}C(W_{\mathcal{T}}) r_{\mathcal{T}}\\
        &\hspace{2cm} + \eta^2 r_{\mathcal{T}}^T C(W_{\mathcal{W}})^T C(W_{\mathcal{T}}) r_{\mathcal{T}}\\
        &\overset{\text{(b)}}{\leq} \|r_{\mathcal{T}}\|_2^2 - \eta r_{\mathcal{T}} \mathcal{J}(W_{\mathcal{T}}) \mathcal{J}(W_{\mathcal{T}})^T r_{\mathcal{T}}\\
        &\hspace{2cm} + \eta^2 \beta^2 r_{\mathcal{T}}^T \mathcal{J}(W_{\mathcal{T}}) \mathcal{J}(W_{\mathcal{T}})^T r_{\mathcal{T}}\\
        &\overset{\text{(c)}}{\leq} \|r_{\mathcal{T}}\|_2^2 - (\eta - \eta^2 \beta^2) \|\mathcal{J} (W_{\mathcal{T}}) ^Tr_{\mathcal{T}}\|_2^2\\
        &\overset{\text{(d)}}{\leq} \|r_{\mathcal{T}}\|_2^2 - \frac{\eta}{2} \|\mathcal{J} (W_{\mathcal{T}}) ^Tr_{\mathcal{T}}\|_2^2\\
        &\overset{\text{(e)}}{\leq} (1 - \frac{\eta \alpha^2}{2}) \|r_{\mathcal{T}}\|_2^2
    \end{align}

At (a) we use the properties of the two norm, i.e. $\|x\|_2^2 = x^Tx$ and then simplify. At (b) we use results from Lemma \ref{lm:general} and the fact that $\|\bar{\mathcal{J}} (W_{\mathcal{T}+1}, W_\mathcal{T})\|_2 \leq \beta$ and Definition \ref{df:AverageJacobian}. At (c) we combine what we have at (b) and use the fact that $r_{\mathcal{T}}^T \mathcal{J}(W_{\mathcal{T}}) \mathcal{J}(W_{\mathcal{T}})^T r_{\mathcal{T}} = \|\mathcal{J} (W_{\mathcal{T}})^T r_{\mathcal{T}}\|_2^2$. At (d) we use the fact that $\eta \leq \frac{1}{2 \beta^2}$ and plug in $\eta$ at (c). At (e) we use the fact that $\|\mathcal{J}(W_{\mathcal{T}})\|_2 \geq \alpha$ and so $-\|\mathcal{J}(W_{\mathcal{T}})\|_2 \leq -\alpha$.
\end{proof}

In the next Lemma, we are going to introduce the perturbed dataset:\\

\begin{lemma}\label{lm:ourbound}
    By following the results from Lemma \ref{lm:PerturbedData} prove that,

    $$\|r_{t+1}\|_2^2 \lessapprox (1-\eta (\frac{\alpha}{2} + 2 E_{J_2}\alpha + 2 E_{J_1} \alpha - 2 \eta E_{J_2} \frac{\beta^3}{2} - 2 \eta E_{J_1} \frac{\beta}{3}))\|r_t\|_2^2$$
\end{lemma}
\begin{proof}
    A portion of the proof follows a similar approach to Lemma \ref{lm:paperbefore}, and we will first present each equality and inequality before explaining them in detail. From Lemma \ref{lm:PerturbedData}, when the perturbation occurs in the data rather than the labels, we have $r_{t+1} = (I - \eta (C(W_t + E_{J_2}\mathcal{J}(W_t)^T + \bar{\mathcal{J}}(W_{t+1}, W_t)E_{J_1})))r_t$. We use similar reasoning to Lemma \ref{lm:paperbefore} to derive the result.

    \begin{align}
        r_{\tau +1}^Tr_{\tau+1} &= (r_\tau^T -\eta r_\tau^TC(W_\tau) - \eta E_{J_2}r_\tau^T\mathcal{J}(W_\tau) - \eta E_{J_1}r_\tau^T \bar{\mathcal{J}}(W_{\tau+1}, W_\tau)^T)\\
        &\hspace{2cm} (r_\tau - \eta C(W_\tau)r_\tau - \eta E_{J_2}\mathcal{J}(W_\tau)^Tr_\tau-\eta \bar{\mathcal{J}}(W_{\tau+1}, W_\tau)E_{J_1}r_\tau)\\
        &\overset{\text{(a)}}{\lessapprox} \|r_\tau\|_2^2 - \frac{\eta \alpha^2}{2} \|r_\tau\|_2^2-2 \eta E_{J_2} r_\tau^T \mathcal{J}(W_\tau)r_\tau -\eta E_{J_1} r_\tau^T \bar{\mathcal{J}}(W_{\tau+1}, W_\tau) r_\tau\\
        &\hspace{2cm}+ 2 \eta^2 E_{J_2}r_\tau^T\mathcal{J}(W_\tau)C(W_\tau)r_\tau + 2 \eta^2 E_{J_1} r_\tau^T \bar{\mathcal{J}}(W_{\tau+1}, W_\tau)C(W_\tau)r_\tau\\
        &\overset{\text{(b)}}{\leq} (1-\eta (\frac{\alpha}{2} + 2 E_{J_2}\alpha + 2 E_{J_1} \alpha - 2 \eta E_{J_2} \frac{\beta^3}{2} - 2 \eta E_{J_1} \frac{\beta}{3}))\|r_t\|_2^2
    \end{align}

    At step (a), we applied the bounds from Lemma \ref{lm:paperbefore}, and the remaining part simplifies the multiplication from the first equality. In step (b), we use the facts that $\|\mathcal{J} (W_\tau)\|_2 \leq \beta$, $\|\bar{\mathcal{J}} (W_{\tau+1}, W_\tau)\|_2 \leq \beta$, and also that $-\|\mathcal{J} (W_\tau)\|_2 \leq -\alpha$ and $-\|\bar{\mathcal{J}} (W_{\tau+1}, W_\tau)\|_2 \leq -\alpha$.

\end{proof}

By continuing the calculations in Lemma \ref{lm:ourbound} back to the first iteration and assuming $\mathcal{T}$ iterations, we obtain:

\begin{equation}\label{eq:mainequationiterations}
    \|r_\mathcal{T}\|_2^2 \lessapprox \left(1-\eta \left(\frac{\alpha}{2} + 2 E_{J_2}\alpha + 2 E_{J_1}\alpha - 2 \eta E_{J_2} \frac{\beta^3}{2} - 2 \eta E_{J_1} \frac{\beta}{3}\right)\right)^\mathcal{T}\|r_0\|_2^2
\end{equation}

\textbf{Now we give the Lemma about the bounds for $T_{total}$ where $T_{total}$ is the number of iterations}

\begin{lemma}\label{lm:iterationsbound} (Bounds for iterations)\\
    Continuing with the results from Equation \ref{eq:mainequationiterations} and if $\nu>0$ to bound the error after $T_{total}$ iterations $\|r_{T_{total}}\|_2=\|f(W_{T_{total}},X)-Y\|_2 \leq \nu$ then after $T_{total}$ iterations the Neural Network $f$ correctly classifies where $T_{total}$ is 

    $$T_{total} \geq \mathcal{O}(\frac{1}{\eta (\frac{\alpha}{2} +  E_{J_2}\alpha +  E_{J_1} \alpha -  \eta E_{J_2} \frac{\beta^3}{2} - \eta E_{J_1} \frac{\beta}{3})} \log \frac{\|r_0\|_2}{\nu})$$
\end{lemma}
\begin{proof}
    We first need to establish some general formulas that will be useful for the subsequent steps. Specifically, we need to show that for any $0 < x < 1$, the inequality $\log \left( \frac{1}{1-x} \right) \geq \log (1+x)$ holds. 

First, we subtract $\log(1+x)$ from both sides of the inequality:
$$
\log \left( \frac{1}{1-x} \right) - \log(1+x) \geq 0.
$$
Using the logarithmic identity $\log(a) - \log(b) = \log\left(\frac{a}{b}\right)$, this simplifies to:
$$
\log \left( \frac{1}{(1-x)(1+x)} \right) \geq 0.
$$

Now, we simplify the expression inside the logarithm:
$$
(1-x)(1+x) = 1 - x^2.
$$
Thus, the inequality becomes:
$$
\log \left( \frac{1}{1 - x^2} \right) \geq 0,
$$
which is equivalent to:
$$
\frac{1}{1 - x^2} \geq 1.
$$

Next, we multiply both sides by $1 - x^2$ (which is positive for $0 < x < 1$):
$$
1 \geq 1 - x^2.
$$
This is true since $0<x<1$ and thus it proves our inequality.\\

Furthermore, we need to prove that $g(x) = \log (1+x) - \frac{x}{2}$ is also increasing. We still need the derivatives and we get 

$$g'(x) = \frac{1}{x+1} - \frac{1}{2}=\frac{1-x}{2(x+1)}$$

Which means that $g'(x)>0$ for $0<x<1$.\\

We can now proceed with the proof of the lemma using the results from Equation \ref{eq:mainequationiterations}. For some $\nu > 0$, we require that $\|r_{T_{total}}\|_2 \leq \nu$, where $T_{total}$ is the number of iterations needed for accurate classification. We want $\nu$ to be very small, which implies that $r_{T_{total}} = f(W_{T_{total}}, X) - y$ should also be very small. Expanding from Equation \ref{eq:mainequationiterations}, we obtain

$$(1 - \eta \left(\frac{\alpha}{2} + 2 E_{J_2} \alpha + 2 E_{J_1} \alpha - 2 \eta E_{J_2} \frac{\beta^3}{2} - 2 \eta E_{J_1} \frac{\beta}{3}\right))^{T_{total}} \|r_0\|_2 \leq \nu.$$

Furthermore, we can express this as 

$$1 - \eta \left(\frac{\alpha}{2} + 2 E_{J_2} \alpha + 2 E_{J_1} \alpha - 2 \eta E_{J_2} \frac{\beta^3}{2} - 2 \eta E_{J_1} \frac{\beta}{3}\right) = 1 - A.$$

\begin{align}
    (1 - A)^{T_{total}} &\overset{\text{(a)}}{\leq} \frac{\nu}{\|r_0\|_2}\\
    T_{total} \log (1 - A) &\overset{\text{(b)}}{\leq} \log \frac{\nu}{\|r_0\|_2}\\
    T_{total} \log (\frac{1}{1-A}) &\overset{\text{(c)}}{\geq} \log (\frac{\|r_0\|_2}{\nu})\\
\end{align}

At (a), this is valid because we move $\|r_0\|_2$ to the other side without changing the sign, as it is positive. In (b), we apply the logarithm and use the property that allows us to bring powers in front of the log. In (c), we multiply both sides by -1 and make the necessary adjustments to the logarithmic functions. Thus, using what we established at the beginning of the proof of this lemma, we obtain

$$T_{total} \log(1 + A) \geq \log \frac{\|r_0\|_2}{\nu}.$$

Moreover, from another result earlier in this lemma, we have

$$T_{total} \frac{A}{2} \geq \log \frac{\|r_0\|_2}{\nu}.$$

This leads us to our final results.

$$T_{total} \geq \frac{2}{A} \log \frac{\|r_0\|_2}{\nu}=\frac{2}{\eta (\frac{\alpha}{2} + 2 E_{J_2}\alpha + 2 E_{J_1} \alpha - 2 \eta E_{J_2} \frac{\beta^3}{2} - 2 \eta E_{J_1} \frac{\beta}{3})} \log \frac{\|r_0\|_2}{\nu}$$

\end{proof}

We now present the Main Theorem, with certain parts modified from the results of Theorem 4.1 in \cite{mirzasoleiman2020coresets}.

\begin{theorem}\label{thm:maintheorem} (Main Theorem)\\
Assume we apply gradient descent using mean-squared error (MSE) loss to train a neural network on a dataset with perturbed data. Suppose the Jacobian is $L$-smooth. The Coresets found by CRUST \cite{mirzasoleiman2020coresets}, considering perturbed datasets where $\delta$ represents the fraction of noisy data and $k$ is the Coreset size, approximate the Jacobian matrix with an error of at most $\epsilon \leq \mathcal{O}\left(\frac{\delta \alpha^2}{k \beta \log (\delta)}\right)$, where $\alpha = \sqrt{r_{min}} \sigma_{min}(\mathcal{J}(W, X_S)) - E_{min}$ and $\beta = \|\mathcal{J}(W, X)\|_2 + \epsilon + E_{max}$, with further details regarding $E_{min}$ and $E_{max}$ provided in the proof. For $L \leq \frac{\alpha \beta}{L \sqrt{2k}}$ and a step size $\eta = \frac{1}{2 \beta^2}$, after $T_{total} \geq \mathcal{O}\left(\frac{1}{\eta \left(\frac{\alpha}{2} + E_{J_2} \alpha + E_{J_1} \alpha - \eta E_{J_2} \frac{\beta^3}{2} - \eta E_{J_1} \frac{\beta}{3}\right)} \log \frac{\|r_0\|_2}{\nu}\right)$ iterations, the network correctly classifies all selected elements.
\end{theorem}
\begin{proof}
    The proof regarding $T_{total}$ and the network's correct classification is based on Lemma \ref{lm:iterationsbound}, which helps us determine the values for $\alpha$ and $\beta$. Consider a vector $v$, where we define $\hat{p} = \mathcal{J}_r(W, X_S + E_{X_S}) v$ and $p = \mathcal{J}(W, X_S + E_{X_S}) v$ (with $\mathcal{J}_r$ as described in \cite{mirzasoleiman2020coresets}). As explained in that reference, to obtain $\hat{p}$, we need to multiply the entries of $p$ by a factor between $r_{min}$ and $r_{max}$. Thus, when working with vectors over $S_+$, we have 

    {\large
    $$\sqrt{r_{min}} \sigma_{min}(\mathcal{J}(W, X_S + E_{X_S}), S_+) \leq \sigma_{i \in [k]}(\mathcal{J}_r(W, X_S + E_{X_S}), S_+) \leq \sqrt{r_{max}} \|\mathcal{J}(W, X_S)\|_2.$$
    }
    
    After analyzing the Jacobian in the middle part and subtracting that value from all three sides, we can simplify the left and right sides to obtain $E_{min}$ and $E_{max}$.
    
    The proof concerning $\epsilon$ is straightforward; we can follow the steps outlined in \cite{mirzasoleiman2020coresets}. In their proof, they focus on the residual $r = f(W, X) - Y$, which involves noisy labels. In our case, however, we are dealing with perturbed images. The key difference is that $\delta$ in our context pertains to the noise introduced by perturbations to the images, rather than being a parameter related to flipped labels.
\end{proof}

\section{Experimental parameters}

\noindent\hspace{0pt}\begin{minipage}{0.99\textwidth}
\RaggedRight

In the following table, we define the hyperparameters used for the different datasets in our experiments. Not that Coreset size refers to the number of samples stored in the individual per-class Coresets. Note that in all experiments, we use five random seeds, and class ordering is shuffled with each seed. Our curriculum involves learning on two classes in the first experience, and learning on a single new class for all subsequent experiences. \textbf{Note that this differs from the standard curriculum set up used for ``splitMNIST'' and ``splitCIFAR''.} Our set up was selected to mimic real world operating conditions for our applications-of-interest where one class may be added at a time (one-class incremental learning), and in many cases, may be a more challenging setting than the standard multiple-classes-per-experience setting. The baseline and prior work approaches all follow very similar training procedure to the main methods (Continual CRUST and cosine-CRUST). I.e., we use the same total number of epochs, same loss functions (when applicable), same per-class memory sizes, etc. Note that we do tune the learning rates for each strategy separately based on one seed to try to compare against the best performance of each comparison method.

\begin{table}[H]
\centering
\resizebox{\textwidth}{!}{\begin{tabular}{|c|c|c|c|c|c|c|c|}
\hline
\textbf{Dataset} & \textbf{\begin{tabular}[c]{@{}c@{}}Model\end{tabular}} & \textbf{\begin{tabular}[c]{@{}c@{}}Learning\\ Rate\end{tabular}} & \textbf{\begin{tabular}[c]{@{}c@{}}Weight\\ Decay\end{tabular}} & \textbf{\begin{tabular}[c]{@{}c@{}}Batch \\ Size\end{tabular}} & \textbf{\begin{tabular}[c]{@{}c@{}}Phase 1\\ Epochs\end{tabular}} & \textbf{\begin{tabular}[c]{@{}c@{}}Phase 2\\ Epochs\end{tabular}} & \textbf{\begin{tabular}[c]{@{}c@{}}Coreset \\ Size\end{tabular}} \\ \hline
MNIST (Label Noise) & SimpleCNN & 1.0e-3 & 1.0e-6 & 32 & 40 & 20 & 300 \\ \hline
MNIST (Instance Noise) & SimpleCNN & 1.0e-3 & 1.0e-6 & 32 & 40 & 20 & 128 \\ \hline
FashionMNIST (Label Noise) & Resnet18 & 3.0e-2 & 1.0e-6 & 32 & 40 & 20 & 300 \\ \hline
FashionMNIST (Instance Noise) & Resnet18 & 1.0e-3 & 1.0e-6 & 64 & 40 & 20 & 128 \\ \hline
CIFAR10 & Resnet18 & 4.0e-2 & 1.0e-6 & 32 & 40 & 20 & 300 \\ \hline
MSTAR (Label Noise) & EfficientNetV2s & 1.0e-3 & 1.0e-6 & 18 & 40 & 20 & 60 \\ \hline
MSTAR (Instance Noise) & EfficientNetV2s & 1.0e-3 & 1.0e-6 & 18 & 40 & 20 & 30 \\ \hline
PathMNIST+ & EfficientNetV2s & 1.0e-3 & 1.0e-5 & 16 & 10 & 20 & 256 \\ \hline

\end{tabular}}
\caption{Experimental parameters for the datasets in our experiments}
\label{tab:HyperparametersNoisyLabels}
\end{table}

\end{minipage}

\newpage
\section{Additional Experiments}

\noindent\hspace{0pt}\begin{minipage}{0.99\textwidth}
\RaggedRight

In this section, we present additional ablations and more fine-grained experimental data that support the results reported in the main text. Specifically, we look at the effect of Coreset size, report full results over all data sets for label flipping noise ratios of \{0.0, 0.1, 0.2, 0.3, 0.4, 0.5\}, and also include experiments for instance-level perturbations over varying amounts of training data on MNIST and MSTAR.

\end{minipage}

\subsection{Effect of Coreset Size}

\noindent\hspace{0pt}\begin{minipage}{0.99\textwidth}
\RaggedRight

In Fig. \ref{fig:mnist_coreset} we show the effect of Coreset size on the performance of our methods. We vary the Coreset size on MNIST \{100, 200, 300 samples per class\} at different levels of label flipping noise. For this dataset, Coreset size does not seem to play a large of a role w.r.t. accuracy and forgetting, except at the largest Coreset sizes there is a slight drop in performance in both accuracy and forgetting metrics. Further ablations on other datasets are warranted. We expect this is because Coreset purity does degrade somewhat with high levels of noise, so it may take larger Coresets to help overcome some of the added noise in the Coreset.

\begin{figure}[H]
  \centering
   \includegraphics[width=0.49\linewidth]{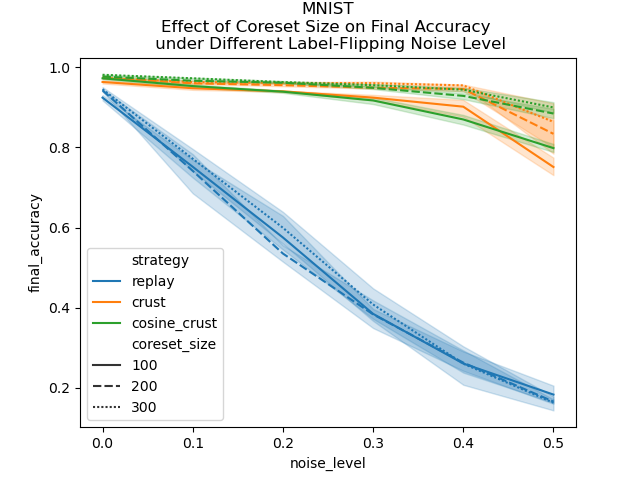}
   \includegraphics[width=0.49\linewidth]{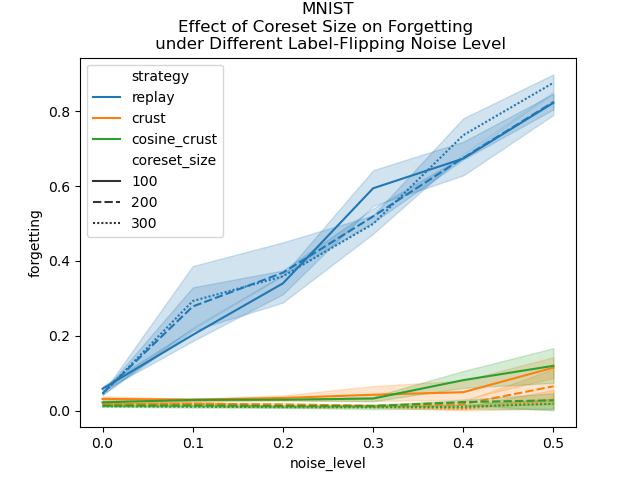}
   \caption{Looking at the final accuracy (left) and forgetting metric (right) at different Coreset sizes (100, 200, and 300 samples per class) for MNIST for random replay, Continual CRUST, and Continual CosineCRUST}
   \label{fig:mnist_coreset}
\end{figure}

\end{minipage}

\subsection{MNIST (Coreset Size=100): Label Noise}
\label{sec:MNIST100results}

\noindent\hspace{0pt}\begin{minipage}{0.99\textwidth}
\RaggedRight
In the prior section, we discussed an ablation using different sized Coresets on MNIST. In the next few sections, we report the fine-grained data associated with this experiment.  For these experiments, we considered per-class Coreset memory sizes of 100, 200, and 300, and we employed a simple Convolutional Neural Network (CNN) model. The architecture consists of two convolutional layers with 32 and 64 channels, respectively. Both layers utilize a kernel size of 3 and a stride of 1. Following the convolutional layers, the model incorporates two linear layers, with an intermediate dimension of 128.

\indent\hspace{0.5cm} Classes are shuffled prior to training. We initiated the first experience with two classes and subsequently added one class at a time for each future experience (a similar approach is used in all the experiments). We utilized the Adam optimizer and cross-entropy loss function, with a batch size of 32.

\indent\hspace{0.5cm} We evaluated three strategies: Continual CosineCRUST, Continual CRUST, and Random Replay. For a Coreset memory size of 100, the results are presented in Table \ref{tab:mem100MNIST}. These results represent the average and standard deviation across five different random seeds. The learning rates were set as follows: 0.0007 for Continual CosineCRUST and CRUST and 0.005 for Random Replay.

\indent\hspace{0.5cm} The training process for Continual CosineCRUST and Continual CRUST consisted of two main phases. First, we trained the models for 40 epochs using the full dataset for the current class along with all the data from previous classes. After this initial training, we continued with an additional 20 epochs, but this time, we only use the Coreset data from the current (updated after every epoch) and previous classes. This two-step approach allows us to first learn from all available data and then fine-tune the models using a smaller, more focused (potentially ``cleaner'') set of examples.

\begin{table}[H]
\centering
\resizebox{\textwidth}{!}{\begin{tabular}{c|cc|cc|cc|cc|cc|cc|}
\cline{2-13}
\textbf{} & \multicolumn{2}{c|}{\textbf{Noise=0.0}} & \multicolumn{2}{||c|}{\textbf{Noise=0.1}} & \multicolumn{2}{||c|}{\textbf{Noise=0.2}} & \multicolumn{2}{||c|}{\textbf{Noise=0.3}} & \multicolumn{2}{||c|}{\textbf{Noise=0.4}} & \multicolumn{2}{||c|}{\textbf{Noise=0.5}} \\ \hline
\multicolumn{1}{|c|}{\textbf{Algorithm}} & \multicolumn{1}{c|}{\textit{\textbf{Acc}}} & \textit{\textbf{Forg.}} & \multicolumn{1}{||c|}{\textit{\textbf{Acc}}} & \textit{\textbf{Forg.}} & \multicolumn{1}{||c|}{\textit{\textbf{Acc}}} & \textit{\textbf{Forg.}} & \multicolumn{1}{||c|}{\textit{\textbf{Acc}}} & \textit{\textbf{Forg.}} & \multicolumn{1}{||c|}
{\textit{\textbf{Acc}}} & \textit{\textbf{Forg.}} & \multicolumn{1}{||c|}{\textit{\textbf{Acc}}} & \textit{\textbf{Forg.}} \\ \hline
\multicolumn{1}{|c|}{\begin{tabular}[c]{@{}c@{}}Random \\ Replay\end{tabular}} & \multicolumn{1}{c|}{\begin{tabular}[c]{@{}c@{}}0.92\\ $\pm$0.01\end{tabular}} & \begin{tabular}[c]{@{}c@{}}0.06\\ $\pm$0.01\end{tabular} & \multicolumn{1}{||c|}{\begin{tabular}[c]{@{}c@{}}0.75\\ $\pm$0.03\end{tabular}} & \begin{tabular}[c]{@{}c@{}}0.2\\ $\pm$0.02\end{tabular} & \multicolumn{1}{||c|}{\begin{tabular}[c]{@{}c@{}}0.58\\ $\pm$0.05\end{tabular}} & \begin{tabular}[c]{@{}c@{}}0.34\\ $\pm$0.03\end{tabular} & \multicolumn{1}{||c|}{\begin{tabular}[c]{@{}c@{}}0.38\\ $\pm$0.01\end{tabular}} & \begin{tabular}[c]{@{}c@{}}0.59\\ $\pm$0.06\end{tabular} & \multicolumn{1}{||c|}{\begin{tabular}[c]{@{}c@{}}0.26\\ $\pm$0.03\end{tabular}} & \begin{tabular}[c]{@{}c@{}}0.67\\ $\pm$0.05\end{tabular} & \multicolumn{1}{||c|}{\begin{tabular}[c]{@{}c@{}}0.18\\ $\pm$0.02\end{tabular}} & \begin{tabular}[c]{@{}c@{}}0.82\\ $\pm$0.03\end{tabular} \\ \hline
\multicolumn{1}{|c|}{\begin{tabular}[c]{@{}c@{}}Continual \\ CRUST\end{tabular}} & \multicolumn{1}{c|}{\begin{tabular}[c]{@{}c@{}}0.96\\ $\pm$0.0\end{tabular}} & \begin{tabular}[c]{@{}c@{}}0.03\\ $\pm$0.01\end{tabular} & \multicolumn{1}{||c|}{\begin{tabular}[c]{@{}c@{}}0.95\\ $\pm$0.01\end{tabular}} & \begin{tabular}[c]{@{}c@{}}0.03\\ $\pm$0.0\end{tabular} & \multicolumn{1}{||c|}{\begin{tabular}[c]{@{}c@{}}0.94\\ $\pm$0.0\end{tabular}} & \begin{tabular}[c]{@{}c@{}}0.03\\ $\pm$0.01\end{tabular} & \multicolumn{1}{||c|}{\begin{tabular}[c]{@{}c@{}}0.92\\ $\pm$0.01\end{tabular}} & \begin{tabular}[c]{@{}c@{}}0.04\\ $\pm$0.02\end{tabular} & \multicolumn{1}{||c|}{\begin{tabular}[c]{@{}c@{}}0.9\\ $\pm$0.02\end{tabular}} & \begin{tabular}[c]{@{}c@{}}0.05\\ $\pm$0.03\end{tabular} & \multicolumn{1}{||c|}{\begin{tabular}[c]{@{}c@{}}0.75\\ $\pm$0.02\end{tabular}} & \begin{tabular}[c]{@{}c@{}}0.11\\ $\pm$0.03\end{tabular} \\ \hline
\multicolumn{1}{|c|}{\begin{tabular}[c]{@{}c@{}}Continual \\ CosineCRUST\end{tabular}} & \multicolumn{1}{c|}{\begin{tabular}[c]{@{}c@{}}0.97\\ $\pm$0.0\end{tabular}} & \begin{tabular}[c]{@{}c@{}}0.02\\ $\pm$0.01\end{tabular} & \multicolumn{1}{||c|}{\begin{tabular}[c]{@{}c@{}}0.95\\ $\pm$0.0\end{tabular}} & \begin{tabular}[c]{@{}c@{}}0.03\\ $\pm$0.0\end{tabular} & \multicolumn{1}{||c|}{\begin{tabular}[c]{@{}c@{}}0.94\\ $\pm$0.0\end{tabular}} & \begin{tabular}[c]{@{}c@{}}0.03\\ $\pm$0.0\end{tabular} & \multicolumn{1}{||c|}{\begin{tabular}[c]{@{}c@{}}0.92\\ $\pm$0.01\end{tabular}} & \begin{tabular}[c]{@{}c@{}}0.03\\ $\pm$0.01\end{tabular} & \multicolumn{1}{||c|}{\begin{tabular}[c]{@{}c@{}}0.87\\ $\pm$0.01\end{tabular}} & \begin{tabular}[c]{@{}c@{}}0.08\\ $\pm$0.03\end{tabular} & \multicolumn{1}{||c|}{\begin{tabular}[c]{@{}c@{}}0.8\\ $\pm$0.01\end{tabular}} & \begin{tabular}[c]{@{}c@{}}0.12\\ $\pm$0.05\end{tabular} \\ \hline
\end{tabular}}
\caption{Final evaluation accuracy and forgetting on MNIST w/ Coreset size of 100 for varying levels of label-flipping noise}
\label{tab:mem100MNIST}
\end{table}

\end{minipage}

\subsection{MNIST (Coreset Size=200): Label Noise}

\noindent\hspace{0pt}\begin{minipage}{0.99\textwidth}
\RaggedRight
\indent\hspace{0.5cm} In our experiments with the MNIST dataset and a Coreset memory of 200, we used the Adam\cite{kingma2014adam} optimizer with a batch size of 32. The learning rates were set to 0.001 for both Continual CosineCRUST and Continual CRUST, while Random Replay used a higher learning rate of 0.007.

\indent\hspace{0.5cm} As mentioned in Section \ref{sec:MNIST100results}, when training Continual CosineCRUST and Continual CRUST, the first phase of training is conducted for 40 epochs and then the second phase is carried out for 20 epochs.

\indent\hspace{0.5cm} Table \ref{tab:mem200MNIST} presents a comparative analysis of the three methods. The results indicate that Continual CRUST and Continual CosineCRUST exhibit similar performance, with Continual CRUST shows a slight edge. Both of these methods consistently outperform Random Replay across all mislabel ratios. Notably, as the mislabel ratio increases, the performance gap between Continual CosineCRUST/CRUST and Random Replay widens significantly. This trend underscores the efficacy of the sample selection strategies employed by Continual CosineCRUST and Continual CRUST in constructing a robust Coreset, particularly in scenarios with higher levels of label noise.

\begin{table}[H]
\centering
\resizebox{\textwidth}{!}{\begin{tabular}{c|cc|cc|cc|cc|cc|cc|}
\cline{2-13}
\textbf{} & \multicolumn{2}{c|}{\textbf{Noise=0.0}} & \multicolumn{2}{||c|}{\textbf{Noise=0.1}} & \multicolumn{2}{||c|}{\textbf{Noise=0.2}} & \multicolumn{2}{||c|}{\textbf{Noise=0.3}} & \multicolumn{2}{||c|}{\textbf{Noise=0.4}} & \multicolumn{2}{||c|}{\textbf{Noise=0.5}} \\ \hline
\multicolumn{1}{|c|}{\textbf{Algorithm}} & \multicolumn{1}{c|}{\textit{\textbf{Acc}}} & \textit{\textbf{Forg.}} & \multicolumn{1}{||c|}{\textit{\textbf{Acc}}} & \textit{\textbf{Forg.}} & \multicolumn{1}{||c|}{\textit{\textbf{Acc}}} & \textit{\textbf{Forg.}} & \multicolumn{1}{||c|}{\textit{\textbf{Acc}}} & \textit{\textbf{Forg.}} & \multicolumn{1}{||c|}
{\textit{\textbf{Acc}}} & \textit{\textbf{Forg.}} & \multicolumn{1}{||c|}{\textit{\textbf{Acc}}} & \textit{\textbf{Forg.}} \\ \hline
\multicolumn{1}{|c|}{\begin{tabular}[c]{@{}c@{}}Random \\ Replay\end{tabular}} & \multicolumn{1}{c|}{\begin{tabular}[c]{@{}c@{}}0.94\\ $\pm$0.0\end{tabular}} & \begin{tabular}[c]{@{}c@{}}0.05\\ $\pm$0.0\end{tabular} & \multicolumn{1}{||c|}{\begin{tabular}[c]{@{}c@{}}0.74\\ $\pm$0.06\end{tabular}} & \begin{tabular}[c]{@{}c@{}}0.28\\ $\pm$0.06\end{tabular} & \multicolumn{1}{||c|}{\begin{tabular}[c]{@{}c@{}}0.53\\ $\pm$0.02\end{tabular}} & \begin{tabular}[c]{@{}c@{}}0.37\\ $\pm$0.01\end{tabular} & \multicolumn{1}{||c|}{\begin{tabular}[c]{@{}c@{}}0.38\\ $\pm$0.04\end{tabular}} & \begin{tabular}[c]{@{}c@{}}0.52\\ $\pm$0.02\end{tabular} & \multicolumn{1}{||c|}{\begin{tabular}[c]{@{}c@{}}0.26\\ $\pm$0.03\end{tabular}} & \begin{tabular}[c]{@{}c@{}}0.68\\ $\pm$0.01\end{tabular} & \multicolumn{1}{||c|}{\begin{tabular}[c]{@{}c@{}}0.17\\ $\pm$0.01\end{tabular}} & \begin{tabular}[c]{@{}c@{}}0.83\\ $\pm$0.02\end{tabular} \\ \hline
\multicolumn{1}{|c|}{\begin{tabular}[c]{@{}c@{}}Continual \\ CRUST\end{tabular}} & \multicolumn{1}{c|}{\begin{tabular}[c]{@{}c@{}}0.97\\ $\pm$0.0\end{tabular}} & \begin{tabular}[c]{@{}c@{}}0.02\\ $\pm$0.0\end{tabular} & \multicolumn{1}{||c|}{\begin{tabular}[c]{@{}c@{}}0.96\\ $\pm$0.0\end{tabular}} & \begin{tabular}[c]{@{}c@{}}0.02\\ $\pm$0.0\end{tabular} & \multicolumn{1}{||c|}{\begin{tabular}[c]{@{}c@{}}0.96\\ $\pm$0.0\end{tabular}} & \begin{tabular}[c]{@{}c@{}}0.02\\ $\pm$0.0\end{tabular} & \multicolumn{1}{||c|}{\begin{tabular}[c]{@{}c@{}}0.95\\ $\pm$0.0\end{tabular}} & \begin{tabular}[c]{@{}c@{}}0.01\\ $\pm$0.0\end{tabular} & \multicolumn{1}{||c|}{\begin{tabular}[c]{@{}c@{}}0.95\\ $\pm$0.01\end{tabular}} & \begin{tabular}[c]{@{}c@{}}0.02\\ $\pm$0.01\end{tabular} & \multicolumn{1}{||c|}{\begin{tabular}[c]{@{}c@{}}0.83\\ $\pm$0.04\end{tabular}} & \begin{tabular}[c]{@{}c@{}}0.06\\ $\pm$0.05\end{tabular} \\ \hline
\multicolumn{1}{|c|}{\begin{tabular}[c]{@{}c@{}}Continual \\ CosineCRUST\end{tabular}} & \multicolumn{1}{c|}{\begin{tabular}[c]{@{}c@{}}0.98\\ $\pm$0.0\end{tabular}} & \begin{tabular}[c]{@{}c@{}}0.01\\ $\pm$0.0\end{tabular} & \multicolumn{1}{||c|}{\begin{tabular}[c]{@{}c@{}}0.97\\ $\pm$0.0\end{tabular}} & \begin{tabular}[c]{@{}c@{}}0.01\\ $\pm$0.0\end{tabular} & \multicolumn{1}{||c|}{\begin{tabular}[c]{@{}c@{}}0.96\\ $\pm$0.0\end{tabular}} & \begin{tabular}[c]{@{}c@{}}0.01\\ $\pm$0.0\end{tabular} & \multicolumn{1}{||c|}{\begin{tabular}[c]{@{}c@{}}0.95\\ $\pm$0.0\end{tabular}} & \begin{tabular}[c]{@{}c@{}}0.01\\ $\pm$0.0\end{tabular} & \multicolumn{1}{||c|}{\begin{tabular}[c]{@{}c@{}}0.93\\ $\pm$0.01\end{tabular}} & \begin{tabular}[c]{@{}c@{}}0.02\\ $\pm$0.01\end{tabular} & \multicolumn{1}{||c|}{\begin{tabular}[c]{@{}c@{}}0.88\\ $\pm$0.01\end{tabular}} & \begin{tabular}[c]{@{}c@{}}0.03\\ $\pm$0.03\end{tabular} \\ \hline
\end{tabular}}
\caption{Final evaluation accuracy and forgetting on MNIST w/ Coreset size of 200 for varying levels of label-flipping noise}
\label{tab:mem200MNIST}
\end{table}

\end{minipage}

\subsection{MNIST (Coreset Size=300): Label Noise}
\noindent\hspace{0pt}\begin{minipage}{0.99\textwidth}
\RaggedRight
\indent\hspace{0.5cm} Following similar procedure to the two prior sections, we conducted experiments with various levels of label noise on the MNIST dataset with a Coreset size of 300 samples per class. Note that the following results are also the results reported/compared against for the MNIST experiments included in the main text.

\begin{table}[H]
\centering
\resizebox{\textwidth}{!}{\begin{tabular}{c|cc|cc|cc|cc|cc|cc|}
\cline{2-13}
\textbf{} & \multicolumn{2}{c|}{\textbf{Noise=0.0}} & \multicolumn{2}{||c|}{\textbf{Noise=0.1}} & \multicolumn{2}{||c|}{\textbf{Noise=0.2}} & \multicolumn{2}{||c|}{\textbf{Noise=0.3}} & \multicolumn{2}{||c|}{\textbf{Noise=0.4}} & \multicolumn{2}{||c|}{\textbf{Noise=0.5}} \\ \hline
\multicolumn{1}{|c|}{\textbf{Algorithm}} & \multicolumn{1}{c|}{\textit{\textbf{Acc}}} & \textit{\textbf{Forg.}} & \multicolumn{1}{||c|}{\textit{\textbf{Acc}}} & \textit{\textbf{Forg.}} & \multicolumn{1}{||c|}{\textit{\textbf{Acc}}} & \textit{\textbf{Forg.}} & \multicolumn{1}{||c|}{\textit{\textbf{Acc}}} & \textit{\textbf{Forg.}} & \multicolumn{1}{||c|}
{\textit{\textbf{Acc}}} & \textit{\textbf{Forg.}} & \multicolumn{1}{||c|}{\textit{\textbf{Acc}}} & \textit{\textbf{Forg.}} \\ \hline
\multicolumn{1}{|c|}{\begin{tabular}[c]{@{}c@{}}Naive \\ Sequential \\ Learner\end{tabular}} & \multicolumn{1}{c|}{\begin{tabular}[c]{@{}c@{}}0.11\\ $\pm$0.0\end{tabular}} & \begin{tabular}[c]{@{}c@{}}1.0\\ $\pm$0.0\end{tabular} & \multicolumn{1}{||c|}{\begin{tabular}[c]{@{}c@{}}0.11\\ $\pm$0.0\end{tabular}} & \begin{tabular}[c]{@{}c@{}}1.0\\ $\pm$0.0\end{tabular} & \multicolumn{1}{||c|}{\begin{tabular}[c]{@{}c@{}}0.11\\ $\pm$0.0\end{tabular}} & \begin{tabular}[c]{@{}c@{}}1.0\\ $\pm$0.0\end{tabular} & \multicolumn{1}{||c|}{\begin{tabular}[c]{@{}c@{}}0.11\\ $\pm$0.0\end{tabular}} & \begin{tabular}[c]{@{}c@{}}1.0\\ $\pm$0.0\end{tabular} & \multicolumn{1}{||c|}{\begin{tabular}[c]{@{}c@{}}0.11\\ $\pm$0.0\end{tabular}} & \begin{tabular}[c]{@{}c@{}}1.0\\ $\pm$0.0\end{tabular} & \multicolumn{1}{||c|}{\begin{tabular}[c]{@{}c@{}}0.11\\ $\pm$0.0\end{tabular}} & \begin{tabular}[c]{@{}c@{}}1.0\\ $\pm$0.0\end{tabular} \\ \hline
\multicolumn{1}{|c|}{\begin{tabular}[c]{@{}c@{}}Joint \\ Learner\end{tabular}} & \multicolumn{1}{c|}{\begin{tabular}[c]{@{}c@{}}0.99\\ $\pm$0.0\end{tabular}} & \begin{tabular}[c]{@{}c@{}} N/A \end{tabular} & \multicolumn{1}{||c|}{\begin{tabular}[c]{@{}c@{}}0.99\\ $\pm$0.0\end{tabular}} & \begin{tabular}[c]{@{}c@{}} N/A \end{tabular} & \multicolumn{1}{||c|}{\begin{tabular}[c]{@{}c@{}}0.98\\ $\pm$0.0\end{tabular}} & \begin{tabular}[c]{@{}c@{}} N/A \end{tabular} & \multicolumn{1}{||c|}{\begin{tabular}[c]{@{}c@{}}0.98\\ $\pm$0.0\end{tabular}} & \begin{tabular}[c]{@{}c@{}} N/A \end{tabular} & \multicolumn{1}{||c|}{\begin{tabular}[c]{@{}c@{}}0.98\\ $\pm$0.0\end{tabular}} & \begin{tabular}[c]{@{}c@{}} N/A \end{tabular} & \multicolumn{1}{||c|}{\begin{tabular}[c]{@{}c@{}}0.98\\ $\pm$0.0\end{tabular}} & \begin{tabular}[c]{@{}c@{}} N/A \end{tabular} \\ \hline
\multicolumn{1}{|c|}{\begin{tabular}[c]{@{}c@{}}Cumulative \\ Learner\end{tabular}} & \multicolumn{1}{c|}{\begin{tabular}[c]{@{}c@{}}0.99\\ $\pm$0.0\end{tabular}} & \begin{tabular}[c]{@{}c@{}}0.0\\ $\pm$0.0\end{tabular} & \multicolumn{1}{||c|}{\begin{tabular}[c]{@{}c@{}}0.99\\ $\pm$0.0\end{tabular}} & \begin{tabular}[c]{@{}c@{}}0.0\\ $\pm$0.0\end{tabular} & \multicolumn{1}{||c|}{\begin{tabular}[c]{@{}c@{}}0.98\\ $\pm$0.0\end{tabular}} & \begin{tabular}[c]{@{}c@{}}0.01\\ $\pm$0.0\end{tabular} & \multicolumn{1}{||c|}{\begin{tabular}[c]{@{}c@{}}0.98\\ $\pm$0.0\end{tabular}} & \begin{tabular}[c]{@{}c@{}}0.01\\ $\pm$0.0\end{tabular} & \multicolumn{1}{||c|}{\begin{tabular}[c]{@{}c@{}}0.98\\ $\pm$0.0\end{tabular}} & \begin{tabular}[c]{@{}c@{}}0.01\\ $\pm$0.0\end{tabular} & \multicolumn{1}{||c|}{\begin{tabular}[c]{@{}c@{}}0.97\\ $\pm$0.0\end{tabular}} & \begin{tabular}[c]{@{}c@{}}0.01\\ $\pm$0.0\end{tabular} \\ \hline
\multicolumn{1}{|c|}{\begin{tabular}[c]{@{}c@{}}Random \\ Replay\end{tabular}} & \multicolumn{1}{c|}{\begin{tabular}[c]{@{}c@{}}0.95\\ $\pm$0.01\end{tabular}} & \begin{tabular}[c]{@{}c@{}}0.04\\ $\pm$0.0\end{tabular} & \multicolumn{1}{||c|}{\begin{tabular}[c]{@{}c@{}}0.77\\ $\pm$0.03\end{tabular}} & \begin{tabular}[c]{@{}c@{}}0.29\\ $\pm$0.1\end{tabular} & \multicolumn{1}{||c|}{\begin{tabular}[c]{@{}c@{}}0.6\\ $\pm$0.05\end{tabular}} & \begin{tabular}[c]{@{}c@{}}0.36\\ $\pm$0.09\end{tabular} & \multicolumn{1}{||c|}{\begin{tabular}[c]{@{}c@{}}0.41\\ $\pm$0.05\end{tabular}} & \begin{tabular}[c]{@{}c@{}}0.5\\ $\pm$0.03\end{tabular} & \multicolumn{1}{||c|}{\begin{tabular}[c]{@{}c@{}}0.26\\ $\pm$0.06\end{tabular}} & \begin{tabular}[c]{@{}c@{}}0.74\\ $\pm$0.06\end{tabular} & \multicolumn{1}{||c|}{\begin{tabular}[c]{@{}c@{}}0.16\\ $\pm$0.02\end{tabular}} & \begin{tabular}[c]{@{}c@{}}0.88\\ $\pm$0.03\end{tabular} \\ \hline
\multicolumn{1}{|c|}{\begin{tabular}[c]{@{}c@{}}Random \\ Replay \\ w/EWC\end{tabular}} & \multicolumn{1}{c|}{\begin{tabular}[c]{@{}c@{}}0.97\\ $\pm$0.01\end{tabular}} & \begin{tabular}[c]{@{}c@{}}0.03\\ $\pm$0.01\end{tabular} & \multicolumn{1}{||c|}{\begin{tabular}[c]{@{}c@{}}0.94\\ $\pm$0.01\end{tabular}} & \begin{tabular}[c]{@{}c@{}}0.06\\ $\pm$0.01\end{tabular} & \multicolumn{1}{||c|}{\begin{tabular}[c]{@{}c@{}}0.85\\ $\pm$0.02\end{tabular}} & \begin{tabular}[c]{@{}c@{}}0.13\\ $\pm$0.01\end{tabular} & \multicolumn{1}{||c|}{\begin{tabular}[c]{@{}c@{}}0.72\\ $\pm$0.03\end{tabular}} & \begin{tabular}[c]{@{}c@{}}0.25\\ $\pm$0.02\end{tabular} & \multicolumn{1}{||c|}{\begin{tabular}[c]{@{}c@{}}0.53\\ $\pm$0.03\end{tabular}} & \begin{tabular}[c]{@{}c@{}}0.42\\ $\pm$0.02\end{tabular} & \multicolumn{1}{||c|}{\begin{tabular}[c]{@{}c@{}}0.3\\ $\pm$0.02\end{tabular}} & \begin{tabular}[c]{@{}c@{}}0.63\\ $\pm$0.02\end{tabular} \\ \hline
\multicolumn{1}{|c|}{\begin{tabular}[c]{@{}c@{}}Dark ER\end{tabular}} & \multicolumn{1}{c|}{\begin{tabular}[c]{@{}c@{}}0.97\\ $\pm$0.01\end{tabular}} & \begin{tabular}[c]{@{}c@{}}0.02\\ $\pm$0.0\end{tabular} & \multicolumn{1}{||c|}{\begin{tabular}[c]{@{}c@{}}0.93\\ $\pm$0.01\end{tabular}} & \begin{tabular}[c]{@{}c@{}}0.05\\ $\pm$0.0\end{tabular} & \multicolumn{1}{||c|}{\begin{tabular}[c]{@{}c@{}}0.85\\ $\pm$0.01\end{tabular}} & \begin{tabular}[c]{@{}c@{}}0.11\\ $\pm$0.01\end{tabular} & \multicolumn{1}{||c|}{\begin{tabular}[c]{@{}c@{}}0.79\\ $\pm$0.02\end{tabular}} & \begin{tabular}[c]{@{}c@{}}0.17\\ $\pm$0.01\end{tabular} & \multicolumn{1}{||c|}{\begin{tabular}[c]{@{}c@{}}0.61\\ $\pm$0.02\end{tabular}} & \begin{tabular}[c]{@{}c@{}}0.3\\ $\pm$0.01\end{tabular} & \multicolumn{1}{||c|}{\begin{tabular}[c]{@{}c@{}}0.44\\ $\pm$0.02\end{tabular}} & \begin{tabular}[c]{@{}c@{}}0.46\\ $\pm$0.02\end{tabular} \\ \hline
\multicolumn{1}{|c|}{\begin{tabular}[c]{@{}c@{}}iCaRL\end{tabular}} & \multicolumn{1}{c|}{\begin{tabular}[c]{@{}c@{}}0.94\\ $\pm$0.01\end{tabular}} & \begin{tabular}[c]{@{}c@{}}-0.09\\ $\pm$0.06\end{tabular} & \multicolumn{1}{||c|}{\begin{tabular}[c]{@{}c@{}}0.93\\ $\pm$0.01\end{tabular}} & \begin{tabular}[c]{@{}c@{}}0.02\\ $\pm$0.0\end{tabular} & \multicolumn{1}{||c|}{\begin{tabular}[c]{@{}c@{}}0.89\\ $\pm$0.01\end{tabular}} & \begin{tabular}[c]{@{}c@{}}0.05\\ $\pm$0.02\end{tabular} & \multicolumn{1}{||c|}{\begin{tabular}[c]{@{}c@{}}0.77\\ $\pm$0.01\end{tabular}} & \begin{tabular}[c]{@{}c@{}}0.14\\ $\pm$0.02\end{tabular} & \multicolumn{1}{||c|}{\begin{tabular}[c]{@{}c@{}}0.55\\ $\pm$0.02\end{tabular}} & \begin{tabular}[c]{@{}c@{}}0.27\\ $\pm$0.04\end{tabular} & \multicolumn{1}{||c|}{\begin{tabular}[c]{@{}c@{}}0.3\\ $\pm$0.15\end{tabular}} & \begin{tabular}[c]{@{}c@{}}0.41\\ $\pm$0.04\end{tabular} \\ \hline
\multicolumn{1}{|c|}{\begin{tabular}[c]{@{}c@{}}Continual \\ CRUST\end{tabular}} & \multicolumn{1}{c|}{\begin{tabular}[c]{@{}c@{}}0.97\\ $\pm$0.0\end{tabular}} & \begin{tabular}[c]{@{}c@{}}0.02\\ $\pm$0.0\end{tabular} & \multicolumn{1}{||c|}{\begin{tabular}[c]{@{}c@{}}0.96\\ $\pm$0.0\end{tabular}} & \begin{tabular}[c]{@{}c@{}}0.02\\ $\pm$0.01\end{tabular} & \multicolumn{1}{||c|}{\begin{tabular}[c]{@{}c@{}}0.96\\ $\pm$0.0\end{tabular}} & \begin{tabular}[c]{@{}c@{}}0.01\\ $\pm$0.0\end{tabular} & \multicolumn{1}{||c|}{\begin{tabular}[c]{@{}c@{}}0.96\\ $\pm$0.0\end{tabular}} & \begin{tabular}[c]{@{}c@{}}0.01\\ $\pm$0.0\end{tabular} & \multicolumn{1}{||c|}{\begin{tabular}[c]{@{}c@{}}0.95\\ $\pm$0.0\end{tabular}} & \begin{tabular}[c]{@{}c@{}}0.0\\ $\pm$0.01\end{tabular} & \multicolumn{1}{||c|}{\begin{tabular}[c]{@{}c@{}}0.86\\ $\pm$0.08\end{tabular}} & \begin{tabular}[c]{@{}c@{}}0.03\\ $\pm$0.03\end{tabular} \\ \hline
\multicolumn{1}{|c|}{\begin{tabular}[c]{@{}c@{}}Continual \\ CosineCRUST\end{tabular}} & \multicolumn{1}{c|}{\begin{tabular}[c]{@{}c@{}}0.98\\ $\pm$0.0\end{tabular}} & \begin{tabular}[c]{@{}c@{}}0.01\\ $\pm$0.0\end{tabular} & \multicolumn{1}{||c|}{\begin{tabular}[c]{@{}c@{}}0.97\\ $\pm$0.0\end{tabular}} & \begin{tabular}[c]{@{}c@{}}0.01\\ $\pm$0.0\end{tabular} & \multicolumn{1}{||c|}{\begin{tabular}[c]{@{}c@{}}0.96\\ $\pm$0.0\end{tabular}} & \begin{tabular}[c]{@{}c@{}}0.01\\ $\pm$0.0\end{tabular} & \multicolumn{1}{||c|}{\begin{tabular}[c]{@{}c@{}}0.96\\ $\pm$0.0\end{tabular}} & \begin{tabular}[c]{@{}c@{}}0.01\\ $\pm$0.01\end{tabular} & \multicolumn{1}{||c|}{\begin{tabular}[c]{@{}c@{}}0.94\\ $\pm$0.01\end{tabular}} & \begin{tabular}[c]{@{}c@{}}0.01\\ $\pm$0.01\end{tabular} & \multicolumn{1}{||c|}{\begin{tabular}[c]{@{}c@{}}0.9\\ $\pm$0.02\end{tabular}} & \begin{tabular}[c]{@{}c@{}}0.02\\ $\pm$0.02\end{tabular} \\ \hline
\end{tabular}}
\caption{Final evaluation accuracy and forgetting on MNIST w/ Coreset size of 300 for varying levels of label-flipping noise}
\label{tab:mem300MNIST}
\end{table}

\indent\hspace{0.5cm} Table \ref{tab:mem300MNIST} presents the results for MNIST with a Coreset memory of 300. The performance of Naive Sequential Learner, Joint Learner, and Cumulative Learner are included as lower and upper bounds for comparison. The data reveals that Continual CosineCRUST and Continual CRUST consistently outperform other strategies, including Random Replay, Random Replay with EWC, Dark ER, and iCaRL. This performance advantage becomes particularly pronounced as the mislabel ratio increases. Notably, even at a high mislabel ratio of 0.5, Continual CosineCRUST and Continual CRUST maintain impressive accuracies of 0.9 and 0.86, respectively. These results further demonstrate the effectiveness of our proposed strategies in selecting representative Coresets for each class.

\end{minipage}

\begin{figure}[H]
  \centering
   \includegraphics[width=0.45\linewidth]{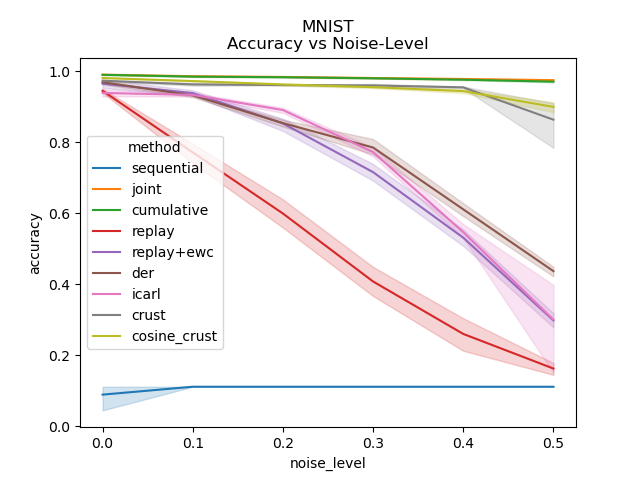}
   \includegraphics[width=0.45\linewidth]{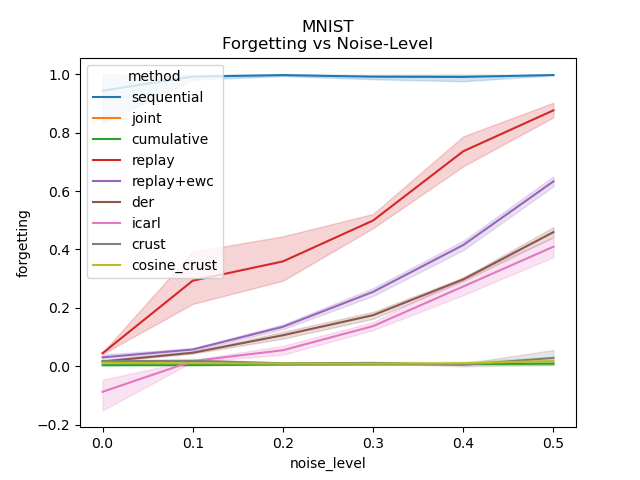}
   \caption{Looking at the final accuracy (left) and forgetting metric (right) at different label flipping noise levels (0.0-0.5) for MNIST for different strategies}
   \label{fig:mnist_noise_supp}
\end{figure}

\newpage
\subsection{FashionMNIST: Label Noise}

\noindent\hspace{0pt}\begin{minipage}{0.99\textwidth}
\RaggedRight
Next we consider experiments run on the FashionMNIST dataset using a Coreset memory of 300. We employ a similar set up as the previously discussed MNIST experiments, varying the mislabel ratio from 0.0 to 0.5 by increments of 0.1. Table \ref{tab:mem300FashionMNIST} presents the results obtained from five different random seeds, reporting both the average performance and standard deviation. As in previous experiments, we include Naive Sequential Learner, Cumulative Learner, and Joint Learner as lower and upper performance bounds for comparison and compared against prior work in the form of Random Replay, Random Replay with EWC, Dark Experience Replay, and iCaRL. 

\begin{table}[H]
\centering
\resizebox{\textwidth}{!}{\begin{tabular}{c|cc|cc|cc|cc|cc|cc|}
\cline{2-13}
\textbf{} & \multicolumn{2}{c|}{\textbf{Noise=0.0}} & \multicolumn{2}{||c|}{\textbf{Noise=0.1}} & \multicolumn{2}{||c|}{\textbf{Noise=0.2}} & \multicolumn{2}{||c|}{\textbf{Noise=0.3}} & \multicolumn{2}{||c|}{\textbf{Noise=0.4}} & \multicolumn{2}{||c|}{\textbf{Noise=0.5}} \\ \hline
\multicolumn{1}{|c|}{\textbf{Algorithm}} & \multicolumn{1}{c|}{\textit{\textbf{Acc}}} & \textit{\textbf{Forg.}} & \multicolumn{1}{||c|}{\textit{\textbf{Acc}}} & \textit{\textbf{Forg.}} & \multicolumn{1}{||c|}{\textit{\textbf{Acc}}} & \textit{\textbf{Forg.}} & \multicolumn{1}{||c|}{\textit{\textbf{Acc}}} & \textit{\textbf{Forg.}} & \multicolumn{1}{||c|}
{\textit{\textbf{Acc}}} & \textit{\textbf{Forg.}} & \multicolumn{1}{||c|}{\textit{\textbf{Acc}}} & \textit{\textbf{Forg.}} \\ \hline
\multicolumn{1}{|c|}{\begin{tabular}[c]{@{}c@{}}Naive \\ Sequential \\ Learner\end{tabular}} & \multicolumn{1}{c|}{\begin{tabular}[c]{@{}c@{}}0.09\\ $\pm$0.04\end{tabular}} & \begin{tabular}[c]{@{}c@{}}0.94\\ $\pm$0.11\end{tabular} & \multicolumn{1}{||c|}{\begin{tabular}[c]{@{}c@{}}0.11\\ $\pm$0.0\end{tabular}} & \begin{tabular}[c]{@{}c@{}}0.99\\ $\pm$0.01\end{tabular} & \multicolumn{1}{||c|}{\begin{tabular}[c]{@{}c@{}}0.11\\ $\pm$0.0\end{tabular}} & \begin{tabular}[c]{@{}c@{}}1.0\\ $\pm$0.0\end{tabular} & \multicolumn{1}{||c|}{\begin{tabular}[c]{@{}c@{}}0.11\\ $\pm$0.0\end{tabular}} & \begin{tabular}[c]{@{}c@{}}0.99\\ $\pm$0.01\end{tabular} & \multicolumn{1}{||c|}{\begin{tabular}[c]{@{}c@{}}0.11\\ $\pm$0.0\end{tabular}} & \begin{tabular}[c]{@{}c@{}}0.99\\ $\pm$0.01\end{tabular} & \multicolumn{1}{||c|}{\begin{tabular}[c]{@{}c@{}}0.11\\ $\pm$0.0\end{tabular}} & \begin{tabular}[c]{@{}c@{}}1.0\\ $\pm$0.0\end{tabular} \\ \hline
\multicolumn{1}{|c|}{\begin{tabular}[c]{@{}c@{}}Joint \\ Learner\end{tabular}} & \multicolumn{1}{c|}{\begin{tabular}[c]{@{}c@{}}0.94\\ $\pm$0.01\end{tabular}} & \begin{tabular}[c]{@{}c@{}} N/A \end{tabular} & \multicolumn{1}{||c|}{\begin{tabular}[c]{@{}c@{}}0.91\\ $\pm$0.01\end{tabular}} & \begin{tabular}[c]{@{}c@{}} N/A \end{tabular} & \multicolumn{1}{||c|}{\begin{tabular}[c]{@{}c@{}}0.9\\ $\pm$0.01\end{tabular}} & \begin{tabular}[c]{@{}c@{}} N/A \end{tabular} & \multicolumn{1}{||c|}{\begin{tabular}[c]{@{}c@{}}0.89\\ $\pm$0.01\end{tabular}} & \begin{tabular}[c]{@{}c@{}} N/A \end{tabular} & \multicolumn{1}{||c|}{\begin{tabular}[c]{@{}c@{}}0.88\\ $\pm$0.01\end{tabular}} & \begin{tabular}[c]{@{}c@{}} N/A \end{tabular} & \multicolumn{1}{||c|}{\begin{tabular}[c]{@{}c@{}}0.87\\ $\pm$0.01\end{tabular}} & \begin{tabular}[c]{@{}c@{}} N/A \end{tabular} \\ \hline
\multicolumn{1}{|c|}{\begin{tabular}[c]{@{}c@{}}Cumulative \\ Learner\end{tabular}} & \multicolumn{1}{c|}{\begin{tabular}[c]{@{}c@{}}0.93\\ $\pm$0.01\end{tabular}} & \begin{tabular}[c]{@{}c@{}}0.02\\ $\pm$0.01\end{tabular} & \multicolumn{1}{||c|}{\begin{tabular}[c]{@{}c@{}}0.9\\ $\pm$0.01\end{tabular}} & \begin{tabular}[c]{@{}c@{}}0.03\\ $\pm$0.01\end{tabular} & \multicolumn{1}{||c|}{\begin{tabular}[c]{@{}c@{}}0.89\\ $\pm$0.01\end{tabular}} & \begin{tabular}[c]{@{}c@{}}0.03\\ $\pm$0.01\end{tabular} & \multicolumn{1}{||c|}{\begin{tabular}[c]{@{}c@{}}0.89\\ $\pm$0.01\end{tabular}} & \begin{tabular}[c]{@{}c@{}}0.04\\ $\pm$0.02\end{tabular} & \multicolumn{1}{||c|}{\begin{tabular}[c]{@{}c@{}}0.87\\ $\pm$0.01\end{tabular}} & \begin{tabular}[c]{@{}c@{}}0.06\\ $\pm$0.04\end{tabular} & \multicolumn{1}{||c|}{\begin{tabular}[c]{@{}c@{}}0.87\\ $\pm$0.0\end{tabular}} & \begin{tabular}[c]{@{}c@{}}0.04\\ $\pm$0.02\end{tabular} \\ \hline
\multicolumn{1}{|c|}{\begin{tabular}[c]{@{}c@{}}Random \\ Replay\end{tabular}} & \multicolumn{1}{c|}{\begin{tabular}[c]{@{}c@{}}0.75\\ $\pm$0.02\end{tabular}} & \begin{tabular}[c]{@{}c@{}}0.26\\ $\pm$0.03\end{tabular} & \multicolumn{1}{||c|}{\begin{tabular}[c]{@{}c@{}}0.42\\ $\pm$0.02\end{tabular}} & \begin{tabular}[c]{@{}c@{}}0.5\\ $\pm$0.03\end{tabular} & \multicolumn{1}{||c|}{\begin{tabular}[c]{@{}c@{}}0.27\\ $\pm$0.04\end{tabular}} & \begin{tabular}[c]{@{}c@{}}0.64\\ $\pm$0.1\end{tabular} & \multicolumn{1}{||c|}{\begin{tabular}[c]{@{}c@{}}0.2\\ $\pm$0.02\end{tabular}} & \begin{tabular}[c]{@{}c@{}}0.79\\ $\pm$0.05\end{tabular} & \multicolumn{1}{||c|}{\begin{tabular}[c]{@{}c@{}}0.18\\ $\pm$0.01\end{tabular}} & \begin{tabular}[c]{@{}c@{}}0.81\\ $\pm$0.02\end{tabular} & \multicolumn{1}{||c|}{\begin{tabular}[c]{@{}c@{}}0.17\\ $\pm$0.03\end{tabular}} & \begin{tabular}[c]{@{}c@{}}0.79\\ $\pm$0.03\end{tabular} \\ \hline
\multicolumn{1}{|c|}{\begin{tabular}[c]{@{}c@{}}Random \\ Replay \\ w/EWC\end{tabular}} & \multicolumn{1}{c|}{\begin{tabular}[c]{@{}c@{}}0.82\\ $\pm$0.02\end{tabular}} & \begin{tabular}[c]{@{}c@{}}0.19\\ $\pm$0.01\end{tabular} & \multicolumn{1}{||c|}{\begin{tabular}[c]{@{}c@{}}0.65\\ $\pm$0.04\end{tabular}} & \begin{tabular}[c]{@{}c@{}}0.29\\ $\pm$0.01\end{tabular} & \multicolumn{1}{||c|}{\begin{tabular}[c]{@{}c@{}}0.5\\ $\pm$0.03\end{tabular}} & \begin{tabular}[c]{@{}c@{}}0.41\\ $\pm$0.05\end{tabular} & \multicolumn{1}{||c|}{\begin{tabular}[c]{@{}c@{}}0.35\\ $\pm$0.03\end{tabular}} & \begin{tabular}[c]{@{}c@{}}0.57\\ $\pm$0.05\end{tabular} & \multicolumn{1}{||c|}{\begin{tabular}[c]{@{}c@{}}0.33\\ $\pm$0.05\end{tabular}} & \begin{tabular}[c]{@{}c@{}}0.61\\ $\pm$0.06\end{tabular} & \multicolumn{1}{||c|}{\begin{tabular}[c]{@{}c@{}}0.27\\ $\pm$0.02\end{tabular}} & \begin{tabular}[c]{@{}c@{}}0.63\\ $\pm$0.04\end{tabular} \\ \hline
\multicolumn{1}{|c|}{\begin{tabular}[c]{@{}c@{}}Dark ER\end{tabular}} & \multicolumn{1}{c|}{\begin{tabular}[c]{@{}c@{}}0.85\\ $\pm$0.02\end{tabular}} & \begin{tabular}[c]{@{}c@{}}0.12\\ $\pm$0.01\end{tabular} & \multicolumn{1}{||c|}{\begin{tabular}[c]{@{}c@{}}0.77\\ $\pm$0.01\end{tabular}} & \begin{tabular}[c]{@{}c@{}}0.23\\ $\pm$0.03\end{tabular} & \multicolumn{1}{||c|}{\begin{tabular}[c]{@{}c@{}}0.69\\ $\pm$0.02\end{tabular}} & \begin{tabular}[c]{@{}c@{}}0.29\\ $\pm$0.05\end{tabular} & \multicolumn{1}{||c|}{\begin{tabular}[c]{@{}c@{}}0.55\\ $\pm$0.03\end{tabular}} & \begin{tabular}[c]{@{}c@{}}0.41\\ $\pm$0.02\end{tabular} & \multicolumn{1}{||c|}{\begin{tabular}[c]{@{}c@{}}0.43\\ $\pm$0.02\end{tabular}} & \begin{tabular}[c]{@{}c@{}}0.52\\ $\pm$0.06\end{tabular} & \multicolumn{1}{||c|}{\begin{tabular}[c]{@{}c@{}}0.29\\ $\pm$0.02\end{tabular}} & \begin{tabular}[c]{@{}c@{}}0.67\\ $\pm$0.06\end{tabular} \\ \hline
\multicolumn{1}{|c|}{\begin{tabular}[c]{@{}c@{}}iCaRL\end{tabular}} & \multicolumn{1}{c|}{\begin{tabular}[c]{@{}c@{}}0.62\\ $\pm$0.03\end{tabular}} & \begin{tabular}[c]{@{}c@{}}0.31\\ $\pm$0.02\end{tabular} & \multicolumn{1}{||c|}{\begin{tabular}[c]{@{}c@{}}0.45\\ $\pm$0.03\end{tabular}} & \begin{tabular}[c]{@{}c@{}}0.38\\ $\pm$0.02\end{tabular} & \multicolumn{1}{||c|}{\begin{tabular}[c]{@{}c@{}}0.44\\ $\pm$0.03\end{tabular}} & \begin{tabular}[c]{@{}c@{}}0.38\\ $\pm$0.06\end{tabular} & \multicolumn{1}{||c|}{\begin{tabular}[c]{@{}c@{}}0.34\\ $\pm$0.04\end{tabular}} & \begin{tabular}[c]{@{}c@{}}0.43\\ $\pm$0.06\end{tabular} & \multicolumn{1}{||c|}{\begin{tabular}[c]{@{}c@{}}0.31\\ $\pm$0.02\end{tabular}} & \begin{tabular}[c]{@{}c@{}}0.49\\ $\pm$0.08\end{tabular} & \multicolumn{1}{||c|}{\begin{tabular}[c]{@{}c@{}}0.27\\ $\pm$0.06\end{tabular}} & \begin{tabular}[c]{@{}c@{}}0.56\\ $\pm$0.15\end{tabular} \\ \hline
\multicolumn{1}{|c|}{\begin{tabular}[c]{@{}c@{}}Continual \\ CRUST\end{tabular}} & \multicolumn{1}{c|}{\begin{tabular}[c]{@{}c@{}}0.88\\ $\pm$0.0\end{tabular}} & \begin{tabular}[c]{@{}c@{}}0.04\\ $\pm$0.01\end{tabular} & \multicolumn{1}{||c|}{\begin{tabular}[c]{@{}c@{}}0.85\\ $\pm$0.01\end{tabular}} & \begin{tabular}[c]{@{}c@{}}0.04\\ $\pm$0.01\end{tabular} & \multicolumn{1}{||c|}{\begin{tabular}[c]{@{}c@{}}0.84\\ $\pm$0.01\end{tabular}} & \begin{tabular}[c]{@{}c@{}}0.04\\ $\pm$0.01\end{tabular} & \multicolumn{1}{||c|}{\begin{tabular}[c]{@{}c@{}}0.82\\ $\pm$0.02\end{tabular}} & \begin{tabular}[c]{@{}c@{}}0.05\\ $\pm$0.02\end{tabular} & \multicolumn{1}{||c|}{\begin{tabular}[c]{@{}c@{}}0.81\\ $\pm$0.01\end{tabular}} & \begin{tabular}[c]{@{}c@{}}0.04\\ $\pm$0.01\end{tabular} & \multicolumn{1}{||c|}{\begin{tabular}[c]{@{}c@{}}0.79\\ $\pm$0.02\end{tabular}} & \begin{tabular}[c]{@{}c@{}}0.04\\ $\pm$0.01\end{tabular} \\ \hline
\multicolumn{1}{|c|}{\begin{tabular}[c]{@{}c@{}}Continual \\ CosineCRUST\end{tabular}} & \multicolumn{1}{c|}{\begin{tabular}[c]{@{}c@{}}0.88\\ $\pm$0.01\end{tabular}} & \begin{tabular}[c]{@{}c@{}}0.06\\ $\pm$0.01\end{tabular} & \multicolumn{1}{||c|}{\begin{tabular}[c]{@{}c@{}}0.85\\ $\pm$0.0\end{tabular}} & \begin{tabular}[c]{@{}c@{}}0.06\\ $\pm$0.01\end{tabular} & \multicolumn{1}{||c|}{\begin{tabular}[c]{@{}c@{}}0.83\\ $\pm$0.01\end{tabular}} & \begin{tabular}[c]{@{}c@{}}0.06\\ $\pm$0.01\end{tabular} & \multicolumn{1}{||c|}{\begin{tabular}[c]{@{}c@{}}0.81\\ $\pm$0.01\end{tabular}} & \begin{tabular}[c]{@{}c@{}}0.07\\ $\pm$0.02\end{tabular} & \multicolumn{1}{||c|}{\begin{tabular}[c]{@{}c@{}}0.79\\ $\pm$0.02\end{tabular}} & \begin{tabular}[c]{@{}c@{}}0.08\\ $\pm$0.03\end{tabular} & \multicolumn{1}{||c|}{\begin{tabular}[c]{@{}c@{}}0.73\\ $\pm$0.02\end{tabular}} & \begin{tabular}[c]{@{}c@{}}0.12\\ $\pm$0.03\end{tabular} \\ \hline
\end{tabular}}
\caption{Final evaluation accuracy and forgetting on FashionMNIST for varying levels of label-flipping noise}
\label{tab:mem300FashionMNIST}
\end{table}
\end{minipage}

\begin{figure}[H]
  \centering
   \includegraphics[width=0.45\linewidth]{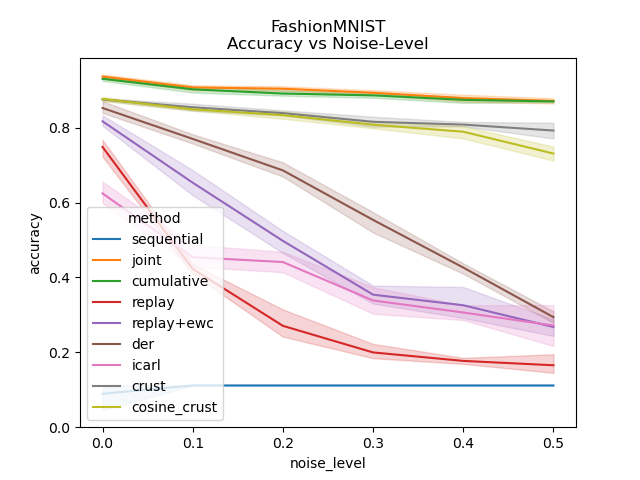}
   \includegraphics[width=0.45\linewidth]{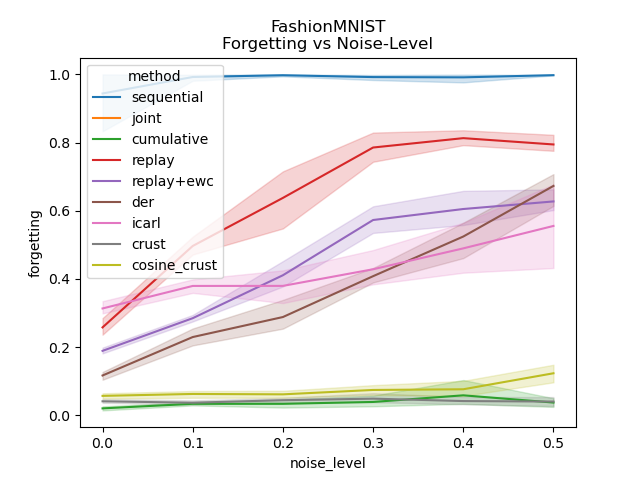}
   \caption{Looking at the final accuracy (left) and forgetting metric (right) at different label flipping noise levels (0.0-0.5) for FashionMNIST for different strategies}
   \label{fig:fashionmnist_noise}
\end{figure}

\noindent\hspace{0pt}\begin{minipage}{0.99\textwidth}
\RaggedRight
\indent\hspace{0.5cm} The model architecture we employ across all strategies for this dataset is ResNet18\cite{he2016deep}, with Adam\cite{kingma2014adam} as the optimizer and a batch size of 32. While the model structure remains consistent, the learning rate is adjusted for each strategy to optimize performance.

\indent\hspace{0.5cm} Our findings demonstrate that Continual CosineCRUST and Continual CRUST consistently outperform all other strategies, with their advantage becoming particularly evident at higher mislabel ratios. This resilience to label noise underscores the robustness of our proposed strategies in maintaining accuracy even as the proportion of mislabeled data increases, while other methods experience significant performance degradation.

\indent\hspace{0.5cm} Among the baseline methods, we observe a clear performance hierarchy. Dark ER exhibits superior performance compared to Random Replay with EWC, which in turn outperforms standard Random Replay. Interestingly, iCaRL consistently shows the lowest performance among these baseline strategies.
\end{minipage}

\newpage
\subsection{CIFAR10: Label Noise}

\noindent\hspace{0pt}\begin{minipage}{0.99\textwidth}
\RaggedRight
Training using continual learning on CIFAR10 presents additional challenges compared to simpler datasets with more consistent structure present in the data samples. As evident from Table \ref{tab:mem300CIFAR10}, forgetting is more pronounced across all settings, and training is more challenging. Despite these challenges, Continual CosineCRUST and Continual CRUST still demonstrate superior performance relative to other strategies.
\end{minipage}

\begin{table}[H]
\centering
\resizebox{\textwidth}{!}{\begin{tabular}{c|cc|cc|cc|cc|cc|cc|}
\cline{2-13}
\textbf{} & \multicolumn{2}{c|}{\textbf{Noise=0.0}} & \multicolumn{2}{||c|}{\textbf{Noise=0.1}} & \multicolumn{2}{||c|}{\textbf{Noise=0.2}} & \multicolumn{2}{||c|}{\textbf{Noise=0.3}} & \multicolumn{2}{||c|}{\textbf{Noise=0.4}} & \multicolumn{2}{||c|}{\textbf{Noise=0.5}} \\ \hline
\multicolumn{1}{|c|}{\textbf{Algorithm}} & \multicolumn{1}{c|}{\textit{\textbf{Acc}}} & \textit{\textbf{Forg.}} & \multicolumn{1}{||c|}{\textit{\textbf{Acc}}} & \textit{\textbf{Forg.}} & \multicolumn{1}{||c|}{\textit{\textbf{Acc}}} & \textit{\textbf{Forg.}} & \multicolumn{1}{||c|}{\textit{\textbf{Acc}}} & \textit{\textbf{Forg.}} & \multicolumn{1}{||c|}
{\textit{\textbf{Acc}}} & \textit{\textbf{Forg.}} & \multicolumn{1}{||c|}{\textit{\textbf{Acc}}} & \textit{\textbf{Forg.}} \\ \hline
\multicolumn{1}{|c|}{\begin{tabular}[c]{@{}c@{}}Naive \\ Sequential \\ Learner\end{tabular}} & \multicolumn{1}{c|}{\begin{tabular}[c]{@{}c@{}}0.11\\ $\pm$0.0\end{tabular}} & \begin{tabular}[c]{@{}c@{}}0.99\\ $\pm$0.0\end{tabular} & \multicolumn{1}{||c|}{\begin{tabular}[c]{@{}c@{}}0.11\\ $\pm$0.0\end{tabular}} & \begin{tabular}[c]{@{}c@{}}0.98\\ $\pm$0.01\end{tabular} & \multicolumn{1}{||c|}{\begin{tabular}[c]{@{}c@{}}0.11\\ $\pm$0.0\end{tabular}} & \begin{tabular}[c]{@{}c@{}}0.98\\ $\pm$0.01\end{tabular} & \multicolumn{1}{||c|}{\begin{tabular}[c]{@{}c@{}}0.11\\ $\pm$0.0\end{tabular}} & \begin{tabular}[c]{@{}c@{}}0.99\\ $\pm$0.01\end{tabular} & \multicolumn{1}{||c|}{\begin{tabular}[c]{@{}c@{}}0.11\\ $\pm$0.0\end{tabular}} & \begin{tabular}[c]{@{}c@{}}0.98\\ $\pm$0.01\end{tabular} & \multicolumn{1}{||c|}{\begin{tabular}[c]{@{}c@{}}0.11\\ $\pm$0.0\end{tabular}} & \begin{tabular}[c]{@{}c@{}}0.96\\ $\pm$0.02\end{tabular} \\ \hline
\multicolumn{1}{|c|}{\begin{tabular}[c]{@{}c@{}}Joint \\ Learner\end{tabular}} & \multicolumn{1}{c|}{\begin{tabular}[c]{@{}c@{}}0.9\\ $\pm$0.0\end{tabular}} & \begin{tabular}[c]{@{}c@{}} N/A \end{tabular} & \multicolumn{1}{||c|}{\begin{tabular}[c]{@{}c@{}}TBD\\ $\pm$TBD\end{tabular}} & \begin{tabular}[c]{@{}c@{}} N/A \end{tabular} & \multicolumn{1}{||c|}{\begin{tabular}[c]{@{}c@{}}0.86\\ $\pm$0.01\end{tabular}} & \begin{tabular}[c]{@{}c@{}} N/A \end{tabular} & \multicolumn{1}{||c|}{\begin{tabular}[c]{@{}c@{}}0.84\\ $\pm$0.01\end{tabular}} & \begin{tabular}[c]{@{}c@{}} N/A \end{tabular} & \multicolumn{1}{||c|}{\begin{tabular}[c]{@{}c@{}}0.81\\ $\pm$0.01\end{tabular}} & \begin{tabular}[c]{@{}c@{}} N/A \end{tabular} & \multicolumn{1}{||c|}{\begin{tabular}[c]{@{}c@{}}0.78\\ $\pm$0.01\end{tabular}} & \begin{tabular}[c]{@{}c@{}} N/A \end{tabular} \\ \hline
\multicolumn{1}{|c|}{\begin{tabular}[c]{@{}c@{}}Cumulative \\ Learner\end{tabular}} & \multicolumn{1}{c|}{\begin{tabular}[c]{@{}c@{}}0.86\\ $\pm$0.02\end{tabular}} & \begin{tabular}[c]{@{}c@{}}0.06\\ $\pm$0.03\end{tabular} & \multicolumn{1}{||c|}{\begin{tabular}[c]{@{}c@{}}0.86\\ $\pm$0.01\end{tabular}} & \begin{tabular}[c]{@{}c@{}}0.04\\ $\pm$0.01\end{tabular} & \multicolumn{1}{||c|}{\begin{tabular}[c]{@{}c@{}}0.82\\ $\pm$0.03\end{tabular}} & \begin{tabular}[c]{@{}c@{}}0.04\\ $\pm$0.02\end{tabular} & \multicolumn{1}{||c|}{\begin{tabular}[c]{@{}c@{}}0.81\\ $\pm$0.01\end{tabular}} & \begin{tabular}[c]{@{}c@{}}0.02\\ $\pm$0.01\end{tabular} & \multicolumn{1}{||c|}{\begin{tabular}[c]{@{}c@{}}0.76\\ $\pm$0.01\end{tabular}} & \begin{tabular}[c]{@{}c@{}}0.07\\ $\pm$0.01\end{tabular} & \multicolumn{1}{||c|}{\begin{tabular}[c]{@{}c@{}}0.71\\ $\pm$0.04\end{tabular}} & \begin{tabular}[c]{@{}c@{}}0.06\\ $\pm$0.03\end{tabular} \\ \hline
\multicolumn{1}{|c|}{\begin{tabular}[c]{@{}c@{}}Random \\ Replay\end{tabular}} & \multicolumn{1}{c|}{\begin{tabular}[c]{@{}c@{}}0.42\\ $\pm$0.01\end{tabular}} & \begin{tabular}[c]{@{}c@{}}0.57\\ $\pm$0.03\end{tabular} & \multicolumn{1}{||c|}{\begin{tabular}[c]{@{}c@{}}0.19\\ $\pm$0.01\end{tabular}} & \begin{tabular}[c]{@{}c@{}}0.79\\ $\pm$0.05\end{tabular} & \multicolumn{1}{||c|}{\begin{tabular}[c]{@{}c@{}}0.16\\ $\pm$0.01\end{tabular}} & \begin{tabular}[c]{@{}c@{}}0.8\\ $\pm$0.07\end{tabular} & \multicolumn{1}{||c|}{\begin{tabular}[c]{@{}c@{}}0.14\\ $\pm$0.0\end{tabular}} & \begin{tabular}[c]{@{}c@{}}0.9\\ $\pm$0.03\end{tabular} & \multicolumn{1}{||c|}{\begin{tabular}[c]{@{}c@{}}0.13\\ $\pm$0.01\end{tabular}} & \begin{tabular}[c]{@{}c@{}}0.86\\ $\pm$0.07\end{tabular} & \multicolumn{1}{||c|}{\begin{tabular}[c]{@{}c@{}}0.12\\ $\pm$0.01\end{tabular}} & \begin{tabular}[c]{@{}c@{}}0.87\\ $\pm$0.03\end{tabular} \\ \hline
\multicolumn{1}{|c|}{\begin{tabular}[c]{@{}c@{}}Random \\ Replay \\ w/EWC\end{tabular}} & \multicolumn{1}{c|}{\begin{tabular}[c]{@{}c@{}}0.52\\ $\pm$0.03\end{tabular}} & \begin{tabular}[c]{@{}c@{}}0.44\\ $\pm$0.02\end{tabular} & \multicolumn{1}{||c|}{\begin{tabular}[c]{@{}c@{}}0.3\\ $\pm$0.03\end{tabular}} & \begin{tabular}[c]{@{}c@{}}0.65\\ $\pm$0.04\end{tabular} & \multicolumn{1}{||c|}{\begin{tabular}[c]{@{}c@{}}0.21\\ $\pm$0.03\end{tabular}} & \begin{tabular}[c]{@{}c@{}}0.72\\ $\pm$0.05\end{tabular} & \multicolumn{1}{||c|}{\begin{tabular}[c]{@{}c@{}}0.19\\ $\pm$0.01\end{tabular}} & \begin{tabular}[c]{@{}c@{}}0.8\\ $\pm$0.02\end{tabular} & \multicolumn{1}{||c|}{\begin{tabular}[c]{@{}c@{}}0.14\\ $\pm$0.01\end{tabular}} & \begin{tabular}[c]{@{}c@{}}0.85\\ $\pm$0.02\end{tabular} & \multicolumn{1}{||c|}{\begin{tabular}[c]{@{}c@{}}0.13\\ $\pm$0.01\end{tabular}} & \begin{tabular}[c]{@{}c@{}}0.85\\ $\pm$0.05\end{tabular} \\ \hline
\multicolumn{1}{|c|}{\begin{tabular}[c]{@{}c@{}}Dark ER\end{tabular}} & \multicolumn{1}{c|}{\begin{tabular}[c]{@{}c@{}}0.68\\ $\pm$0.01\end{tabular}} & \begin{tabular}[c]{@{}c@{}}0.32\\ $\pm$0.03\end{tabular} & \multicolumn{1}{||c|}{\begin{tabular}[c]{@{}c@{}}0.48\\ $\pm$0.03\end{tabular}} & \begin{tabular}[c]{@{}c@{}}0.5\\ $\pm$0.04\end{tabular} & \multicolumn{1}{||c|}{\begin{tabular}[c]{@{}c@{}}0.41\\ $\pm$0.06\end{tabular}} & \begin{tabular}[c]{@{}c@{}}0.54\\ $\pm$0.04\end{tabular} & \multicolumn{1}{||c|}{\begin{tabular}[c]{@{}c@{}}0.29\\ $\pm$0.04\end{tabular}} & \begin{tabular}[c]{@{}c@{}}0.67\\ $\pm$0.03\end{tabular} & \multicolumn{1}{||c|}{\begin{tabular}[c]{@{}c@{}}0.23\\ $\pm$0.03\end{tabular}} & \begin{tabular}[c]{@{}c@{}}0.75\\ $\pm$0.04\end{tabular} & \multicolumn{1}{||c|}{\begin{tabular}[c]{@{}c@{}}0.17\\ $\pm$0.02\end{tabular}} & \begin{tabular}[c]{@{}c@{}}0.78\\ $\pm$0.04\end{tabular} \\ \hline
\multicolumn{1}{|c|}{\begin{tabular}[c]{@{}c@{}}iCaRL\end{tabular}} & \multicolumn{1}{c|}{\begin{tabular}[c]{@{}c@{}}0.61\\ $\pm$0.03\end{tabular}} & \begin{tabular}[c]{@{}c@{}}0.22\\ $\pm$0.06\end{tabular} & \multicolumn{1}{||c|}{\begin{tabular}[c]{@{}c@{}}0.52\\ $\pm$0.01\end{tabular}} & \begin{tabular}[c]{@{}c@{}}0.29\\ $\pm$0.03\end{tabular} & \multicolumn{1}{||c|}{\begin{tabular}[c]{@{}c@{}}0.44\\ $\pm$0.02\end{tabular}} & \begin{tabular}[c]{@{}c@{}}0.38\\ $\pm$0.05\end{tabular} & \multicolumn{1}{||c|}{\begin{tabular}[c]{@{}c@{}}0.34\\ $\pm$0.02\end{tabular}} & \begin{tabular}[c]{@{}c@{}}0.45\\ $\pm$0.05\end{tabular} & \multicolumn{1}{||c|}{\begin{tabular}[c]{@{}c@{}}0.27\\ $\pm$0.04\end{tabular}} & \begin{tabular}[c]{@{}c@{}}0.48\\ $\pm$0.05\end{tabular} & \multicolumn{1}{||c|}{\begin{tabular}[c]{@{}c@{}}0.25\\ $\pm$0.02\end{tabular}} & \begin{tabular}[c]{@{}c@{}}0.44\\ $\pm$0.03\end{tabular} \\ \hline
\multicolumn{1}{|c|}{\begin{tabular}[c]{@{}c@{}}Continual \\ CRUST\end{tabular}} & \multicolumn{1}{c|}{\begin{tabular}[c]{@{}c@{}}0.75\\ $\pm$0.02\end{tabular}} & \begin{tabular}[c]{@{}c@{}}0.17\\ $\pm$0.02\end{tabular} & \multicolumn{1}{||c|}{\begin{tabular}[c]{@{}c@{}}0.67\\ $\pm$0.02\end{tabular}} & \begin{tabular}[c]{@{}c@{}}0.19\\ $\pm$0.02\end{tabular} & \multicolumn{1}{||c|}{\begin{tabular}[c]{@{}c@{}}0.63\\ $\pm$0.01\end{tabular}} & \begin{tabular}[c]{@{}c@{}}0.2\\ $\pm$0.02\end{tabular} & \multicolumn{1}{||c|}{\begin{tabular}[c]{@{}c@{}}0.56\\ $\pm$0.02\end{tabular}} & \begin{tabular}[c]{@{}c@{}}0.21\\ $\pm$0.01\end{tabular} & \multicolumn{1}{||c|}{\begin{tabular}[c]{@{}c@{}}0.48\\ $\pm$0.02\end{tabular}} & \begin{tabular}[c]{@{}c@{}}0.23\\ $\pm$0.01\end{tabular} & \multicolumn{1}{||c|}{\begin{tabular}[c]{@{}c@{}}0.36\\ $\pm$0.02\end{tabular}} & \begin{tabular}[c]{@{}c@{}}0.21\\ $\pm$0.03\end{tabular} \\ \hline
\multicolumn{1}{|c|}{\begin{tabular}[c]{@{}c@{}}Continual \\ CosineCRUST\end{tabular}} & \multicolumn{1}{c|}{\begin{tabular}[c]{@{}c@{}}0.78\\ $\pm$0.01\end{tabular}} & \begin{tabular}[c]{@{}c@{}}0.18\\ $\pm$0.02\end{tabular} & \multicolumn{1}{||c|}{\begin{tabular}[c]{@{}c@{}}0.69\\ $\pm$0.01\end{tabular}} & \begin{tabular}[c]{@{}c@{}}0.16\\ $\pm$0.01\end{tabular} & \multicolumn{1}{||c|}{\begin{tabular}[c]{@{}c@{}}0.64\\ $\pm$0.02\end{tabular}} & \begin{tabular}[c]{@{}c@{}}0.17\\ $\pm$0.01\end{tabular} & \multicolumn{1}{||c|}{\begin{tabular}[c]{@{}c@{}}0.57\\ $\pm$0.01\end{tabular}} & \begin{tabular}[c]{@{}c@{}}0.19\\ $\pm$0.03\end{tabular} & \multicolumn{1}{||c|}{\begin{tabular}[c]{@{}c@{}}0.49\\ $\pm$0.01\end{tabular}} & \begin{tabular}[c]{@{}c@{}}0.21\\ $\pm$0.01\end{tabular} & \multicolumn{1}{||c|}{\begin{tabular}[c]{@{}c@{}}0.4\\ $\pm$0.01\end{tabular}} & \begin{tabular}[c]{@{}c@{}}0.22\\ $\pm$0.04\end{tabular} \\ \hline
\end{tabular}}
\caption{Final evaluation accuracy and forgetting on CIFAR10 for varying levels of label-flipping noise}
\label{tab:mem300CIFAR10}
\end{table}

\begin{figure}[H]
  \centering
   \includegraphics[width=0.45\linewidth]{graphics/CIFAR10_accuracy.png}
   \includegraphics[width=0.45\linewidth]{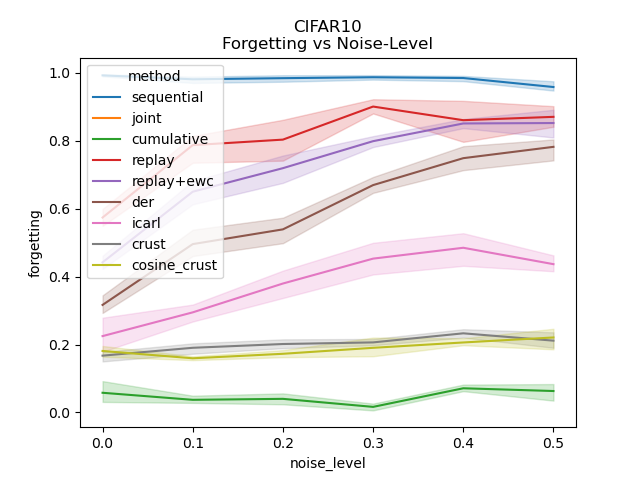}
   \caption{Looking at the final accuracy (left) and forgetting metric (right) at different label flipping noise levels (0.0-0.5) for CIFAR10 for different strategies}
   \label{fig:cifar_noise}
\end{figure}

\noindent\hspace{0pt}\begin{minipage}{0.99\textwidth}
\RaggedRight
\indent\hspace{0.5cm} For all strategies, we employ a ResNet18\cite{he2016deep} architecture, utilizing Adam\cite{kingma2014adam} as the optimizer with a batch size of 32 and Coreset memory is set to 300 samples per class. This setup aligns with common practices in the field of continual learning for image classification tasks. Our experimental protocol mirrors that used for the other datasets: we first randomly shuffle the classes, then initialize the first experience with two classes, subsequently adding one class at a time for each new experience. \textbf{Note that this is a different setup than the standard splitCIFAR10 benchmark, which spans five experiences (vs ten for our experiments) and includes two new classes per experience (vs one for our experiments)}.

\indent\hspace{0.5cm} This approach allows us to evaluate the robustness of our methods in a more complex domain, where the increased diversity and difficulty of CIFAR10 images amplify the challenges of catastrophic forgetting and knowledge transfer between tasks. The consistent superiority of our proposed methods underscores their efficacy in managing these challenges even in more demanding scenarios.
\end{minipage}

\newpage
\subsection{MSTAR: Label Noise}

\noindent\hspace{0pt}\begin{minipage}{0.99\textwidth}
\RaggedRight
For our experiments on the MSTAR dataset, we employ the EfficientNetV2s\cite{tan2019efficientnet} architecture, which has demonstrated superior performance in various image classification tasks. We utilize the Adam\cite{kingma2014adam} optimizer with a batch size of 18 across all strategies. The results for our proposed strategies and other comparative methods are presented in Table \ref{tab:mem60MSTAR}, with the Coreset memory fixed at 60 samples per class.

Our findings reveal that Continual CosineCRUST exhibits superior performance compared to Continual CRUST when the mislabel ratio is low. As the mislabel ratio increases, both strategies demonstrate comparable performance. Notably, both Continual CosineCRUST and Continual CRUST consistently outperform Random Replay, Random Replay with EWC, iCaRL, and Dark ER across all mislabel ratios. Among the baseline methods, Dark ER shows the best performance, surpassing Random Replay, Random Replay with EWC, and iCaRL.
\end{minipage}

\begin{table}[H]
\centering
\resizebox{\textwidth}{!}{\begin{tabular}{c|cc|cc|cc|cc|cc|cc|}
\cline{2-13}
\textbf{} & \multicolumn{2}{c|}{\textbf{Noise=0.0}} & \multicolumn{2}{||c|}{\textbf{Noise=0.1}} & \multicolumn{2}{||c|}{\textbf{Noise=0.2}} & \multicolumn{2}{||c|}{\textbf{Noise=0.3}} & \multicolumn{2}{||c|}{\textbf{Noise=0.4}} & \multicolumn{2}{||c|}{\textbf{Noise=0.5}} \\ \hline
\multicolumn{1}{|c|}{\textbf{Algorithm}} & \multicolumn{1}{c|}{\textit{\textbf{Acc}}} & \textit{\textbf{Forg.}} & \multicolumn{1}{||c|}{\textit{\textbf{Acc}}} & \textit{\textbf{Forg.}} & \multicolumn{1}{||c|}{\textit{\textbf{Acc}}} & \textit{\textbf{Forg.}} & \multicolumn{1}{||c|}{\textit{\textbf{Acc}}} & \textit{\textbf{Forg.}} & \multicolumn{1}{||c|}
{\textit{\textbf{Acc}}} & \textit{\textbf{Forg.}} & \multicolumn{1}{||c|}{\textit{\textbf{Acc}}} & \textit{\textbf{Forg.}} \\ \hline
\multicolumn{1}{|c|}{\begin{tabular}[c]{@{}c@{}}Naive \\ Sequential \\ Learner\end{tabular}} & \multicolumn{1}{c|}{\begin{tabular}[c]{@{}c@{}}0.11\\ $\pm$0.0\end{tabular}} & \begin{tabular}[c]{@{}c@{}}1.0\\ $\pm$0.0\end{tabular} & \multicolumn{1}{||c|}{\begin{tabular}[c]{@{}c@{}}0.11\\ $\pm$0.0\end{tabular}} & \begin{tabular}[c]{@{}c@{}}1.0\\ $\pm$0.0\end{tabular} & \multicolumn{1}{||c|}{\begin{tabular}[c]{@{}c@{}}0.11\\ $\pm$0.0\end{tabular}} & \begin{tabular}[c]{@{}c@{}}0.97\\ $\pm$0.04\end{tabular} & \multicolumn{1}{||c|}{\begin{tabular}[c]{@{}c@{}}0.11\\ $\pm$0.0\end{tabular}} & \begin{tabular}[c]{@{}c@{}}0.95\\ $\pm$0.05\end{tabular} & \multicolumn{1}{||c|}{\begin{tabular}[c]{@{}c@{}}0.11\\ $\pm$0.0\end{tabular}} & \begin{tabular}[c]{@{}c@{}}0.98\\ $\pm$0.01\end{tabular} & \multicolumn{1}{||c|}{\begin{tabular}[c]{@{}c@{}}0.11\\ $\pm$0.0\end{tabular}} & \begin{tabular}[c]{@{}c@{}}0.93\\ $\pm$0.06\end{tabular} \\ \hline
\multicolumn{1}{|c|}{\begin{tabular}[c]{@{}c@{}}Joint \\ Learner\end{tabular}} & \multicolumn{1}{c|}{\begin{tabular}[c]{@{}c@{}}0.97\\ $\pm$0.01\end{tabular}} & \begin{tabular}[c]{@{}c@{}} N/A \end{tabular} & \multicolumn{1}{||c|}{\begin{tabular}[c]{@{}c@{}}0.9\\ $\pm$0.02\end{tabular}} & \begin{tabular}[c]{@{}c@{}} N/A \end{tabular} & \multicolumn{1}{||c|}{\begin{tabular}[c]{@{}c@{}}0.89\\ $\pm$0.02\end{tabular}} & \begin{tabular}[c]{@{}c@{}} N/A \end{tabular} & \multicolumn{1}{||c|}{\begin{tabular}[c]{@{}c@{}}0.78\\ $\pm$0.02\end{tabular}} & \begin{tabular}[c]{@{}c@{}} N/A \end{tabular} & \multicolumn{1}{||c|}{\begin{tabular}[c]{@{}c@{}}0.72\\ $\pm$0.06\end{tabular}} & \begin{tabular}[c]{@{}c@{}} N/A \end{tabular} & \multicolumn{1}{||c|}{\begin{tabular}[c]{@{}c@{}}0.58\\ $\pm$0.05\end{tabular}} & \begin{tabular}[c]{@{}c@{}} N/A \end{tabular} \\ \hline
\multicolumn{1}{|c|}{\begin{tabular}[c]{@{}c@{}}Cumulative \\ Learner\end{tabular}} & \multicolumn{1}{c|}{\begin{tabular}[c]{@{}c@{}}0.99\\ $\pm$0.0\end{tabular}} & \begin{tabular}[c]{@{}c@{}}-0.09\\ $\pm$0.07\end{tabular} & \multicolumn{1}{||c|}{\begin{tabular}[c]{@{}c@{}}0.96\\ $\pm$0.01\end{tabular}} & \begin{tabular}[c]{@{}c@{}}-0.07\\ $\pm$0.09\end{tabular} & \multicolumn{1}{||c|}{\begin{tabular}[c]{@{}c@{}}0.95\\ $\pm$0.01\end{tabular}} & \begin{tabular}[c]{@{}c@{}}-0.02\\ $\pm$0.07\end{tabular} & \multicolumn{1}{||c|}{\begin{tabular}[c]{@{}c@{}}0.92\\ $\pm$0.02\end{tabular}} & \begin{tabular}[c]{@{}c@{}}-0.01\\ $\pm$0.08\end{tabular} & \multicolumn{1}{||c|}{\begin{tabular}[c]{@{}c@{}}0.9\\ $\pm$0.03\end{tabular}} & \begin{tabular}[c]{@{}c@{}}0.01\\ $\pm$0.04\end{tabular} & \multicolumn{1}{||c|}{\begin{tabular}[c]{@{}c@{}}0.8\\ $\pm$0.05\end{tabular}} & \begin{tabular}[c]{@{}c@{}}-0.03\\ $\pm$0.06\end{tabular} \\ \hline
\multicolumn{1}{|c|}{\begin{tabular}[c]{@{}c@{}}Random \\ Replay\end{tabular}} & \multicolumn{1}{c|}{\begin{tabular}[c]{@{}c@{}}0.91\\ $\pm$0.04\end{tabular}} & \begin{tabular}[c]{@{}c@{}}0.04\\ $\pm$0.08\end{tabular} & \multicolumn{1}{||c|}{\begin{tabular}[c]{@{}c@{}}0.81\\ $\pm$0.02\end{tabular}} & \begin{tabular}[c]{@{}c@{}}0.16\\ $\pm$0.03\end{tabular} & \multicolumn{1}{||c|}{\begin{tabular}[c]{@{}c@{}}0.67\\ $\pm$0.03\end{tabular}} & \begin{tabular}[c]{@{}c@{}}0.28\\ $\pm$0.16\end{tabular} & \multicolumn{1}{||c|}{\begin{tabular}[c]{@{}c@{}}0.57\\ $\pm$0.03\end{tabular}} & \begin{tabular}[c]{@{}c@{}}0.31\\ $\pm$0.11\end{tabular} & \multicolumn{1}{||c|}{\begin{tabular}[c]{@{}c@{}}0.44\\ $\pm$0.01\end{tabular}} & \begin{tabular}[c]{@{}c@{}}0.52\\ $\pm$0.05\end{tabular} & \multicolumn{1}{||c|}{\begin{tabular}[c]{@{}c@{}}0.36\\ $\pm$0.04\end{tabular}} & \begin{tabular}[c]{@{}c@{}}0.7\\ $\pm$0.07\end{tabular} \\ \hline
\multicolumn{1}{|c|}{\begin{tabular}[c]{@{}c@{}}Random \\ Replay \\ w/EWC\end{tabular}} & \multicolumn{1}{c|}{\begin{tabular}[c]{@{}c@{}}0.86\\ $\pm$0.03\end{tabular}} & \begin{tabular}[c]{@{}c@{}}0.1\\ $\pm$0.08\end{tabular} & \multicolumn{1}{||c|}{\begin{tabular}[c]{@{}c@{}}0.65\\ $\pm$0.03\end{tabular}} & \begin{tabular}[c]{@{}c@{}}0.28\\ $\pm$0.09\end{tabular} & \multicolumn{1}{||c|}{\begin{tabular}[c]{@{}c@{}}0.51\\ $\pm$0.06\end{tabular}} & \begin{tabular}[c]{@{}c@{}}0.36\\ $\pm$0.06\end{tabular} & \multicolumn{1}{||c|}{\begin{tabular}[c]{@{}c@{}}0.37\\ $\pm$0.04\end{tabular}} & \begin{tabular}[c]{@{}c@{}}0.43\\ $\pm$0.05\end{tabular} & \multicolumn{1}{||c|}{\begin{tabular}[c]{@{}c@{}}0.29\\ $\pm$0.02\end{tabular}} & \begin{tabular}[c]{@{}c@{}}0.61\\ $\pm$0.06\end{tabular} & \multicolumn{1}{||c|}{\begin{tabular}[c]{@{}c@{}}0.18\\ $\pm$0.09\end{tabular}} & \begin{tabular}[c]{@{}c@{}}0.64\\ $\pm$0.02\end{tabular} \\ \hline
\multicolumn{1}{|c|}{\begin{tabular}[c]{@{}c@{}}Dark ER\end{tabular}} & \multicolumn{1}{c|}{\begin{tabular}[c]{@{}c@{}}0.93\\ $\pm$0.03\end{tabular}} & \begin{tabular}[c]{@{}c@{}}-0.01\\ $\pm$0.07\end{tabular} & \multicolumn{1}{||c|}{\begin{tabular}[c]{@{}c@{}}0.87\\ $\pm$0.02\end{tabular}} & \begin{tabular}[c]{@{}c@{}}0.07\\ $\pm$0.07\end{tabular} & \multicolumn{1}{||c|}{\begin{tabular}[c]{@{}c@{}}0.75\\ $\pm$0.02\end{tabular}} & \begin{tabular}[c]{@{}c@{}}0.25\\ $\pm$0.03\end{tabular} & \multicolumn{1}{||c|}{\begin{tabular}[c]{@{}c@{}}0.63\\ $\pm$0.02\end{tabular}} & \begin{tabular}[c]{@{}c@{}}0.36\\ $\pm$0.06\end{tabular} & \multicolumn{1}{||c|}{\begin{tabular}[c]{@{}c@{}}0.49\\ $\pm$0.03\end{tabular}} & \begin{tabular}[c]{@{}c@{}}0.52\\ $\pm$0.06\end{tabular} & \multicolumn{1}{||c|}{\begin{tabular}[c]{@{}c@{}}0.38\\ $\pm$0.01\end{tabular}} & \begin{tabular}[c]{@{}c@{}}0.64\\ $\pm$0.08\end{tabular} \\ \hline
\multicolumn{1}{|c|}{\begin{tabular}[c]{@{}c@{}}iCaRL\end{tabular}} & \multicolumn{1}{c|}{\begin{tabular}[c]{@{}c@{}}0.86\\ $\pm$0.01\end{tabular}} & \begin{tabular}[c]{@{}c@{}}0.12\\ $\pm$0.02\end{tabular} & \multicolumn{1}{||c|}{\begin{tabular}[c]{@{}c@{}}0.72\\ $\pm$0.04\end{tabular}} & \begin{tabular}[c]{@{}c@{}}0.19\\ $\pm$0.03\end{tabular} & \multicolumn{1}{||c|}{\begin{tabular}[c]{@{}c@{}}0.61\\ $\pm$0.01\end{tabular}} & \begin{tabular}[c]{@{}c@{}}0.31\\ $\pm$0.03\end{tabular} & \multicolumn{1}{||c|}{\begin{tabular}[c]{@{}c@{}}0.45\\ $\pm$0.02\end{tabular}} & \begin{tabular}[c]{@{}c@{}}0.43\\ $\pm$0.08\end{tabular} & \multicolumn{1}{||c|}{\begin{tabular}[c]{@{}c@{}}0.34\\ $\pm$0.03\end{tabular}} & \begin{tabular}[c]{@{}c@{}}0.49\\ $\pm$0.04\end{tabular} & \multicolumn{1}{||c|}{\begin{tabular}[c]{@{}c@{}}0.25\\ $\pm$0.02\end{tabular}} & \begin{tabular}[c]{@{}c@{}}0.59\\ $\pm$0.08\end{tabular} \\ \hline
\multicolumn{1}{|c|}{\begin{tabular}[c]{@{}c@{}}Continual \\ CRUST\end{tabular}} & \multicolumn{1}{c|}{\begin{tabular}[c]{@{}c@{}}0.93\\ $\pm$0.01\end{tabular}} & \begin{tabular}[c]{@{}c@{}}0.02\\ $\pm$0.01\end{tabular} & \multicolumn{1}{||c|}{\begin{tabular}[c]{@{}c@{}}0.89\\ $\pm$0.02\end{tabular}} & \begin{tabular}[c]{@{}c@{}}-0.01\\ $\pm$0.06\end{tabular} & \multicolumn{1}{||c|}{\begin{tabular}[c]{@{}c@{}}0.89\\ $\pm$0.02\end{tabular}} & \begin{tabular}[c]{@{}c@{}}0.01\\ $\pm$0.05\end{tabular} & \multicolumn{1}{||c|}{\begin{tabular}[c]{@{}c@{}}0.84\\ $\pm$0.02\end{tabular}} & \begin{tabular}[c]{@{}c@{}}0.01\\ $\pm$0.01\end{tabular} & \multicolumn{1}{||c|}{\begin{tabular}[c]{@{}c@{}}0.81\\ $\pm$0.01\end{tabular}} & \begin{tabular}[c]{@{}c@{}}0.07\\ $\pm$0.03\end{tabular} & \multicolumn{1}{||c|}{\begin{tabular}[c]{@{}c@{}}0.71\\ $\pm$0.02\end{tabular}} & \begin{tabular}[c]{@{}c@{}}0.05\\ $\pm$0.05\end{tabular} \\ \hline
\multicolumn{1}{|c|}{\begin{tabular}[c]{@{}c@{}}Continual \\ CosineCRUST\end{tabular}} & \multicolumn{1}{c|}{\begin{tabular}[c]{@{}c@{}}0.97\\ $\pm$0.01\end{tabular}} & \begin{tabular}[c]{@{}c@{}}0.0\\ $\pm$0.01\end{tabular} & \multicolumn{1}{||c|}{\begin{tabular}[c]{@{}c@{}}0.95\\ $\pm$0.01\end{tabular}} & \begin{tabular}[c]{@{}c@{}}0.01\\ $\pm$0.02\end{tabular} & \multicolumn{1}{||c|}{\begin{tabular}[c]{@{}c@{}}0.91\\ $\pm$0.01\end{tabular}} & \begin{tabular}[c]{@{}c@{}}-0.01\\ $\pm$0.03\end{tabular} & \multicolumn{1}{||c|}{\begin{tabular}[c]{@{}c@{}}0.85\\ $\pm$0.01\end{tabular}} & \begin{tabular}[c]{@{}c@{}}0.0\\ $\pm$0.03\end{tabular} & \multicolumn{1}{||c|}{\begin{tabular}[c]{@{}c@{}}0.81\\ $\pm$0.01\end{tabular}} & \begin{tabular}[c]{@{}c@{}}-0.03\\ $\pm$0.06\end{tabular} & \multicolumn{1}{||c|}{\begin{tabular}[c]{@{}c@{}}0.73\\ $\pm$0.02\end{tabular}} & \begin{tabular}[c]{@{}c@{}}0.0\\ $\pm$0.05\end{tabular} \\ \hline
\end{tabular}}
\caption{Final evaluation accuracy and forgetting on MSTAR for varying levels of label-flipping noise}
\label{tab:mem60MSTAR}
\end{table}

\begin{figure}[H]
  \centering
   \includegraphics[width=0.45\linewidth]{graphics/MSTAR_accuracy.png}
   \includegraphics[width=0.45\linewidth]{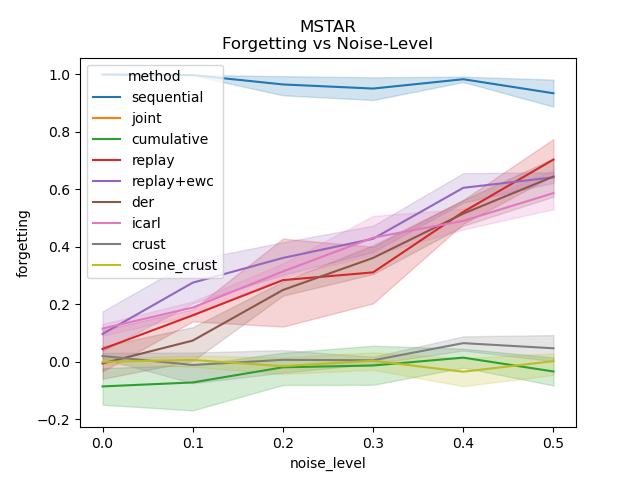}
   \caption{Looking at the final accuracy (left) and forgetting metric (right) at different label flipping noise levels (0.0-0.5) for MSTAR for different strategies}
   \label{fig:mstar_noise}
\end{figure}

\newpage
\subsection{PathMNIST+: Label Noise}
\noindent\hspace{0pt}\begin{minipage}{0.99\textwidth}
\RaggedRight
Our final data is PathMNIST+, a part of MedMNISTv2, a collection of large-scale benchmark datasets for medical and biological image analysis \cite{yang2023medmnist}. We use a subset of the dataset consisting of 18,000 training and 2,700 test non-overlapping RGB images, split evenly across the nine classes in the dataset. Every image is 224x224 pixels.

\indent\hspace{0.5cm} We follow similar procedure as with the other datasets. As with MSTAR, We use EfficientNetV2s as our model architecture. Note that for this dataset, we found it unnecessary to train for many epochs. To avoid overfitting, we used just ten epochs for phase 1 training and twenty epochs for phase 2 training. 

\indent\hspace{0.5cm} Results for this dataset shows some interesting behavior not seen in other settings. We notice the joint learner learns well under high noise (0.81 Acc for noise=0.5), however, the cumulative learner performs much worse (0.49 Acc) for the same noise level. This suggests that this dataset may exhibit a stronger compounding effect between noisy labels and forgetting compared to other datasets where the cumulative learner generally matches or outperforms the joint learner regardless of noise level. We see another interesting trend that Random Replay actually outperforms some of the more powerful strategies under most noise levels, which also may suggest this dataset has some unique property that makes continual learning challenging. As with other datasets, Continual (Cosine)CRUST notably outperforms the other strategies from prior work.

\end{minipage}

\begin{table}[H]
\centering
\resizebox{\textwidth}{!}{\begin{tabular}{c|cc|cc|cc|cc|cc|cc|}
\cline{2-13}
\textbf{} & \multicolumn{2}{c|}{\textbf{Noise=0.0}} & \multicolumn{2}{||c|}{\textbf{Noise=0.1}} & \multicolumn{2}{||c|}{\textbf{Noise=0.2}} & \multicolumn{2}{||c|}{\textbf{Noise=0.3}} & \multicolumn{2}{||c|}{\textbf{Noise=0.4}} & \multicolumn{2}{||c|}{\textbf{Noise=0.5}} \\ \hline
\multicolumn{1}{|c|}{\textbf{Algorithm}} & \multicolumn{1}{c|}{\textit{\textbf{Acc}}} & \textit{\textbf{Forg.}} & \multicolumn{1}{||c|}{\textit{\textbf{Acc}}} & \textit{\textbf{Forg.}} & \multicolumn{1}{||c|}{\textit{\textbf{Acc}}} & \textit{\textbf{Forg.}} & \multicolumn{1}{||c|}{\textit{\textbf{Acc}}} & \textit{\textbf{Forg.}} & \multicolumn{1}{||c|}
{\textit{\textbf{Acc}}} & \textit{\textbf{Forg.}} & \multicolumn{1}{||c|}{\textit{\textbf{Acc}}} & \textit{\textbf{Forg.}} \\ \hline
\multicolumn{1}{|c|}{\begin{tabular}[c]{@{}c@{}}Naive \\ Sequential \\ Learner\end{tabular}} & \multicolumn{1}{c|}{\begin{tabular}[c]{@{}c@{}}0.12\\ $\pm$0.0\end{tabular}} & \begin{tabular}[c]{@{}c@{}}1.0\\ $\pm$0.0\end{tabular} & \multicolumn{1}{||c|}{\begin{tabular}[c]{@{}c@{}}0.12\\ $\pm$0.0\end{tabular}} & \begin{tabular}[c]{@{}c@{}}1.0\\ $\pm$0.0\end{tabular} & \multicolumn{1}{||c|}{\begin{tabular}[c]{@{}c@{}}0.12\\ $\pm$0.0\end{tabular}} & \begin{tabular}[c]{@{}c@{}}0.98\\ $\pm$0.02\end{tabular} & \multicolumn{1}{||c|}{\begin{tabular}[c]{@{}c@{}}0.12\\ $\pm$0.0\end{tabular}} & \begin{tabular}[c]{@{}c@{}}0.99\\ $\pm$0.01\end{tabular} & \multicolumn{1}{||c|}{\begin{tabular}[c]{@{}c@{}}0.12\\ $\pm$0.0\end{tabular}} & \begin{tabular}[c]{@{}c@{}}0.98\\ $\pm$0.01\end{tabular} & \multicolumn{1}{||c|}{\begin{tabular}[c]{@{}c@{}}0.12\\ $\pm$0.0\end{tabular}} & \begin{tabular}[c]{@{}c@{}}0.99\\ $\pm$0.0\end{tabular} \\ \hline
\multicolumn{1}{|c|}{\begin{tabular}[c]{@{}c@{}}Joint \\ Learner\end{tabular}} & \multicolumn{1}{c|}{\begin{tabular}[c]{@{}c@{}}0.92\\ $\pm$0.01\end{tabular}} & \begin{tabular}[c]{@{}c@{}} N/A \end{tabular} & \multicolumn{1}{||c|}{\begin{tabular}[c]{@{}c@{}}0.82\\ $\pm$0.02\end{tabular}} & \begin{tabular}[c]{@{}c@{}} N/A \end{tabular} & \multicolumn{1}{||c|}{\begin{tabular}[c]{@{}c@{}}0.86\\ $\pm$0.03\end{tabular}} & \begin{tabular}[c]{@{}c@{}} N/A \end{tabular} & \multicolumn{1}{||c|}{\begin{tabular}[c]{@{}c@{}}0.84\\ $\pm$0.01\end{tabular}} & \begin{tabular}[c]{@{}c@{}} N/A \end{tabular} & \multicolumn{1}{||c|}{\begin{tabular}[c]{@{}c@{}}0.81\\ $\pm$0.03\end{tabular}} & \begin{tabular}[c]{@{}c@{}} N/A \end{tabular} & \multicolumn{1}{||c|}{\begin{tabular}[c]{@{}c@{}}0.81\\ $\pm$0.02\end{tabular}} & \begin{tabular}[c]{@{}c@{}} N/A \end{tabular} \\ \hline
\multicolumn{1}{|c|}{\begin{tabular}[c]{@{}c@{}}Cumulative \\ Learner\end{tabular}} & \multicolumn{1}{c|}{\begin{tabular}[c]{@{}c@{}}0.91\\ $\pm$0.01\end{tabular}} & \begin{tabular}[c]{@{}c@{}}0.04\\ $\pm$0.03\end{tabular} & \multicolumn{1}{||c|}{\begin{tabular}[c]{@{}c@{}}0.82\\ $\pm$0.01\end{tabular}} & \begin{tabular}[c]{@{}c@{}}0.07\\ $\pm$0.01\end{tabular} & \multicolumn{1}{||c|}{\begin{tabular}[c]{@{}c@{}}0.75\\ $\pm$0.02\end{tabular}} & \begin{tabular}[c]{@{}c@{}}0.1\\ $\pm$0.01\end{tabular} & \multicolumn{1}{||c|}{\begin{tabular}[c]{@{}c@{}}0.66\\ $\pm$0.03\end{tabular}} & \begin{tabular}[c]{@{}c@{}}0.14\\ $\pm$0.03\end{tabular} & \multicolumn{1}{||c|}{\begin{tabular}[c]{@{}c@{}}0.57\\ $\pm$0.03\end{tabular}} & \begin{tabular}[c]{@{}c@{}}0.18\\ $\pm$0.02\end{tabular} & \multicolumn{1}{||c|}{\begin{tabular}[c]{@{}c@{}}0.49\\ $\pm$0.03\end{tabular}} & \begin{tabular}[c]{@{}c@{}}0.22\\ $\pm$0.02\end{tabular} \\ \hline
\multicolumn{1}{|c|}{\begin{tabular}[c]{@{}c@{}}Random \\ Replay\end{tabular}} & \multicolumn{1}{c|}{\begin{tabular}[c]{@{}c@{}}0.89\\ $\pm$0.02\end{tabular}} & \begin{tabular}[c]{@{}c@{}}0.05\\ $\pm$0.03\end{tabular} & \multicolumn{1}{||c|}{\begin{tabular}[c]{@{}c@{}}0.79\\ $\pm$0.01\end{tabular}} & \begin{tabular}[c]{@{}c@{}}0.1\\ $\pm$0.03\end{tabular} & \multicolumn{1}{||c|}{\begin{tabular}[c]{@{}c@{}}0.7\\ $\pm$0.04\end{tabular}} & \begin{tabular}[c]{@{}c@{}}0.13\\ $\pm$0.03\end{tabular} & \multicolumn{1}{||c|}{\begin{tabular}[c]{@{}c@{}}0.58\\ $\pm$0.03\end{tabular}} & \begin{tabular}[c]{@{}c@{}}0.21\\ $\pm$0.09\end{tabular} & \multicolumn{1}{||c|}{\begin{tabular}[c]{@{}c@{}}0.53\\ $\pm$0.03\end{tabular}} & \begin{tabular}[c]{@{}c@{}}0.23\\ $\pm$0.04\end{tabular} & \multicolumn{1}{||c|}{\begin{tabular}[c]{@{}c@{}}0.45\\ $\pm$0.04\end{tabular}} & \begin{tabular}[c]{@{}c@{}}0.29\\ $\pm$0.08\end{tabular} \\ \hline
\multicolumn{1}{|c|}{\begin{tabular}[c]{@{}c@{}}Random \\ Replay \\ w/EWC\end{tabular}} & \multicolumn{1}{c|}{\begin{tabular}[c]{@{}c@{}}0.82\\ $\pm$0.05\end{tabular}} & \begin{tabular}[c]{@{}c@{}}0.1\\ $\pm$0.03\end{tabular} & \multicolumn{1}{||c|}{\begin{tabular}[c]{@{}c@{}}0.64\\ $\pm$0.03\end{tabular}} & \begin{tabular}[c]{@{}c@{}}0.28\\ $\pm$0.06\end{tabular} & \multicolumn{1}{||c|}{\begin{tabular}[c]{@{}c@{}}0.44\\ $\pm$0.05\end{tabular}} & \begin{tabular}[c]{@{}c@{}}0.45\\ $\pm$0.04\end{tabular} & \multicolumn{1}{||c|}{\begin{tabular}[c]{@{}c@{}}0.36\\ $\pm$0.03\end{tabular}} & \begin{tabular}[c]{@{}c@{}}0.52\\ $\pm$0.06\end{tabular} & \multicolumn{1}{||c|}{\begin{tabular}[c]{@{}c@{}}0.3\\ $\pm$0.07\end{tabular}} & \begin{tabular}[c]{@{}c@{}}0.57\\ $\pm$0.04\end{tabular} & \multicolumn{1}{||c|}{\begin{tabular}[c]{@{}c@{}}0.24\\ $\pm$0.07\end{tabular}} & \begin{tabular}[c]{@{}c@{}}0.68\\ $\pm$0.04\end{tabular} \\ \hline
\multicolumn{1}{|c|}{\begin{tabular}[c]{@{}c@{}}Dark ER\end{tabular}} & \multicolumn{1}{c|}{\begin{tabular}[c]{@{}c@{}}0.72\\ $\pm$0.29\end{tabular}} & \begin{tabular}[c]{@{}c@{}}0.17\\ $\pm$0.24\end{tabular} & \multicolumn{1}{||c|}{\begin{tabular}[c]{@{}c@{}}0.75\\ $\pm$0.03\end{tabular}} & \begin{tabular}[c]{@{}c@{}}0.16\\ $\pm$0.09\end{tabular} & \multicolumn{1}{||c|}{\begin{tabular}[c]{@{}c@{}}0.59\\ $\pm$0.04\end{tabular}} & \begin{tabular}[c]{@{}c@{}}0.35\\ $\pm$0.07\end{tabular} & \multicolumn{1}{||c|}{\begin{tabular}[c]{@{}c@{}}0.5\\ $\pm$0.04\end{tabular}} & \begin{tabular}[c]{@{}c@{}}0.43\\ $\pm$0.05\end{tabular} & \multicolumn{1}{||c|}{\begin{tabular}[c]{@{}c@{}}0.37\\ $\pm$0.05\end{tabular}} & \begin{tabular}[c]{@{}c@{}}0.57\\ $\pm$0.05\end{tabular} & \multicolumn{1}{||c|}{\begin{tabular}[c]{@{}c@{}}0.32\\ $\pm$0.07\end{tabular}} & \begin{tabular}[c]{@{}c@{}}0.61\\ $\pm$0.07\end{tabular} \\ \hline
\multicolumn{1}{|c|}{\begin{tabular}[c]{@{}c@{}}iCaRL\end{tabular}} & \multicolumn{1}{c|}{\begin{tabular}[c]{@{}c@{}}0.75\\ $\pm$0.05\end{tabular}} & \begin{tabular}[c]{@{}c@{}}0.04\\ $\pm$0.07\end{tabular} & \multicolumn{1}{||c|}{\begin{tabular}[c]{@{}c@{}}0.71\\ $\pm$0.04\end{tabular}} & \begin{tabular}[c]{@{}c@{}}0.11\\ $\pm$0.04\end{tabular} & \multicolumn{1}{||c|}{\begin{tabular}[c]{@{}c@{}}0.49\\ $\pm$0.07\end{tabular}} & \begin{tabular}[c]{@{}c@{}}0.42\\ $\pm$0.07\end{tabular} & \multicolumn{1}{||c|}{\begin{tabular}[c]{@{}c@{}}0.4\\ $\pm$0.06\end{tabular}} & \begin{tabular}[c]{@{}c@{}}0.48\\ $\pm$0.07\end{tabular} & \multicolumn{1}{||c|}{\begin{tabular}[c]{@{}c@{}}0.29\\ $\pm$0.06\end{tabular}} & \begin{tabular}[c]{@{}c@{}}0.67\\ $\pm$0.07\end{tabular} & \multicolumn{1}{||c|}{\begin{tabular}[c]{@{}c@{}}0.19\\ $\pm$0.03\end{tabular}} & \begin{tabular}[c]{@{}c@{}}0.76\\ $\pm$0.05\end{tabular} \\ \hline
\multicolumn{1}{|c|}{\begin{tabular}[c]{@{}c@{}}Continual \\ CRUST\end{tabular}} & \multicolumn{1}{c|}{\begin{tabular}[c]{@{}c@{}}0.88\\ $\pm$0.02\end{tabular}} & \begin{tabular}[c]{@{}c@{}}0.06\\ $\pm$0.03\end{tabular} & \multicolumn{1}{||c|}{\begin{tabular}[c]{@{}c@{}}0.87\\ $\pm$0.02\end{tabular}} & \begin{tabular}[c]{@{}c@{}}0.08\\ $\pm$0.02\end{tabular} & \multicolumn{1}{||c|}{\begin{tabular}[c]{@{}c@{}}0.84\\ $\pm$0.04\end{tabular}} & \begin{tabular}[c]{@{}c@{}}0.05\\ $\pm$0.02\end{tabular} & \multicolumn{1}{||c|}{\begin{tabular}[c]{@{}c@{}}0.8\\ $\pm$0.02\end{tabular}} & \begin{tabular}[c]{@{}c@{}}0.06\\ $\pm$0.02\end{tabular} & \multicolumn{1}{||c|}{\begin{tabular}[c]{@{}c@{}}0.78\\ $\pm$0.04\end{tabular}} & \begin{tabular}[c]{@{}c@{}}0.06\\ $\pm$0.04\end{tabular} & \multicolumn{1}{||c|}{\begin{tabular}[c]{@{}c@{}}0.65\\ $\pm$0.08\end{tabular}} & \begin{tabular}[c]{@{}c@{}}0.16\\ $\pm$0.05\end{tabular} \\ \hline
\multicolumn{1}{|c|}{\begin{tabular}[c]{@{}c@{}}Continual \\ CosineCRUST\end{tabular}} & \multicolumn{1}{c|}{\begin{tabular}[c]{@{}c@{}}0.89\\ $\pm$0.02\end{tabular}} & \begin{tabular}[c]{@{}c@{}}0.05\\ $\pm$0.03\end{tabular} & \multicolumn{1}{||c|}{\begin{tabular}[c]{@{}c@{}}0.86\\ $\pm$0.03\end{tabular}} & \begin{tabular}[c]{@{}c@{}}0.06\\ $\pm$0.03\end{tabular} & \multicolumn{1}{||c|}{\begin{tabular}[c]{@{}c@{}}0.83\\ $\pm$0.03\end{tabular}} & \begin{tabular}[c]{@{}c@{}}0.07\\ $\pm$0.05\end{tabular} & \multicolumn{1}{||c|}{\begin{tabular}[c]{@{}c@{}}0.81\\ $\pm$0.02\end{tabular}} & \begin{tabular}[c]{@{}c@{}}0.07\\ $\pm$0.04\end{tabular} & \multicolumn{1}{||c|}{\begin{tabular}[c]{@{}c@{}}0.73\\ $\pm$0.02\end{tabular}} & \begin{tabular}[c]{@{}c@{}}0.11\\ $\pm$0.05\end{tabular} & \multicolumn{1}{||c|}{\begin{tabular}[c]{@{}c@{}}0.65\\ $\pm$0.03\end{tabular}} & \begin{tabular}[c]{@{}c@{}}0.12\\ $\pm$0.03\end{tabular} \\ \hline
\end{tabular}}
\caption{Final evaluation accuracy and forgetting on PathMNIST+ for varying levels of label-flipping noise}
\end{table}

\begin{figure}[H]
  \centering
   \includegraphics[width=0.45\linewidth]{graphics/PathMNIST_accuracy.png}
   \includegraphics[width=0.45\linewidth]{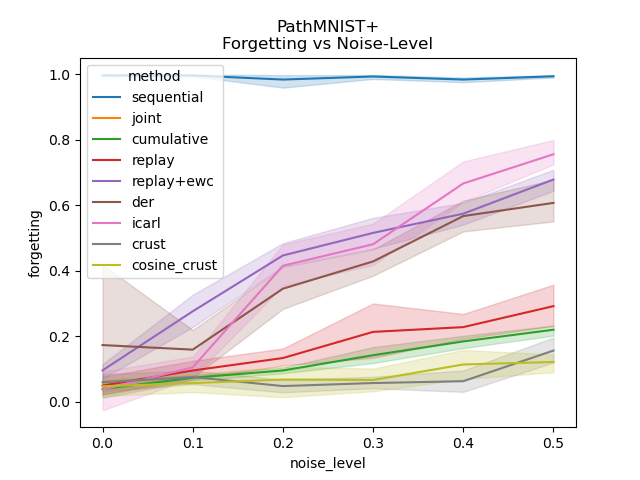}
   \caption{Looking at the final accuracy (left) and forgetting metric (right) at different label flipping noise levels (0.0-0.5) for PathMNIST+ for different strategies}
   \label{fig:pathmnist_noise}
\end{figure}

\newpage
\subsection{MNIST: Instance Noise}
\noindent\hspace{0pt}\begin{minipage}{0.99\textwidth}
\RaggedRight
In the following two experiments, we examine the robustness of the continual learning strategies to perturbations on the data instances as a function of the number of perturbed images. For these experiments, we look at the effect of varying the proportion of perturbed instances over ratios of \{0.2, 0.4, 0.6, and 0.8\}. For these experiments with MNIST, we use a Coreset size of 128 samples per class.

\indent\hspace{0.5cm} In Table \ref{tab:mnist_perturbed}, we see that most methods are not strongly affected by noisy samples. We expect this is because of the robustness of convolutional neural networks to small perturbations. Similarly, we expect this is because even though we used a large amount of noise (90\% of pixels perturbed), we added salt and pepper noise via a linear combination with the original image, which prevented the noise from overpowering the signal too strongly.
\end{minipage}

\begin{table}[H]
\centering
\begin{tabular}{c|cc|cc|cc|cc|}
\cline{2-9}
\textbf{} & \multicolumn{2}{c|}{\textbf{Noise=0.2}} & \multicolumn{2}{||c|}{\textbf{Noise=0.4}} & \multicolumn{2}{||c|}{\textbf{Noise=0.6}} & \multicolumn{2}{||c|}{\textbf{Noise=0.8}} \\ \hline
\multicolumn{1}{|c|}{\textbf{Algorithm}} & \multicolumn{1}{c|}{\textit{\textbf{Acc}}} & \textit{\textbf{Forg.}} & \multicolumn{1}{||c|}{\textit{\textbf{Acc}}} & \textit{\textbf{Forg.}} & \multicolumn{1}{||c|}{\textit{\textbf{Acc}}} & \textit{\textbf{Forg.}} & \multicolumn{1}{||c|}{\textit{\textbf{Acc}}} & \textit{\textbf{Forg.}} \\ \hline
\multicolumn{1}{|c|}{\begin{tabular}[c]{@{}c@{}}Naive \\ Sequential \\ Learner\end{tabular}} & \multicolumn{1}{c|}{\begin{tabular}[c]{@{}c@{}}0.11\\ $\pm$0.0\end{tabular}} & \begin{tabular}[c]{@{}c@{}}1.0\\ $\pm$0.0\end{tabular} & \multicolumn{1}{||c|}{\begin{tabular}[c]{@{}c@{}}0.11\\ $\pm$0.0\end{tabular}} & \begin{tabular}[c]{@{}c@{}}1.0\\ $\pm$0.0\end{tabular} & \multicolumn{1}{||c|}{\begin{tabular}[c]{@{}c@{}}0.11\\ $\pm$0.0\end{tabular}} & \begin{tabular}[c]{@{}c@{}}1.0\\ $\pm$0.0\end{tabular} & \multicolumn{1}{||c|}{\begin{tabular}[c]{@{}c@{}}0.11\\ $\pm$0.0\end{tabular}} & \begin{tabular}[c]{@{}c@{}}1.0\\ $\pm$0.0\end{tabular} \\ \hline
\multicolumn{1}{|c|}{\begin{tabular}[c]{@{}c@{}}Joint \\ Learner\end{tabular}} & \multicolumn{1}{c|}{\begin{tabular}[c]{@{}c@{}}0.99\\ $\pm$0.0\end{tabular}} & \begin{tabular}[c]{@{}c@{}} N/A \end{tabular} & \multicolumn{1}{||c|}{\begin{tabular}[c]{@{}c@{}}0.99\\ $\pm$0.0\end{tabular}} & \begin{tabular}[c]{@{}c@{}} N/A \end{tabular} & \multicolumn{1}{||c|}{\begin{tabular}[c]{@{}c@{}}0.99\\ $\pm$0.0\end{tabular}} & \begin{tabular}[c]{@{}c@{}} N/A \end{tabular} & \multicolumn{1}{||c|}{\begin{tabular}[c]{@{}c@{}}0.98\\ $\pm$0.0\end{tabular}} & \begin{tabular}[c]{@{}c@{}} N/A \end{tabular} \\ \hline
\multicolumn{1}{|c|}{\begin{tabular}[c]{@{}c@{}}Cumulative \\ Learner\end{tabular}} & \multicolumn{1}{c|}{\begin{tabular}[c]{@{}c@{}}0.99\\ $\pm$0.0\end{tabular}} & \begin{tabular}[c]{@{}c@{}}0.0\\ $\pm$0.0\end{tabular} & \multicolumn{1}{||c|}{\begin{tabular}[c]{@{}c@{}}0.99\\ $\pm$0.0\end{tabular}} & \begin{tabular}[c]{@{}c@{}}0.0\\ $\pm$0.0\end{tabular} & \multicolumn{1}{||c|}{\begin{tabular}[c]{@{}c@{}}0.99\\ $\pm$0.0\end{tabular}} & \begin{tabular}[c]{@{}c@{}}0.0\\ $\pm$0.0\end{tabular} & \multicolumn{1}{||c|}{\begin{tabular}[c]{@{}c@{}}0.99\\ $\pm$0.0\end{tabular}} & \begin{tabular}[c]{@{}c@{}}0.0\\ $\pm$0.0\end{tabular} \\ \hline
\multicolumn{1}{|c|}{\begin{tabular}[c]{@{}c@{}}Random \\ Replay\end{tabular}} & \multicolumn{1}{c|}{\begin{tabular}[c]{@{}c@{}}0.95\\ $\pm$0.01\end{tabular}} & \begin{tabular}[c]{@{}c@{}}0.04\\ $\pm$0.0\end{tabular} & \multicolumn{1}{||c|}{\begin{tabular}[c]{@{}c@{}}0.95\\ $\pm$0.01\end{tabular}} & \begin{tabular}[c]{@{}c@{}}0.05\\ $\pm$0.0\end{tabular} & \multicolumn{1}{||c|}{\begin{tabular}[c]{@{}c@{}}0.94\\ $\pm$0.01\end{tabular}} & \begin{tabular}[c]{@{}c@{}}0.05\\ $\pm$0.0\end{tabular} & \multicolumn{1}{||c|}{\begin{tabular}[c]{@{}c@{}}0.93\\ $\pm$0.01\end{tabular}} & \begin{tabular}[c]{@{}c@{}}0.05\\ $\pm$0.0\end{tabular} \\ \hline
\multicolumn{1}{|c|}{\begin{tabular}[c]{@{}c@{}}Random \\ Replay \\ w/EWC\end{tabular}} & \multicolumn{1}{c|}{\begin{tabular}[c]{@{}c@{}}0.97\\ $\pm$0.0\end{tabular}} & \begin{tabular}[c]{@{}c@{}}0.03\\ $\pm$0.0\end{tabular} & \multicolumn{1}{||c|}{\begin{tabular}[c]{@{}c@{}}0.97\\ $\pm$0.0\end{tabular}} & \begin{tabular}[c]{@{}c@{}}0.03\\ $\pm$0.01\end{tabular} & \multicolumn{1}{||c|}{\begin{tabular}[c]{@{}c@{}}0.97\\ $\pm$0.0\end{tabular}} & \begin{tabular}[c]{@{}c@{}}0.03\\ $\pm$0.0\end{tabular} & \multicolumn{1}{||c|}{\begin{tabular}[c]{@{}c@{}}0.96\\ $\pm$0.0\end{tabular}} & \begin{tabular}[c]{@{}c@{}}0.04\\ $\pm$0.0\end{tabular} \\ \hline
\multicolumn{1}{|c|}{\begin{tabular}[c]{@{}c@{}}Dark ER\end{tabular}} & \multicolumn{1}{c|}{\begin{tabular}[c]{@{}c@{}}0.96\\ $\pm$0.0\end{tabular}} & \begin{tabular}[c]{@{}c@{}}0.03\\ $\pm$0.0\end{tabular} & \multicolumn{1}{||c|}{\begin{tabular}[c]{@{}c@{}}0.96\\ $\pm$0.0\end{tabular}} & \begin{tabular}[c]{@{}c@{}}0.03\\ $\pm$0.0\end{tabular} & \multicolumn{1}{||c|}{\begin{tabular}[c]{@{}c@{}}0.94\\ $\pm$0.01\end{tabular}} & \begin{tabular}[c]{@{}c@{}}0.03\\ $\pm$0.0\end{tabular} & \multicolumn{1}{||c|}{\begin{tabular}[c]{@{}c@{}}0.94\\ $\pm$0.01\end{tabular}} & \begin{tabular}[c]{@{}c@{}}0.03\\ $\pm$0.0\end{tabular} \\ \hline
\multicolumn{1}{|c|}{\begin{tabular}[c]{@{}c@{}}iCaRL\end{tabular}} & \multicolumn{1}{c|}{\begin{tabular}[c]{@{}c@{}}0.92\\ $\pm$0.01\end{tabular}} & \begin{tabular}[c]{@{}c@{}}-0.05\\ $\pm$0.01\end{tabular} & \multicolumn{1}{||c|}{\begin{tabular}[c]{@{}c@{}}0.91\\ $\pm$0.01\end{tabular}} & \begin{tabular}[c]{@{}c@{}}-0.03\\ $\pm$0.02\end{tabular} & \multicolumn{1}{||c|}{\begin{tabular}[c]{@{}c@{}}0.92\\ $\pm$0.01\end{tabular}} & \begin{tabular}[c]{@{}c@{}}-0.03\\ $\pm$0.04\end{tabular} & \multicolumn{1}{||c|}{\begin{tabular}[c]{@{}c@{}}0.92\\ $\pm$0.01\end{tabular}} & \begin{tabular}[c]{@{}c@{}}-0.02\\ $\pm$0.02\end{tabular} \\ \hline
\multicolumn{1}{|c|}{\begin{tabular}[c]{@{}c@{}}Continual \\ CRUST\end{tabular}} & \multicolumn{1}{c|}{\begin{tabular}[c]{@{}c@{}}0.97\\ $\pm$0.0\end{tabular}} & \begin{tabular}[c]{@{}c@{}}0.02\\ $\pm$0.0\end{tabular} & \multicolumn{1}{||c|}{\begin{tabular}[c]{@{}c@{}}0.97\\ $\pm$0.0\end{tabular}} & \begin{tabular}[c]{@{}c@{}}0.03\\ $\pm$0.0\end{tabular} & \multicolumn{1}{||c|}{\begin{tabular}[c]{@{}c@{}}0.97\\ $\pm$0.0\end{tabular}} & \begin{tabular}[c]{@{}c@{}}0.02\\ $\pm$0.0\end{tabular} & \multicolumn{1}{||c|}{\begin{tabular}[c]{@{}c@{}}0.96\\ $\pm$0.0\end{tabular}} & \begin{tabular}[c]{@{}c@{}}0.03\\ $\pm$0.0\end{tabular} \\ \hline
\multicolumn{1}{|c|}{\begin{tabular}[c]{@{}c@{}}Continual \\ CosineCRUST\end{tabular}} & \multicolumn{1}{c|}{\begin{tabular}[c]{@{}c@{}}0.98\\ $\pm$0.0\end{tabular}} & \begin{tabular}[c]{@{}c@{}}0.02\\ $\pm$0.0\end{tabular} & \multicolumn{1}{||c|}{\begin{tabular}[c]{@{}c@{}}0.97\\ $\pm$0.0\end{tabular}} & \begin{tabular}[c]{@{}c@{}}0.02\\ $\pm$0.0\end{tabular} & \multicolumn{1}{||c|}{\begin{tabular}[c]{@{}c@{}}0.97\\ $\pm$0.0\end{tabular}} & \begin{tabular}[c]{@{}c@{}}0.02\\ $\pm$0.0\end{tabular} & \multicolumn{1}{||c|}{\begin{tabular}[c]{@{}c@{}}0.97\\ $\pm$0.0\end{tabular}} & \begin{tabular}[c]{@{}c@{}}0.02\\ $\pm$0.0\end{tabular} \\ \hline
\end{tabular}
\caption{Final evaluation accuracy and forgetting on MNIST for varying amounts of perturbed instances w/ 128 samples per class for the Coreset}
\label{tab:mnist_perturbed}
\end{table}

\newpage
\subsection{MSTAR: Instance Noise}
\noindent\hspace{0pt}\begin{minipage}{0.99\textwidth}
\RaggedRight

In the second instance perturbation additional study, we look at perturbations on MSTAR with a Coreset size of 30 instance per class. In Table \ref{tab:mstar_perturbed}, we see that some of the methods are weakly affected by added instance-level noise. Continual (Cosine)CRUST do exhibit better performance with less forgetting compared to the other strategies, suggesting that it is more noise-tolerant w.r.t. instance-level noise.

\end{minipage}

\begin{table}[H]
\centering
\begin{tabular}{c|cc|cc|cc|cc|}
\cline{2-9}
\textbf{} & \multicolumn{2}{c|}{\textbf{Noise=0.2}} & \multicolumn{2}{||c|}{\textbf{Noise=0.4}} & \multicolumn{2}{||c|}{\textbf{Noise=0.6}} & \multicolumn{2}{||c|}{\textbf{Noise=0.8}} \\ \hline
\multicolumn{1}{|c|}{\textbf{Algorithm}} & \multicolumn{1}{c|}{\textit{\textbf{Acc}}} & \textit{\textbf{Forg.}} & \multicolumn{1}{||c|}{\textit{\textbf{Acc}}} & \textit{\textbf{Forg.}} & \multicolumn{1}{||c|}{\textit{\textbf{Acc}}} & \textit{\textbf{Forg.}} & \multicolumn{1}{||c|}{\textit{\textbf{Acc}}} & \textit{\textbf{Forg.}} \\ \hline
\multicolumn{1}{|c|}{\begin{tabular}[c]{@{}c@{}}Naive \\ Sequential \\ Learner\end{tabular}} & \multicolumn{1}{c|}{\begin{tabular}[c]{@{}c@{}}0.11\\ $\pm$0.0\end{tabular}} & \begin{tabular}[c]{@{}c@{}}0.99\\ $\pm$0.0\end{tabular} & \multicolumn{1}{||c|}{\begin{tabular}[c]{@{}c@{}}0.11\\ $\pm$0.0\end{tabular}} & \begin{tabular}[c]{@{}c@{}}0.97\\ $\pm$0.04\end{tabular} & \multicolumn{1}{||c|}{\begin{tabular}[c]{@{}c@{}}0.11\\ $\pm$0.0\end{tabular}} & \begin{tabular}[c]{@{}c@{}}0.97\\ $\pm$0.04\end{tabular} & \multicolumn{1}{||c|}{\begin{tabular}[c]{@{}c@{}}0.11\\ $\pm$0.0\end{tabular}} & \begin{tabular}[c]{@{}c@{}}0.97\\ $\pm$0.04\end{tabular} \\ \hline
\multicolumn{1}{|c|}{\begin{tabular}[c]{@{}c@{}}Joint \\ Learner\end{tabular}} & \multicolumn{1}{c|}{\begin{tabular}[c]{@{}c@{}}0.98\\ $\pm$0.01\end{tabular}} & \begin{tabular}[c]{@{}c@{}} N/A \end{tabular} & \multicolumn{1}{||c|}{\begin{tabular}[c]{@{}c@{}}0.97\\ $\pm$0.01\end{tabular}} & \begin{tabular}[c]{@{}c@{}} N/A \end{tabular} & \multicolumn{1}{||c|}{\begin{tabular}[c]{@{}c@{}}0.98\\ $\pm$0.01\end{tabular}} & \begin{tabular}[c]{@{}c@{}} N/A \end{tabular} & \multicolumn{1}{||c|}{\begin{tabular}[c]{@{}c@{}}0.98\\ $\pm$0.01\end{tabular}} & \begin{tabular}[c]{@{}c@{}} N/A \end{tabular} \\ \hline
\multicolumn{1}{|c|}{\begin{tabular}[c]{@{}c@{}}Cumulative \\ Learner\end{tabular}} & \multicolumn{1}{c|}{\begin{tabular}[c]{@{}c@{}}0.99\\ $\pm$0.0\end{tabular}} & \begin{tabular}[c]{@{}c@{}}-0.03\\ $\pm$0.02\end{tabular} & \multicolumn{1}{||c|}{\begin{tabular}[c]{@{}c@{}}0.99\\ $\pm$0.01\end{tabular}} & \begin{tabular}[c]{@{}c@{}}0.02\\ $\pm$0.04\end{tabular} & \multicolumn{1}{||c|}{\begin{tabular}[c]{@{}c@{}}0.99\\ $\pm$0.0\end{tabular}} & \begin{tabular}[c]{@{}c@{}}-0.0\\ $\pm$0.01\end{tabular} & \multicolumn{1}{||c|}{\begin{tabular}[c]{@{}c@{}}0.98\\ $\pm$0.01\end{tabular}} & \begin{tabular}[c]{@{}c@{}}-0.03\\ $\pm$0.03\end{tabular} \\ \hline
\multicolumn{1}{|c|}{\begin{tabular}[c]{@{}c@{}}Random \\ Replay\end{tabular}} & \multicolumn{1}{c|}{\begin{tabular}[c]{@{}c@{}}0.93\\ $\pm$0.01\end{tabular}} & \begin{tabular}[c]{@{}c@{}}0.04\\ $\pm$0.03\end{tabular} & \multicolumn{1}{||c|}{\begin{tabular}[c]{@{}c@{}}0.91\\ $\pm$0.02\end{tabular}} & \begin{tabular}[c]{@{}c@{}}0.03\\ $\pm$0.04\end{tabular} & \multicolumn{1}{||c|}{\begin{tabular}[c]{@{}c@{}}0.91\\ $\pm$0.02\end{tabular}} & \begin{tabular}[c]{@{}c@{}}0.04\\ $\pm$0.01\end{tabular} & \multicolumn{1}{||c|}{\begin{tabular}[c]{@{}c@{}}0.89\\ $\pm$0.01\end{tabular}} & \begin{tabular}[c]{@{}c@{}}0.08\\ $\pm$0.04\end{tabular} \\ \hline
\multicolumn{1}{|c|}{\begin{tabular}[c]{@{}c@{}}Random \\ Replay \\ w/EWC\end{tabular}} & \multicolumn{1}{c|}{\begin{tabular}[c]{@{}c@{}}0.75\\ $\pm$0.05\end{tabular}} & \begin{tabular}[c]{@{}c@{}}0.22\\ $\pm$0.03\end{tabular} & \multicolumn{1}{||c|}{\begin{tabular}[c]{@{}c@{}}0.73\\ $\pm$0.06\end{tabular}} & \begin{tabular}[c]{@{}c@{}}0.26\\ $\pm$0.02\end{tabular} & \multicolumn{1}{||c|}{\begin{tabular}[c]{@{}c@{}}0.67\\ $\pm$0.05\end{tabular}} & \begin{tabular}[c]{@{}c@{}}0.26\\ $\pm$0.02\end{tabular} & \multicolumn{1}{||c|}{\begin{tabular}[c]{@{}c@{}}0.62\\ $\pm$0.02\end{tabular}} & \begin{tabular}[c]{@{}c@{}}0.32\\ $\pm$0.01\end{tabular} \\ \hline
\multicolumn{1}{|c|}{\begin{tabular}[c]{@{}c@{}}Dark ER\end{tabular}} & \multicolumn{1}{c|}{\begin{tabular}[c]{@{}c@{}}0.93\\ $\pm$0.0\end{tabular}} & \begin{tabular}[c]{@{}c@{}}0.06\\ $\pm$0.03\end{tabular} & \multicolumn{1}{||c|}{\begin{tabular}[c]{@{}c@{}}0.93\\ $\pm$0.01\end{tabular}} & \begin{tabular}[c]{@{}c@{}}0.06\\ $\pm$0.02\end{tabular} & \multicolumn{1}{||c|}{\begin{tabular}[c]{@{}c@{}}0.93\\ $\pm$0.02\end{tabular}} & \begin{tabular}[c]{@{}c@{}}0.04\\ $\pm$0.04\end{tabular} & \multicolumn{1}{||c|}{\begin{tabular}[c]{@{}c@{}}0.9\\ $\pm$0.02\end{tabular}} & \begin{tabular}[c]{@{}c@{}}0.06\\ $\pm$0.02\end{tabular} \\ \hline
\multicolumn{1}{|c|}{\begin{tabular}[c]{@{}c@{}}iCaRL\end{tabular}} & \multicolumn{1}{c|}{\begin{tabular}[c]{@{}c@{}}0.86\\ $\pm$0.02\end{tabular}} & \begin{tabular}[c]{@{}c@{}}0.09\\ $\pm$0.01\end{tabular} & \multicolumn{1}{||c|}{\begin{tabular}[c]{@{}c@{}}0.83\\ $\pm$0.02\end{tabular}} & \begin{tabular}[c]{@{}c@{}}0.13\\ $\pm$0.02\end{tabular} & \multicolumn{1}{||c|}{\begin{tabular}[c]{@{}c@{}}0.77\\ $\pm$0.02\end{tabular}} & \begin{tabular}[c]{@{}c@{}}0.12\\ $\pm$0.04\end{tabular} & \multicolumn{1}{||c|}{\begin{tabular}[c]{@{}c@{}}0.7\\ $\pm$0.06\end{tabular}} & \begin{tabular}[c]{@{}c@{}}0.18\\ $\pm$0.06\end{tabular} \\ \hline
\multicolumn{1}{|c|}{\begin{tabular}[c]{@{}c@{}}Continual \\ CRUST\end{tabular}} & \multicolumn{1}{c|}{\begin{tabular}[c]{@{}c@{}}0.91\\ $\pm$0.01\end{tabular}} & \begin{tabular}[c]{@{}c@{}}-0.01\\ $\pm$0.09\end{tabular} & \multicolumn{1}{||c|}{\begin{tabular}[c]{@{}c@{}}0.91\\ $\pm$0.01\end{tabular}} & \begin{tabular}[c]{@{}c@{}}-0.07\\ $\pm$0.09\end{tabular} & \multicolumn{1}{||c|}{\begin{tabular}[c]{@{}c@{}}0.9\\ $\pm$0.02\end{tabular}} & \begin{tabular}[c]{@{}c@{}}-0.07\\ $\pm$0.09\end{tabular} & \multicolumn{1}{||c|}{\begin{tabular}[c]{@{}c@{}}0.88\\ $\pm$0.02\end{tabular}} & \begin{tabular}[c]{@{}c@{}}-0.03\\ $\pm$0.1\end{tabular} \\ \hline
\multicolumn{1}{|c|}{\begin{tabular}[c]{@{}c@{}}Continual \\ CosineCRUST\end{tabular}} & \multicolumn{1}{c|}{\begin{tabular}[c]{@{}c@{}}0.94\\ $\pm$0.01\end{tabular}} & \begin{tabular}[c]{@{}c@{}}-0.13\\ $\pm$0.07\end{tabular} & \multicolumn{1}{||c|}{\begin{tabular}[c]{@{}c@{}}0.96\\ $\pm$0.01\end{tabular}} & \begin{tabular}[c]{@{}c@{}}-0.04\\ $\pm$0.05\end{tabular} & \multicolumn{1}{||c|}{\begin{tabular}[c]{@{}c@{}}0.95\\ $\pm$0.01\end{tabular}} & \begin{tabular}[c]{@{}c@{}}-0.15\\ $\pm$0.06\end{tabular} & \multicolumn{1}{||c|}{\begin{tabular}[c]{@{}c@{}}0.95\\ $\pm$0.01\end{tabular}} & \begin{tabular}[c]{@{}c@{}}-0.06\\ $\pm$0.04\end{tabular} \\ \hline
\end{tabular}
\caption{Final evaluation accuracy and forgetting on MSTAR for varying amounts of perturbed instances w/ 30 samples per class for the Coreset}
\label{tab:mstar_perturbed}
\end{table}

\newpage
\subsection{FashionMNIST: Purity Under Instance Noise w/ Total Corruption}

\noindent\hspace{0pt}\begin{minipage}{0.99\textwidth}
\RaggedRight
We run a follow-up experiment to Section \ref{sec:purity_main} where instead of corrupting images with salt and pepper noise, we completely corrupt the instances (i.e., ``noisy'' instances are purely uniform random noise). Experiments are run on FashionMNIST with a Coreset size of 128 samples per class and perturbed-to-unperturbed data ratios of \{0.0, 0.1, 0.2, 0.3, 0.4, 0.5\}. In this case (Table \ref{tab:data_purity_fashionmnist_2}), we see that Continual CRUST and Continual CosineCRUST select much cleaner Coresets (\textbf{purity more than }) for $\geq$ 0.4 fraction of noisy instances compared to random selection and compared to the salt-and-pepper setting in Table \ref{tab:data_purity_fashionmnist}. We do observe that CosineCRUST appears to be less noise-tolerant in this regime of total sample corruption, failing for a noise level of 0.5. We also observed that our model was overparameterized and thus, could still overfit the pure noise samples while achieving reasonable accuracy, which may explain lack of purity. We expect, if we re-ran these experiments with a model less prone to overfitting, our proposed approach may be able to find higher-purity Coresets.

\end{minipage}

\indent\hspace{0.5cm} 

\begin{table*}[h!]
\centering
\resizebox{0.95\textwidth}{!}{
\begin{tabular}{c|cccccc|}
\cline{2-7}
\multicolumn{1}{l|}{} & \multicolumn{6}{c|}{\textbf{FashionMNIST: Coreset Purity for Fully Corrupted Samples at Noise-Level N}} \\ \hline
\multicolumn{1}{|c|}{\textbf{Algorithm}} & \multicolumn{1}{c|}{\textbf{N=0.0}} & \multicolumn{1}{c|}{\textbf{N=0.1}} & \multicolumn{1}{c|}{\textbf{N=0.2}} & \multicolumn{1}{c|}{\textbf{N=0.3}} & \multicolumn{1}{c|}{\textbf{N=0.4}} & \textbf{N=0.5} \\ \hline
\multicolumn{1}{|c|}{Random Replay} & \multicolumn{1}{c|}{1.0 $\pm$ 0.0} & \multicolumn{1}{c|}{0.89 $\pm$ 0.02} & \multicolumn{1}{c|}{0.8 $\pm$ 0.03} & \multicolumn{1}{c|}{0.7 $\pm$ 0.04} & \multicolumn{1}{c|}{0.6 $\pm$ 0.05} & \multicolumn{1}{c|}{0.51 $\pm$ 0.05} \\ \hline
\multicolumn{1}{|c|}{Continual CRUST} & \multicolumn{1}{c|}{1.0 $\pm$ 0.0} & \multicolumn{1}{c|}{0.98 $\pm$ 0.06} & \multicolumn{1}{c|}{0.92 $\pm$ 0.16} & \multicolumn{1}{c|}{0.87 $\pm$ 0.2} & \multicolumn{1}{c|}{0.77 $\pm$ 0.24} & \multicolumn{1}{c|}{0.6 $\pm$ 0.29} \\ \hline
\multicolumn{1}{|c|}{Continual CosineCRUST} & \multicolumn{1}{c|}{1.0 $\pm$ 0.0} & \multicolumn{1}{c|}{0.96 $\pm$ 0.06} & \multicolumn{1}{c|}{0.86 $\pm$ 0.1} & \multicolumn{1}{c|}{0.76 $\pm$ 0.13} & \multicolumn{1}{c|}{0.65 $\pm$ 0.14} & \multicolumn{1}{c|}{0.53 $\pm$ 0.1} \\ \hline
\end{tabular}}
\caption{Coreset purity for FashionMNIST with perturbed data for salt and pepper noise (top) and samples consisting of uniform random noise (complete sample corruption) (bottom). }
\label{tab:data_purity_fashionmnist_2}
\end{table*}



\end{document}